\documentclass[10pt,twocolumn,letterpaper]{article}

\usepackage[pagenumbers]{cvpr}      %

\definecolor{cvprblue}{rgb}{0.21,0.49,0.74}

\def\paperID{13014} %
\def\confName{CVPR}
\def\confYear{2025}

\title{Feature-Preserving Mesh Decimation for Normal Integration}

\author{%
Moritz Heep\\%
PhenoRob\\
University of Bonn\\
{\tt\small mheep@uni-bonn.de}
\and
Sven Behnke\\
Autonomous Intelligent Systems\\
University of Bonn\\
{\tt\small behnke@ais.uni-bonn.de}
\and
Eduard Zell
\vspace{.5\normalbaselineskip}\\
Independent Researcher
\vspace{.5\normalbaselineskip}\\
{\tt\small ezell@hotmail.de}%
}

\usepackage{graphicx}
\usepackage{amsmath}
\usepackage{amssymb}
\usepackage{booktabs}

\usepackage{multirow}

\usepackage[english]{firstauthorslastname}

\newcommand{\Edges}{\mathcal{E}}
\newcommand{\Vertices}{\mathcal{V}}
\newcommand{\Faces}{\mathcal{F}}
\newcommand{\Pixels}{\mathcal{P}}
\newcommand{\Star}{\mathcal{S}}

\usepackage{color}
\usepackage{xcolor}
\definecolor{darkred}{rgb}{0.6,0,0}
\definecolor{darkgreen}{rgb}{0.1,0.3,0.1}
\definecolor{darkorange}{rgb}{0.6,0.4,0.2}
\definecolor{darkpurple}{rgb}{0.4,0.1,0.4}
\definecolor{lightblue}{rgb}{0.5,0.5,0.75}
\definecolor{blue}{rgb}{0,0,0.75}

\def\cpp{{C\nolinebreak[4]\hspace{-.05em}\raisebox{.4ex}{\tiny\bf ++}}}

\usepackage{tabularray}
\usepackage{placeins}
\usepackage{siunitx}

\usepackage[pagebackref,breaklinks,colorlinks,allcolors=cvprblue]{hyperref}

\begin{document}
\maketitle
\begin{abstract}
Normal integration reconstructs 3D surfaces from normal maps obtained \eg by photometric stereo. These normal maps capture surface details down to the pixel level but require large computational resources for integration at high resolutions. In this work, we replace the dense pixel grid with a sparse anisotropic triangle mesh prior to normal integration. We adapt the triangle mesh to the local geometry in the case of complex surface structures and remove oversampling from flat featureless regions.
For high-resolution images, the resulting compression reduces normal integration runtimes from hours to minutes while maintaining high surface accuracy.
Our main contribution is the derivation of the well-known quadric error measure from mesh decimation for screen space applications and its combination with optimal Delaunay triangulation.
Code is available at \url{https://moritzheep.github.io/anisotropic-screen-meshing}.
\end{abstract}
\begin{figure}[ht]
    \centering
    \begin{subfigure}[t]{.47\linewidth}
        \centering
        \normalsize{\textbf{Isotropic Remeshing}}\\
        \vspace{1mm}
        \includegraphics[trim={128 8 128 8}, clip, width=\linewidth]{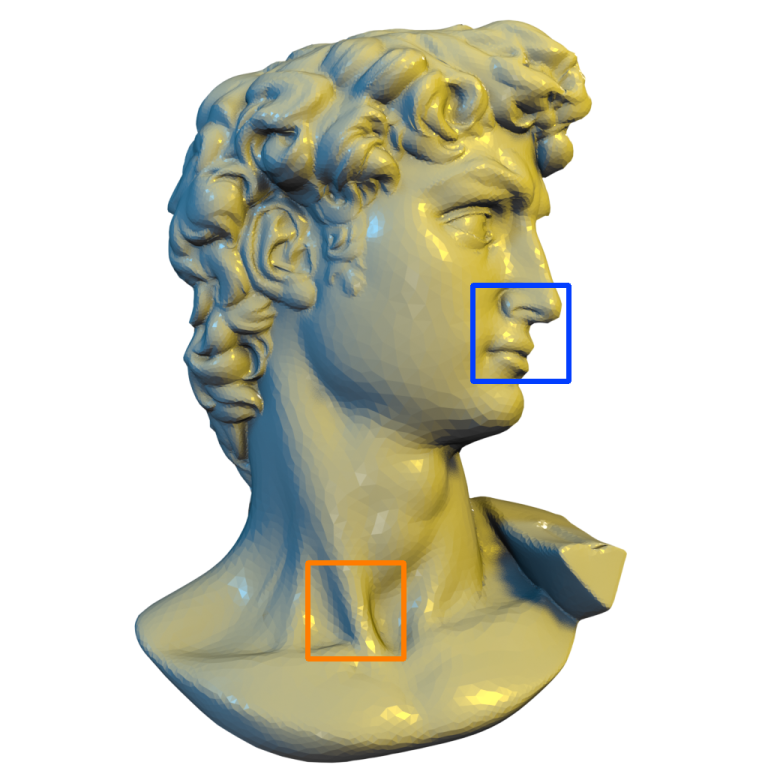}
        \\
        \vspace{1.5mm}
        \includegraphics[width=.48\linewidth]{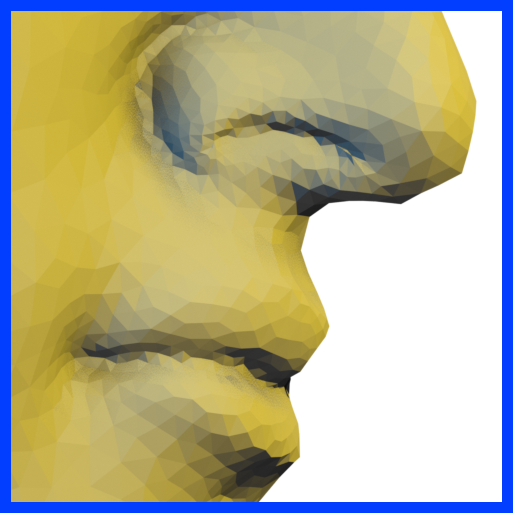}
        \includegraphics[width=.48\linewidth]{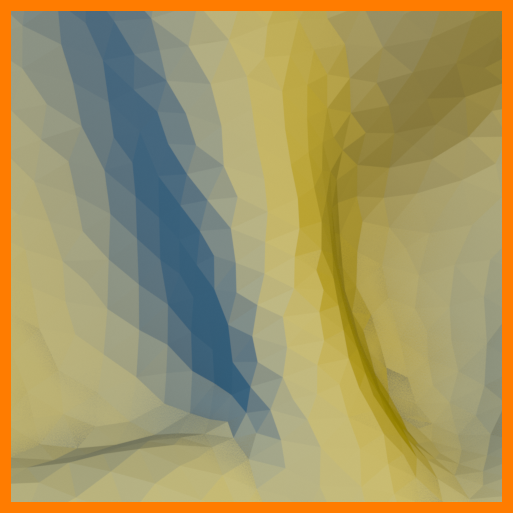}
        21k Vertices
    \end{subfigure}
    \hspace{1mm}
    \begin{subfigure}[t]{.47\linewidth}
        \centering
        \normalsize{\textbf{Our Decimation}}\\
        \vspace{1mm}
        \includegraphics[trim={128 8 128 8}, clip, width=\linewidth]{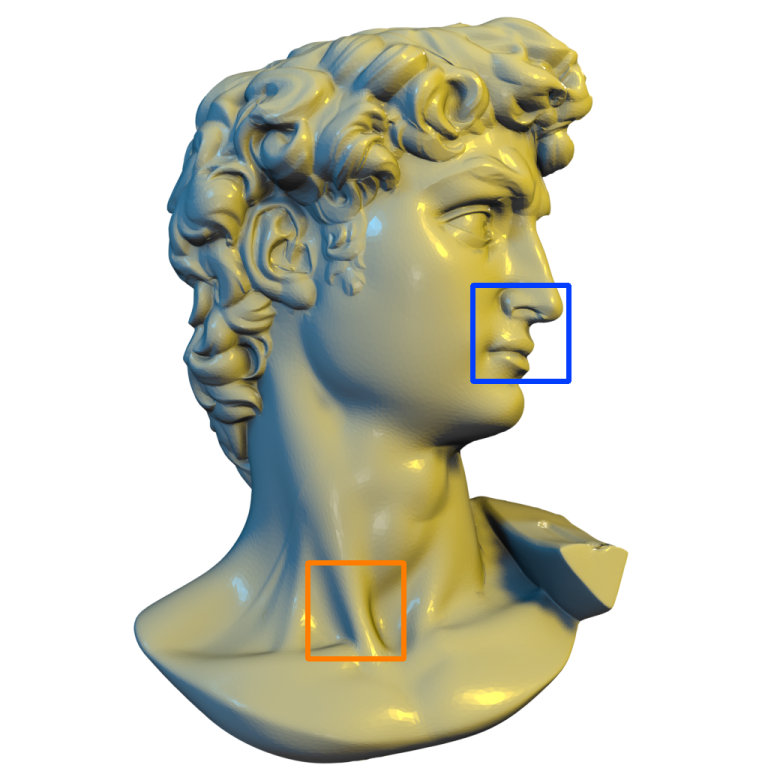}
        \\
        \vspace{1.5mm}
        \includegraphics[width=.48\linewidth]{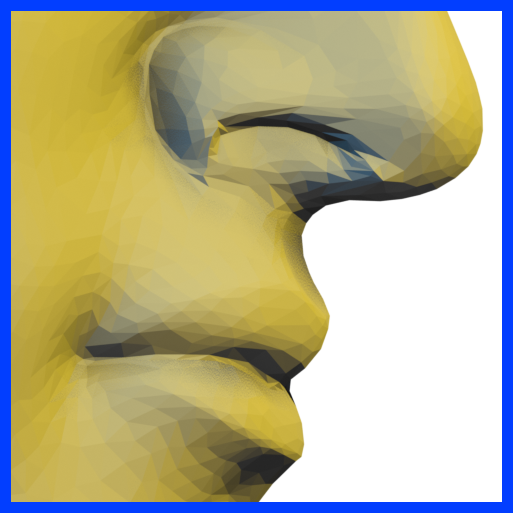}
        \includegraphics[width=.48\linewidth]{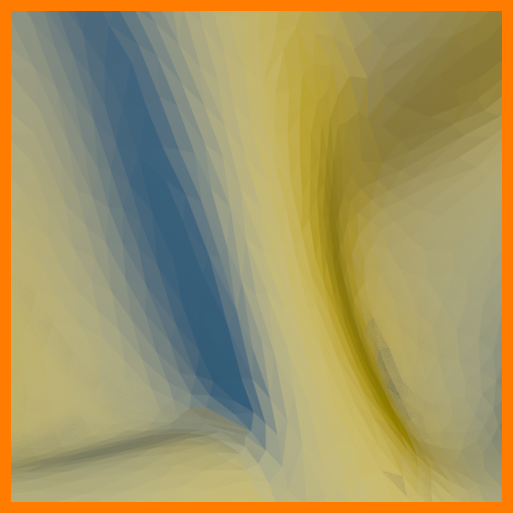}
        20k Vertices
    \end{subfigure}
    \vspace{-1mm}
    \caption{Visual comparison between the isotropic remeshing~\cite{Heep:2025} (\emph{left}) and our anisotropic decimation (\emph{right}). In both examples, the $2048^2$ normal map is compressed into a triangle mesh with approximately 20k vertices \emph{before} integration. By aligning vertices and edges to ridges and furrows, we achieve higher accuracy with the same number of vertices.}
    \label{fig:not_a_teaser}
\end{figure}
\section{Introduction} %
\label{sec:intro}
Normal maps can be estimated from images through shape-from-shading~\cite{Horn:1970} or photometric stereo~\cite{Woodham:1980} down to the pixel level, capturing delicate surface details. Normal integration reconstructs the underlying surface using these normals.
The surface is usually estimated as a pixel-based depth map~\cite{Du:2007} %
by solving a linear system. At the pixel level, doubling the geometric resolution requires doubling both, image height and width. Due to this quadratic growth of the number of variables in the linear system, normal integration scales poorly to higher image resolutions. The fundamental problem of a regular grid is that fine details somewhere on screen require increasing the resolution everywhere on screen.
\par
Recent research proposes substituting the pixel grid with an adaptive 2D triangle mesh \emph{prior to} integration~\cite{Heep:2025}. Triangle meshes allow local control of the resolution: the resolution remains high in areas of delicate details and is lowered within smooth areas. While this reduces the number of vertices compared to the number of pixels (and consequently runtime), this approach is isotropic and struggles with ridges and furrows by design.
Our proposed method overcomes this limitation by shifting to an anisotropic formulation: 
By analysing the normal maps, we align edges and vertices of the mesh with ridges and furrows of the underlying geometry. Our mesh decimation technique maintains delicate surface details more effectively than previous methods, even with high compression ratios, see~\cref{fig:not_a_teaser}. 
In summary, our contributions are:
\begin{itemize}
    \item We derive quadrics~\cite{Garland:1997}, originally designed for 3D meshes and crucial to mesh decimation methods but take normal maps as input to facilitate mesh decimation in screen space.
    \item By taking advantage of the similarity between quadrics and a generalized Delaunay criterion~\cite{Chen:2004}, we align triangle edges with ridges and furrows, thus outperforming previous methods~\cite{Heep:2025}. 
    \item We present a simple iterative algorithm, consisting only of three local mesh operations: edge collapse, edge flip, and vertex position update.
\end{itemize}
Our results suggest that even at around 90\% compression, we achieve surfaces within the sub-millimetre range of pixel-based approaches. Furthermore, given a similar time range we can triangulate, decimate and integrate a 16MP normal map, while pixel-based integration succeeds only to integrate 4MP. A reference implementation is available under \url{http://moritzheep.github.io/anisotropic-screen-meshing}.
\section{Related Work} %
\paragraph{Normal Integration}
The core of recent normal integration methods is a functional~\cite{Du:2007} quantifying the difference between the gradients of the actual depth-map and observed gradients, \eg from photometric stereo. Normal integration is then achieved by finding the depth map that minimizes this functional. If the $L^2$ norm is used to measure the difference, the optimal depth map can be found by solving a Poisson equation in the form of a sparse linear system. This linear system is either found by discretising the functional itself~\cite{Durou:2007} or using functional analysis and discretising the resulting Poisson equation~\cite{Horn:1986}. An overview of the topic is given by Qu{\'e}au et al.~\cite{Queau:2018}. More recently, authors have raised concerns about various artefacts~\cite{Zhu:2020,Cao:2021} %
occurring in the functional setting. Although these artefacts can be avoided by using additional variables~\cite{Cao:2021} or larger stencil sizes for the partial derivatives~\cite{Zhou:2020}, replacing the functional over gradients by a functional over normals represents a more straightforward solution~\cite{Cao:2022, Heep:2022}. Our method is derived from the same normal-based functional~\cite{Heep:2025} and remains unaffected by the aforementioned artefacts.
\paragraph{Mesh-Based Integration}
Early mesh-based methods interpreted the pixel-grid as a grid of quadrilateral facets~\cite{Xie:2014, Xie:2015}. These methods iterate between aligning the quads with the normal directions and gluing adjacent quads into a continuous or even discontinuous~\cite{Xie:2019} surface. However, this approach maintains a one-to-one relation between pixels and quads. More recently, a variational approach to normal integration was introduced for arbitrary triangle meshes~\cite{Heep:2025} together with an isotropic screen space meshing algorithm. By calculating curvatures from the normal maps, the triangle density is adjusted locally, facilitating much sparser representations and reducing runtimes without sacrificing quality. Unlike this approach, we abandon isotropy in favour of an alignment of vertices and edges to ridges and furrows of the underlying geometry. The result is a much more faithful representation.
\paragraph{Remeshing and Decimation}
Modification or simplification of an existing triangle mesh has been extensively studied in the field of geometry processing~\cite{Khan:2020}:
Decimation aims to preserve the shape with a reduced number of vertices. Remeshing aims to improve mesh regularity and might increase or decrease the vertex number. A wide range of remeshing algorithms build either on Centroidal Voronoi Tessellations (CVT)~\cite{Du:1999} or Optimal Delaunay Triangulations (ODT)~\cite{Chen:2004}. Both approaches create uniform or isotropic meshes, \ie meshes with equilateral triangles of constant or varying size by iteratively moving vertices to the centroid of a local neighbourhood. Curvature-based densities~\cite{Alliez:2005, Chen:2012} are a common choice, as they ensure an accurate representation of the geometry~\cite{Dunyach:2013}.
Extending CVT or ODT to create anisotropic meshes is more elaborate. One option is to 'lift' vertex positions and normals into a six-dimensional space~\cite{Levy:2013,Zhong:2013,Nivoliers:2015}. The resulting CVT is isotropic in this space but anisotropic in 3D. Since the change in normals is proportional to curvature, these triangulations still implicitly adapt to curvature.
\par
Mesh simplification typically reduces the vertex count of an existing triangle mesh by collapsing edges that are inconsequential for the overall shape. Different metrics have been proposed to determine these collapse candidates, such as volume preservation~\cite{Lindstrom:1998} or an estimate of the distance to the original mesh~\cite{Garland:1997}. Especially the algorithm by Garland and Heckbert~\cite{Garland:1997} is part of most mesh-processing libraries. Both approaches use quadratic functions -- quadrics -- to determine how much a collapse would distort the original geometry. More recently, probabilistic versions of both algorithms have been proposed to handle noisy input data~\cite{Trettner:2020}.
Although devised in the context of mesh simplification, quadrics have also found application in remeshing: Xu et al.~\cite{Xu:2024} combine quadrics and CVT in an isotropic meshing approach. In contrast, our method uses anisotropy to achieve a more faithful representation.
Our method is partly inspired by the aforementioned methods, but differs significantly in one aspect: 
To overcome the runtime limitation of pixel-based integration, we have to introduce the triangulation \emph{before} normal integration. Hence, our method cannot rely on 3D geometry but can only use normal maps as input. 
\section{Screen Space Mesh Decimation} 
Transforming a regular pixel grid into an irregular 2D triangle mesh allows local adaptation to surface details. This can reduce runtime~\cite{Heep:2025} while maintaining reconstruction quality. However, the creation of the triangulation requires careful consideration:
First, the placement of vertices $v\in\Vertices$ but also the choice of edges $e\in\Edges$ to connect them is of utmost importance for an accurate yet sparse representation. Second, the vertex and edge placement must occur without knowing the 3D surface in advance. 
\par
In this work, we advance the idea of triangulating surfaces based on normal maps by introducing an algorithm that decimates near-redundant vertices to achieve high compression ratios. To maintain high-accuracy, the remaining vertices and edges are aligned to the underlying geometry. Our algorithm is guided by the objective function:
\begin{equation}
    \label{eqn:objective_function}
    E=E_\text{Geo}+\lambda\cdot E_\text{ODT}\,.
\end{equation}
For an existing surface, the two energy terms can be written as surface integrals over $\vec{x}\in\R^3$.
The first term estimates the deviation between the mesh and the underlying surface by relying on quadrics~\cite{Garland:1997, Xu:2024}: 
\begin{equation}
    \label{eqn:gh_objective}
    E_\text{Geo}(\vec{x}_1,...,\vec{x}_{|\Vertices|})=\sum_{v\in\Vertices}\int_{\Star_v}\langle\vec{n}(\vec{x}),\vec{x}_v-\vec{x}\rangle^2\exterior^2x,
\end{equation}
where $\vec{x}_v$ for $v\in\Vertices$ are the unknown 3D vertex positions and $\langle\cdot,\cdot\rangle$ is the standard scalar product. The star $\Star_v$ of $v$ is the surface patch formed by all triangles touching $v$. 
The second term
\begin{equation}
    \label{eqn:odt_objective}
    E_\text{ODT}(\vec{x}_1,...,\vec{x}_{|\Vertices|})=\sum_{v\in\Vertices}\int_{\Star_v}\|\vec{x}_v-\vec{x}\|^2\exterior^2x
\end{equation}
guarantees a well-defined behaviour and an even vertex distribution in flat areas. We will see in \cref{sec:generalized_delaunay} that $E_\text{ODT}$ is closely connected to Delaunay triangulations~\cite{Chen:2004}. This fundamental insight is crucial to align edges to features and to facilitate anisotropic meshing.
\par
While it is easy to compute the energy terms for an existing surface in 3D, the depth coordinate of each $\vec{x}$ is unknown in our case. A coarse 2D triangle mesh must be computed \emph{before} the normal integration to gain any speed-up. 
In the following, we will show how both energy functions can be evaluated in screen space and derive a novel quadric formulation with normal maps as input. %
In \cref{sec:algorithm}, we focus on the practical aspects and implementation details. We break down the mathematical concepts into simple mesh operations, \ie vertex relocation, edge flip and edge collapse.
\subsection{Calculating the Objective Function On Screen}
\label{sec:objective_function}
Image-based reconstruction can be understood as finding a projection
\begin{equation}
    \phi:\Omega\rightarrow\R^3
\end{equation}
from the image foreground $\Omega\subset\R^2$ into 3D space. While existing approaches parametrise the projection through a pixel-wise depth map, we choose a 2D triangle mesh instead. The projection $\phi$ is then determined by one depth value $z_v$ per vertex and bilinear interpolation everywhere else.
In \cref{sec:normal_integration}, we derive these depth values from normal maps by solving the normal integration problem for a 2D triangle mesh, c.f.~\cite{Heep:2025}.
\par
A unified treatment of $E_\text{Geo}$ and $E_\text{ODT}$ is obtained by introducing the norm
\begin{equation}
    \|\vec{x}\|_M^2:=\langle\vec{x},M\vec{x}\rangle
    \label{eqn:generalized_norm}
\end{equation}
with a constant $M=\1$ for $E_\text{ODT}$ and a locally varying $M(\vec{x})=\vec{n}(\vec{x})\cdot\vec{n}^t(\vec{x})$ for $E_\text{Geo}$. With this shorthand notation, both energy functions become a sum over 
\begin{equation}
    Q_v:=\sum_{f\in\Faces}\int_f\left\|\vec{x}_v-\vec{x}\right\|^2_{M(\vec{x})}\exterior^2x
\end{equation}
for each vertex and with the respective choice for $M$. Such quadratic functions $Q_v$ are commonly referred to as quadrics~\cite{Garland:1997}.
With a known projection $\phi$, this integral could be evaluated in coordinates by substituting 
$\vec{x}_v-\vec{x}\rightarrow\phi(\vec{u}_v)-\phi(\vec{u})$ where $\vec{u}$ and $\vec{u}_p$ are the respective 2D coordinates on screen.
However, runtime advantages are only achieved by decimating \emph{before} the integration, \ie before $\phi$ is known. Instead, we replace $\vec{x}_p-\vec{x}=J_f\cdot(\vec{u}_p-\vec{u})$ using a Taylor expansion where the face Jacobian $J_f$ of $\phi$ is constant due to linear interpolation.
\par
Unlike $\phi$ - which is only available after an expensive integration - the Jacobian $J_f$ can be obtained directly from normals: The Jacobian $J=(\partial_u\phi,\partial_v\phi)$ consists of two surface tangents as columns. Both tangents are orthogonal to the surface normals and hence
\begin{equation}
    \begin{aligned}
        \label{eqn:jacobian_ortho}
        \partial_u\phi  &=  \vec{e}_x-\frac{n_x}{n_z}\cdot\vec{e}_z  &
        \partial_v\phi  &=  \vec{e}_y-\frac{n_y}{n_z}\cdot\vec{e}_z
    \end{aligned}
\end{equation}
in the orthographic case and
\begin{equation}
    \begin{aligned}
        \label{eqn:jacobian_persp}
        \partial_i\phi  &=  \left(\partial_i\vec{r}-\frac{\langle\vec{n},\partial_i\vec{r}\rangle}{\langle\vec{n},\vec{r}\rangle}\cdot\vec{r}\right)\cdot z  &  \text{for }i\in\{u,v\}
    \end{aligned}
\end{equation}
in the perspective case where $\vec{r}$ is the camera ray given by the intrinsics matrix, \cf\eg~\cite{Heep:2025}. In the latter case, we adopt the weak perspective projection and assume a constant camera-to-object distance. These two equations allow calculating tangent vectors from normals maps and we will require both equations throughout \cref{sec:algorithm} to calculate normals for vertices, edges and faces.
\par
To determine the influence of a vertex on the mesh surface and to facilitate finding optimal positions (\cref{sec:tangential_smoothing}), we need to know the change in $Q_v$ if vertex $v$ is moved by $\delta\vec{x}_v$:
\begin{equation}
    Q_v(\delta\vec{x}_v):=\sum_{f\in\Faces_v}\int_f\|J_f(\vec{u}_v-\vec{u})+\delta\vec{x}_v\|^2_{M(\vec{x})}\exterior\Omega,
\end{equation}
where $\exterior\Omega=|J_f^tJ_f|^\frac{1}{2}\exterior^2u$ compensates for the distorted area on the screen due to foreshortening. This ensures that all faces contribute as if the integral was evaluated over the 3D surface.
\subsection{Generalized Delaunay Triangulations}
\label{sec:generalized_delaunay}
To achieve accurate mesh representations despite reducing the number of vertices, we must align edges and vertices to ridges and furrows of the underlying surface. Notably, $E_\text{ODT}$ in \cref{eqn:odt_objective} is closely linked to Delaunay triangulations: For a point set in 2D, the associated Delaunay triangulation minimizes $E_\text{ODT}$.
As we have seen in \cref{sec:objective_function}, both $E_\text{Geo}$ and $E_\text{ODT}$ can be written in terms of a norm $\|\vec{x}\|_M^2=\langle\vec{x},M\vec{x}\rangle$ with the respective choices for $M$. We can combine $E_\text{Geo}$ and $E_\text{ODT}$ into a single norm with $M=\vec{n}\cdot\vec{n}^t+\lambda\cdot\1$. In this way, our objective function is an extension of $E_\text{ODT}$ with a more general norm. 
Replacing $\|\cdot\|\rightarrow\|\cdot\|_M$ to generalize Delaunay triangulations in 2D was previously suggested~\cite{Chen:2004}. However, the presented algorithm is based on convex hulls and does not extend to surfaces in 3D or locally varying $M\in\R^{3\times 3}$.
In \cref{sec:edge_flips}, we will present a local version of this algorithm that extends the original idea to surfaces in 3D and with varying $M\in\R^{3\times 3}$. Furthermore, we show how this local version can be computed in screen space.
\subsection{Normal Integration}
\label{sec:normal_integration}
A unified description of the orthographic and perspective integration problem is achieved by minimising the functional
\begin{equation}
    E_\text{Int}=\int_\Omega\left\|\langle\vec{n},\vec{r}\rangle\cdot
    \begin{pmatrix}
        \partial_u z    \\  \partial_v z
    \end{pmatrix}
    +D\cdot
    \begin{pmatrix}
        n_x \\ n_y
    \end{pmatrix}
    \right\|^2\exterior^2u\,.
    \label{eqn:unified_integration}
\end{equation}
over the depth map $z:\Omega\rightarrow\R$~\cite{Queau:2018, Cao:2022}. For the orthographic projection, $\vec{r}=\vec{e}_z$ is the unit vector in the $z$-direction and $D=1$. For the perspective projection, $\vec{r}$ is the camera ray given by the intrinsics matrix and $D$ is the inverse of the focal length. Furthermore, we need to take the exponential of $z$ to obtain the final depth map in the perspective case.
\par
Only recently, a discretisation of \cref{eqn:unified_integration} for triangle meshes has been proposed~\cite{Heep:2025}. This mesh version of the normal integration problem resembles the seminal Cotan-Laplacian~\cite{Pinkall:1993} in geometry processing.
As for pixel-wise integration, it leads to a sparse linear system but replaces pixel neighbourhoods with adjacent vertices. Please refer to the supplementary material for further discussion of the nuances of mesh-based integration.
\begin{figure}
    \centering
    \includegraphics[trim={0 270 600 60}, clip, width=0.9\linewidth]{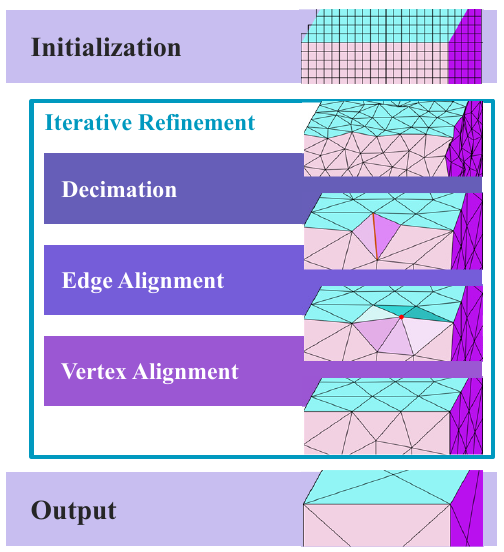}
    \caption{Schematic representation of our iterative screen space decimation: Decimation drastically reduces the number of vertices but may introduce misaligned edges and vertices which are fixed by our edge and vertex alignment procedures.}
    \label{fig:overview}
\end{figure}
\section{Algorithm} %
\label{sec:algorithm}
Our algorithm decimates an existing triangle mesh, ~\cref{fig:overview}, by iteratively repeating the following three steps: We collapse edges (\cref{sec:edge_collapse}) to remove redundant triangles in smooth, featureless regions, we update vertex positions (\cref{sec:tangential_smoothing}) and flip edges (\cref{sec:edge_flips}) to align better with ridges and furrows of the surface. The triangle mesh is initialized by converting each foreground pixel into two triangles.
\subsection{Quadrics from Normal Maps}
Calculating the screen space quadrics is a two-step procedure. First, we calculate face normals for each triangle by
\begin{equation}
    \vec{n}_f:=\normalize\left(\textstyle\sum_{p\in\Pixels_f}\vec{n}_p\right)
\end{equation}
where $\Pixels_f$ contains all pixels covered by triangle $f$ and $\vec{n}_p$ are the input pixel normals. From the face normals $\vec{n}_f$, we calculate the face Jacobians $J_f$ using \cref{eqn:jacobian_ortho,eqn:jacobian_persp} depending on the projection.
\par
Secondly, we calculate the quadrics
\begin{equation}
    Q_v(\delta\vec{x})=\sum_{f\in\Faces_v}\frac{A^{(3)}_f}{|\Pixels_f|}\sum_{p\in\Pixels_f}\left\|J_f\delta\vec{u}_{vp}+\delta\vec{x}\right\|_{M_p}^2,
    \label{eqn:discrete_integral}
\end{equation}
where $\delta\vec{u}_{vp}=\vec{u}_v-\vec{u}_p$ is the vector on screen pointing from the pixel to the vertex. The norm $\|\cdot\|_{M_p}$ at pixel $p$ is given through the matrix
\begin{equation}
\label{eqn:M_p_definition}
    M_p:=\vec{n}_p\cdot\vec{n}_p^t+\lambda\cdot\1, 
\end{equation}
where $\vec{n}_p$ the normal at pixel $p$. This matrix combines the contributions from $E_\text{Geo}$ and $E_\text{ODT}$.
The unforeshortened area $A^{(3)}_f=|J_f^tJ_f|^\frac{1}{2}\cdot A^{(2)}_f$ of $f$ can be calculated from the triangle area $A^{(2)}_f$ on screen by using the Jacobian.
\subsection{Vertex Alignment}
\label{sec:tangential_smoothing}
The key idea behind our vertex alignment algorithm is to place vertices along ridges and furrows of the surface to achieve a close representation even at low mesh resolutions. To translate a vertex displacement $\delta\vec{u}_v$ on screen into a vertex displacement $\delta\vec{x}_v$ in 3D space, we assume that the vertex is moving along the surface, \ie in the tangent space. We define vertex normals
\begin{equation}
    \vec{n}_v:=\normalize\left(\textstyle\sum_{f\in\Faces_v}A^{(3)}_f\cdot\vec{n}_f\right)\,,
\end{equation}
as the area-weighted average of the adjacent face normals and use \cref{eqn:jacobian_ortho,eqn:jacobian_persp} to calculate the Jacobian $J_v$ that characterizes the tangent space. Then, moving the vertex by $\delta\vec{u}_v$ on screen causes a displacement $\delta\vec{x}_v=J_v\delta\vec{u}_v$ in 3D space. To move the vertex to its optimal position, we minimise
\begin{equation}
    \label{eqn:screen_quadrics}
    \widetilde{Q}_v(\delta\vec{u}_v):=Q_v(J_v\delta\vec{u}_v)\,.
\end{equation}
Since quadrics are quadratic functions, the minimizer is found by solving a linear system. Finally, we apply the displacement
\begin{equation}
    \begin{aligned}
        \vec{u}_v&\leftarrow\vec{u}_v+\alpha\cdot\delta\vec{u}_v    &&\text{for }v\in\Vertices
    \end{aligned}
\end{equation}
scaled by a step width $\alpha=0.5$.
\begin{figure}
    \centering
    \includegraphics[width=.9\linewidth]{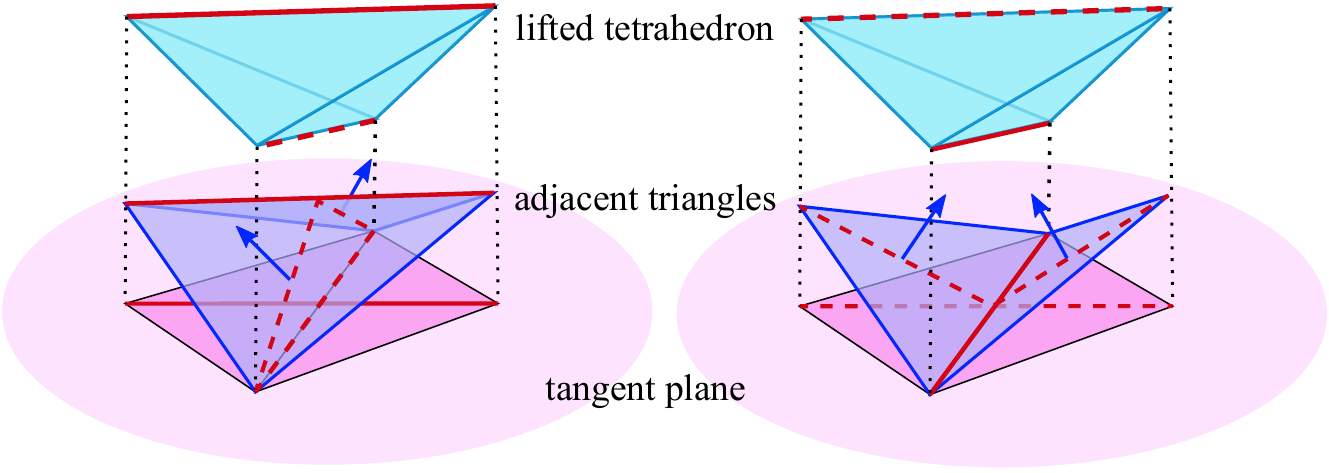}
    \caption{Schematic representation of our edge flip criterion: Each triangle patch is projected into its tangent plane. We then use $\|\cdot\|_{M_e}$ to lift the flattened patch. Among the two diagonals of the patch, the aligned edge is the lower edge in the resulting tetrahedron.}
    \label{fig:edge_flip_schematic}
\end{figure}
\subsection{Edge Alignment}
\label{sec:edge_flips}
Aside from placing vertices along ridges and furrows of the surface, it is equally important to align the triangle edges with these. We identify and flip non-aligned edges using the theory of generalized Delaunay triangulations in 2D, see \cref{sec:generalized_delaunay}.
For each non-boundary edge $e$, we consider the patch made of its two adjacent triangles $f,f'$. As for the vertices, we assign a normal
\begin{equation}
    \vec{n}_e:=\normalize\left(A_f^{(3)}\cdot\vec{n}_f+A^{(3)}_{f'}\cdot\vec{n}_{f'}\right)
\end{equation}
to the edge. Similarly, we approximate the quadric in the patch through
\begin{equation}
    M_e=\frac{A^{(3)}_f}{|\Pixels_f|}\sum_{p\in\Pixels_f}M_p+\frac{A^{(3)}_{f'}}{|\Pixels_{f'}|}\sum_{p\in\Pixels_{f'}}M_p\,.
\end{equation}
By projecting the four vertices forming the patch into the tangent plane, we obtain a flattened approximation of the patch. Similarly to the 2D generalized Delaunay triangulation, we use $\|\cdot\|_{M_e}$ to lift the vertices out of the plane to form a tetrahedron, see \cref{fig:edge_flip_schematic}. Different to the 2D version, our localized version is free of any expensive convex hull calculations. Instead, the current edge is geometry-aligned if it is \emph{lower} in the tetrahedron than the other diagonal of the two adjacent triangles. If this is not the case, we perform an edge flip.
\subsection{Decimation}
\label{sec:edge_collapse}
We remove vertices that least influence the object's overall surface through edge collapses~\cite{Garland:1997}.
During a collapse the edge $(v,w)$ between the vertices $v$ and $w$ is contracted into a single vertex. We judge the influence of a collapse on the overall shape by calculating a cost
\begin{equation}
    C_{vw}:=\min_{\vec{u}_{vw}}\widetilde{Q}_v(\vec{u}-\vec{u}_v)+\widetilde{Q}_w(\vec{u}-\vec{u}_w)\,,
    \label{eqn:collapse_cost}
\end{equation}
\ie the joint value of the screen-quadrics \cref{eqn:screen_quadrics} of the edge endpoints when contracted into an optimally positioned vertex at $\vec{u}_{vw}$. We restrict $\vec{u}_{vw}$ to the edge itself, \ie the solution is found by solving a 1D linear system.
\par
All collapsible edges are placed in a priority queue with non-decreasing $C_{vw}$. During a collapse, we contract the edge into the optimal position $\vec{u}_{vw}$. After the collapse, the quadric for this newly positioned vertex needs to be calculated and the optimal collapses for adjacent edges need to be updated. In line with previous work~\cite{Garland:1997}, the quadric at the new vertex is approximated as the sum of the two just removed vertex quadrics, instead of evaluating the computationally more involved \cref{eqn:discrete_integral}.
\subsection{Implementation Details}
Our decimation algorithm is written in \cpp\ and relies on \textsc{SurfaceMesh}~\cite{Sieger:2023} to represent triangle meshes. We employ the \textsc{nvdiffrast} renderer~\cite{Laine:2020} to translate between triangles and the pixel grid. All linear systems are solved using \textsc{Eigen}~\cite{Guennebaud:2010}. We present two ways to control the mesh resolution: Either by setting the vertex target or by providing a threshold on edge collapses, \cf \cref{eqn:collapse_cost}. In both cases, we perform five iterations of decimation each followed by edge- and vertex-alignment. If an error threshold is provided, each decimation step will perform all edge collapses with a cost below this threshold. If a vertex target is provided, we exponentially decrease the vertex count over five iterations starting at ten times the vertex target.
We use $\lambda=10^{-5}$ and $\alpha=0.5$ in all our experiments. Code is available at \url{https://moritzheep.github.io/anisotropic-screen-meshing}.
\begin{figure*}
    \centering \small
    \begin{tabular}{cccccccc}
        \multicolumn{3}{c}{RGBN~\cite{Toler-Franklin:2007} (uint8)} &
        \multicolumn{3}{c}{LUCES~\cite{Mecca:2021} (uint16)} &
        \multicolumn{2}{c}{PS~\cite{Frankot:1988} (float)} \\
        \includegraphics[height=3.1cm]{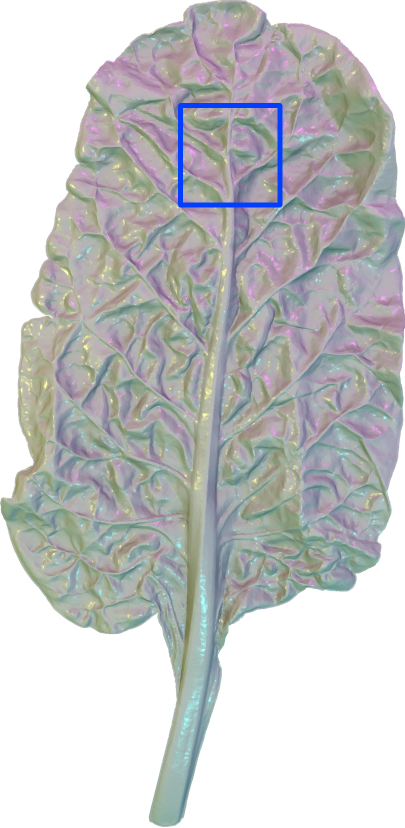}   &
        \includegraphics[height=3.1cm]{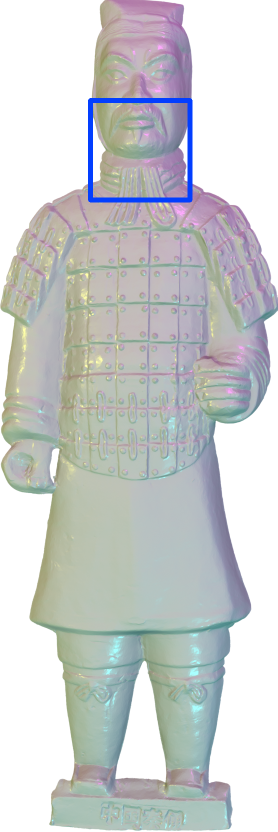}  &
        \includegraphics[height=3.1cm]{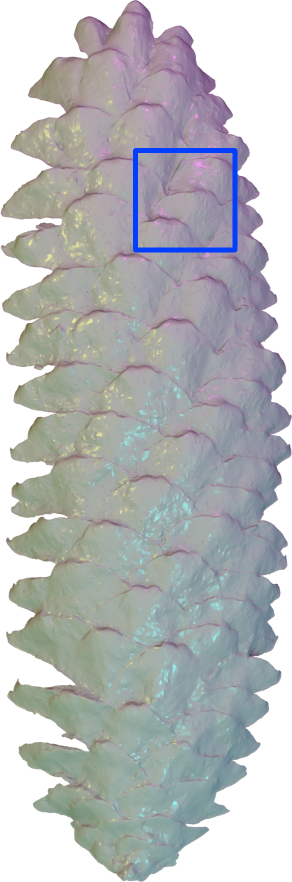}    &
        \includegraphics[height=3.1cm]{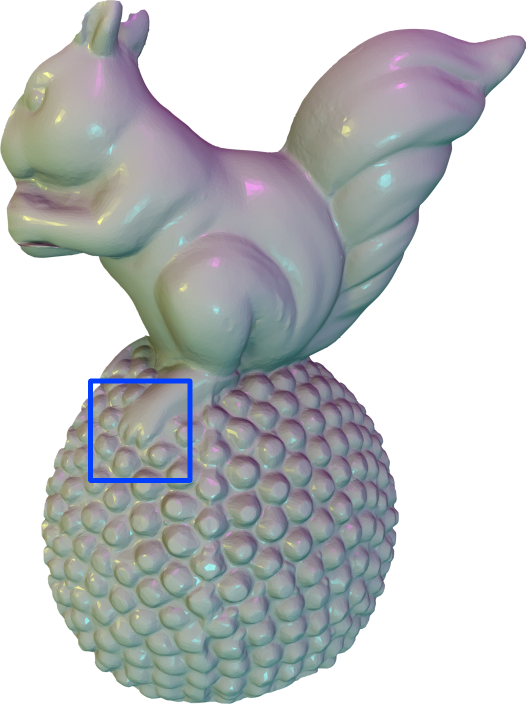} &
        \includegraphics[height=3.1cm]{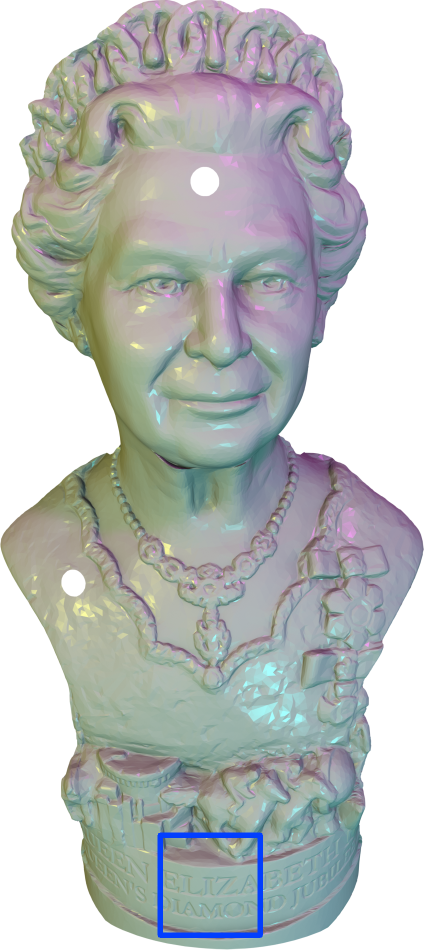}    &
        \includegraphics[height=3.1cm]{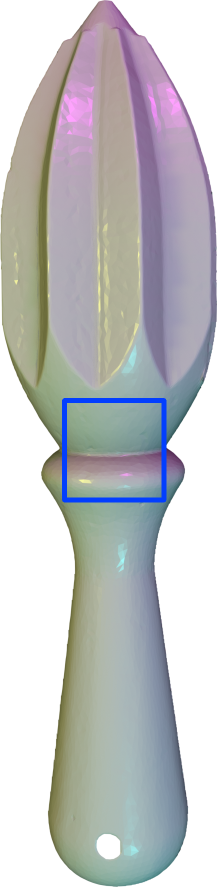}    &
        \shortstack{\includegraphics[height=1.7cm]{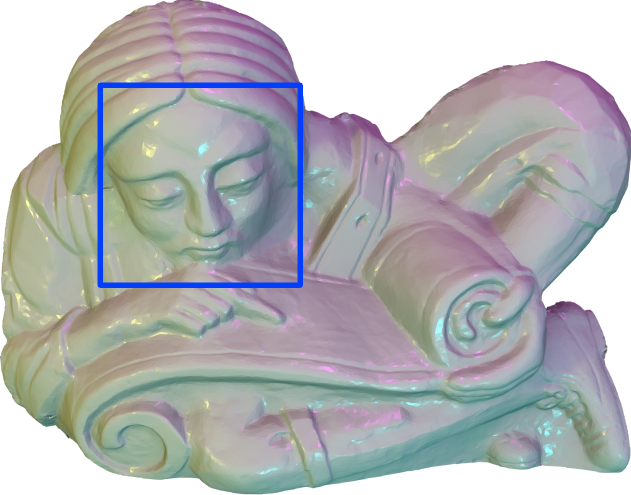}    \\
        \vspace{2mm}
        \includegraphics[height=1.3cm]{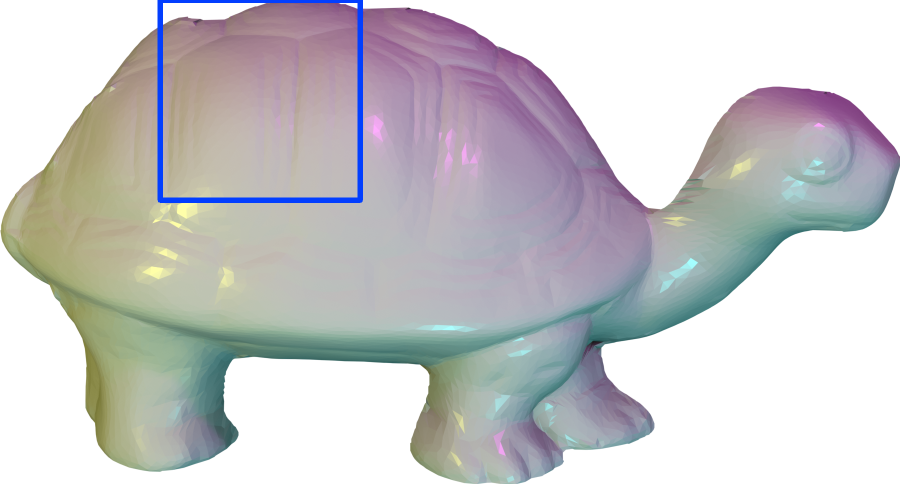}} &
        \includegraphics[height=3.1cm]{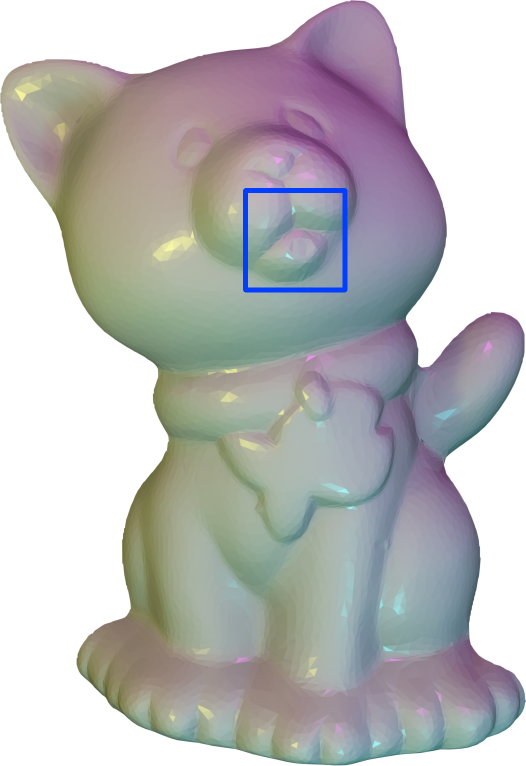}    \\
        \hline
        98.2\% &    98.5\%    & 98.06\% &  97.5\%  &   96.8\%   & 97.8\%  &   95.9\% / 96.9\%   &   96.8\%  \\
        \includegraphics[width=1.3cm]{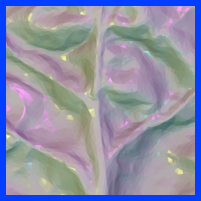}  &
        \includegraphics[width=1.3cm]{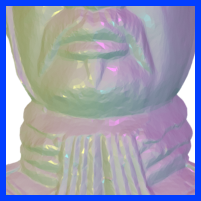}  &
        \includegraphics[width=1.3cm]{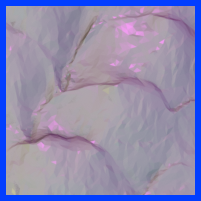}  &
        \includegraphics[width=1.3cm]{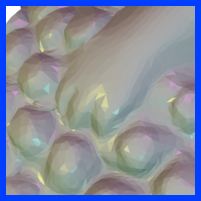}  &
        \includegraphics[width=1.3cm]{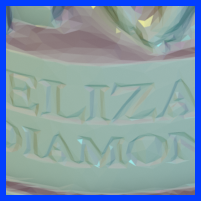}  &
        \includegraphics[width=1.3cm]{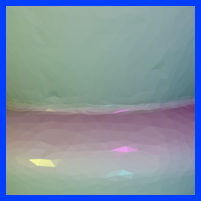}  &
        \includegraphics[width=1.3cm]{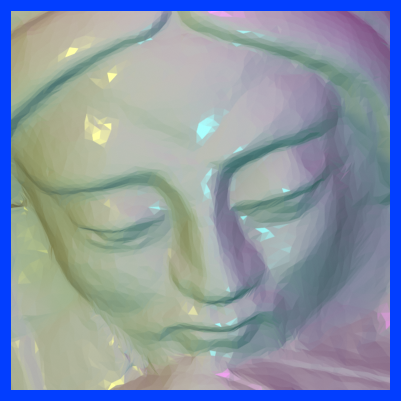}  
        \includegraphics[width=1.3cm]{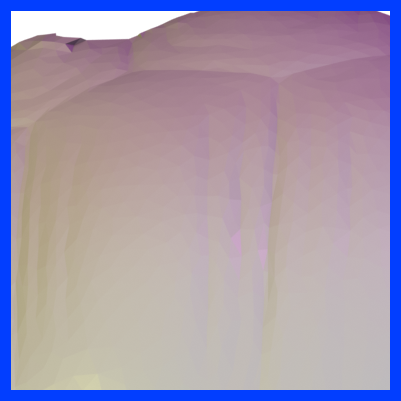} &
        \includegraphics[width=1.3cm]{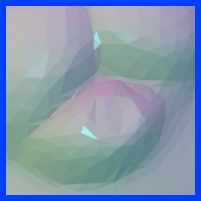}
    \end{tabular}
    \caption{Results from testing different datasets and normal map discretisations. 
    Even with compression rates beyond 95\%, fine details are well preserved and visible. The compression ratios refer to the number of depth variables and do not reflect file size. Any holes within objects are subject to the provided masks of the dataset.}
    \label{fig:additional_results}
\end{figure*}
\section{Evaluation}
We test our algorithm, quantitively and qualitatively on four publicly available datasets~\cite{Frankot:1988, Toler-Franklin:2007, Li:2020, Mecca:2021} and additional synthetical data. 
The DiLiGenT-MV~\cite{Li:2020} dataset contains 5 objects from 20 different views and LUCES~\cite{Mecca:2021} consists of 14 objects. We observe that the these datasets are either low resolution~\cite{Li:2020, Mecca:2021} or lack ground-truth geometry~\cite{Frankot:1988, Toler-Franklin:2007}. We mitigate this issue by complementing these published datasets with rendered normal maps. \Cref{fig:additional_results} illustrates some results on a range of datasets at varying discretisation accuracy: floating points~\cite{Frankot:1988,Li:2020}, 16-bit~\cite{Mecca:2021} or 8-bit unsigned integers~\cite{Toler-Franklin:2007}. 
Despite over 95\% fewer vertices that pixels in the input normal map, fine surface details remain clearly visible. Further results can be found in the supplementary material. 
\begin{figure}[b]
    \centering
    \begin{subfigure}[c]{.44\linewidth}
        \centering
        \small{\textbf{Isotropic}}
        \\
        \vspace{-2mm}
        \hrulefill\\
        \vspace{1mm}
        \begin{subfigure}{.48\linewidth}
            \includegraphics[width=\linewidth]{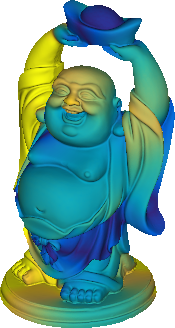}
            \caption*{5.94mm}
        \end{subfigure}
        \begin{subfigure}{.48\linewidth}
            \includegraphics[width=\linewidth]{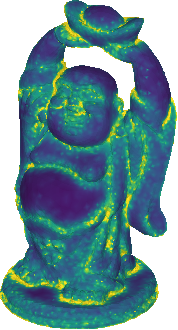}
            \caption*{Density}
        \end{subfigure}
    \end{subfigure}
    \begin{subfigure}[c]{.44\linewidth}
        \centering
        \small{\textbf{Ours}}
        \\
        \vspace{-2mm}
        \hrulefill\\
        \vspace{1mm}
        \begin{subfigure}{.48\linewidth}
            \includegraphics[width=\linewidth]{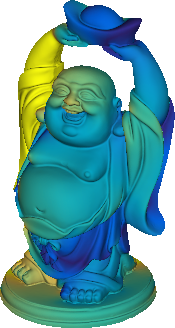}
            \caption*{5.24mm}
        \end{subfigure}
        \begin{subfigure}{.48\linewidth}
            \includegraphics[width=\linewidth]{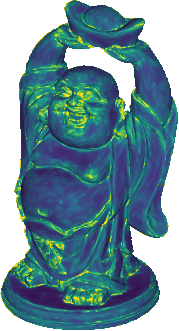}
            \caption*{Density}
        \end{subfigure}
    \end{subfigure}
    \begin{subfigure}[c]{.1\linewidth}
        \includegraphics[width=\linewidth]{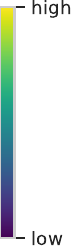}
    \end{subfigure}
    \caption{Absolute error and vertex density for isotropic remeshing (\emph{left}) and our decimation (\emph{right}). The isotropic algorithm places more vertices around ridges and furrows. Our algorithm places fewer vertices in these regions but still obtains a lower error due to the careful vertex and edge alignment.}
    \label{fig:vertex_density}
\end{figure}
\begin{table}[b]
    \centering
    \resizebox{\linewidth}{!}{
    \begin{tabular}{l|S[table-format=1.2]|S[table-format=1.2]S[table-format=1.2]S[table-format=1.2]|S[table-format=1.2]S[table-format=1.2]S[table-format=1.2]}
        \toprule
        &   \multicolumn{1}{c|}{\cite{Cao:2021}} 
        &   \multicolumn{3}{c|}{{Isotropic~\cite{Heep:2025}}}
        &   \multicolumn{3}{c}{Ours}    \\
        \midrule
        Dataset  &  \multicolumn{1}{c|}{Ref} & \multicolumn{1}{c}{low} & \multicolumn{1}{c}{mid} & \multicolumn{1}{c|}{high} & \multicolumn{1}{c}{low} & \multicolumn{1}{c}{mid} & \multicolumn{1}{c}{high}   \\
        \midrule
            \textsc{Bear}	    & 2.97 &	3.95 &	3.65 &	3.37 &	3.84 &	3.38 &	3.04  \\
            \textsc{Buddha}	    & 6.74 &	7.74 &	7.54 &	7.33 &	6.86 &	6.68 &	6.61  \\
            \textsc{Cow}	    & 2.45 &	3.42 &	3.12 &	2.96 &	3.07 &	2.85 &	2.74  \\
            \textsc{Pot2}	    & 5.15 &	5.89 &	5.77 &	5.65 &	5.63 &	5.47 &	5.29  \\
            \textsc{Reading}	& 6.34 &	7.08 &	6.93 &	6.83 &	6.82 &	6.67 &	6.50  \\
        \bottomrule
    \end{tabular}
    }
    \caption{Average RMSE over all 20 views of the DiLiGenT-MV dataset in mm using orthographic projection.}
    \label{tab:mesh_comparison}
\end{table}
\begin{figure*}
    \centering \small
    \begin{tblr}{colsep=1pt, colspec={b{.02\linewidth}b{.11\linewidth}b{.055\linewidth, rightsep=2pt}b{.11\linewidth}b{.055\linewidth, rightsep=2pt}b{.11\linewidth}b{.055\linewidth, rightsep=2pt}b{.11\linewidth}b{.055\linewidth, rightsep=4pt}b{.11\linewidth}b{.055\linewidth, rightsep=2pt}b{.064\linewidth}}, rowsep=1pt}
        &    \SetCell[c=2]{c}{\textsc{Bear}}    
        &&   \SetCell[c=2]{c}{\textsc{Buddha}}  
        &&   \SetCell[c=2]{c}{\textsc{Cow}}   
        &&   \SetCell[c=2]{c}{\textsc{Pot2}}   
        &&   \SetCell[c=2]{c}{\textsc{Reading}}    
        \\
        \cline{2-11}
        \hline[white]
        \SetCell[r=3]{l}{\rotatebox{90}{\textsc{Isotropic}}}
        &
        \includegraphics[width=\linewidth]{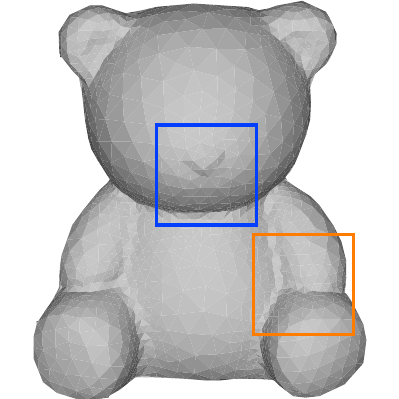}    &
        \includegraphics[width=\linewidth]{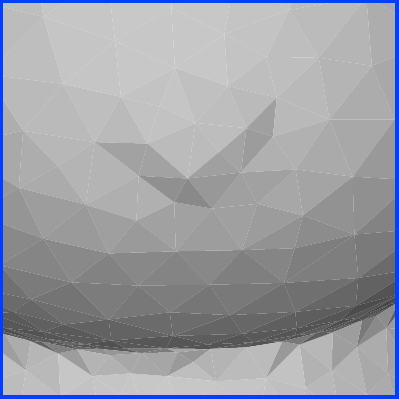} \newline
        \includegraphics[width=\linewidth]{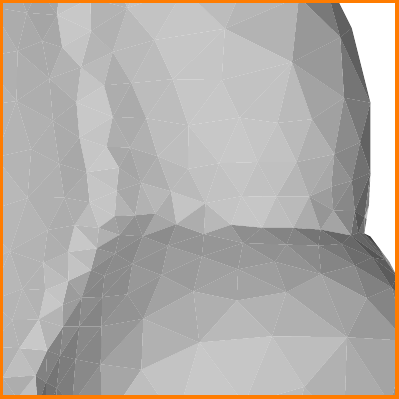} &
        \includegraphics[width=\linewidth]{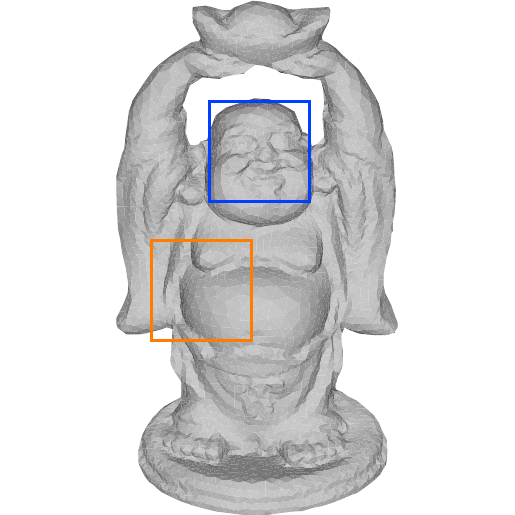}  &
        \includegraphics[width=\linewidth]{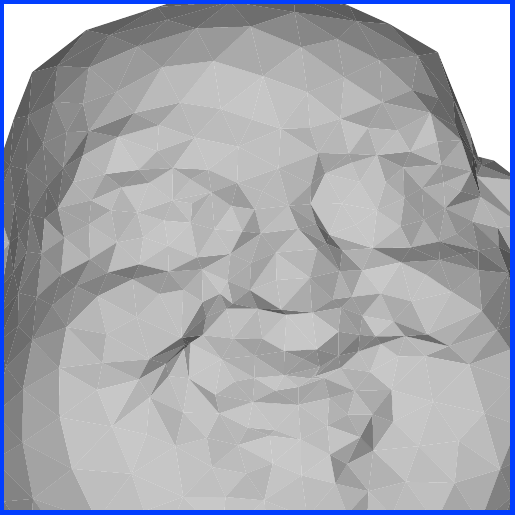} \newline
        \includegraphics[width=\linewidth]{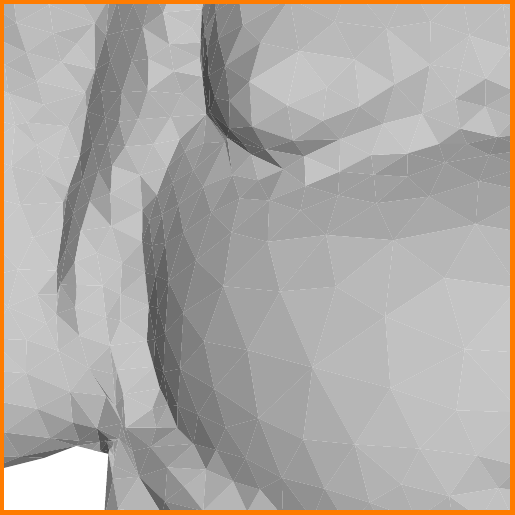} &
        \includegraphics[width=\linewidth]{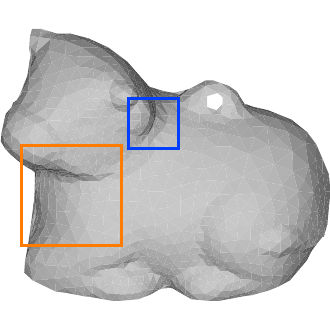} &
        \includegraphics[width=\linewidth]{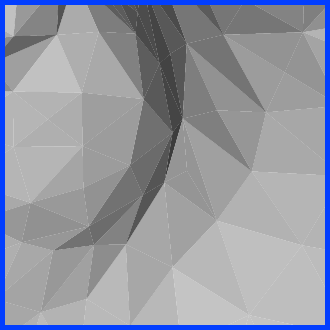} \newline
        \includegraphics[width=\linewidth]{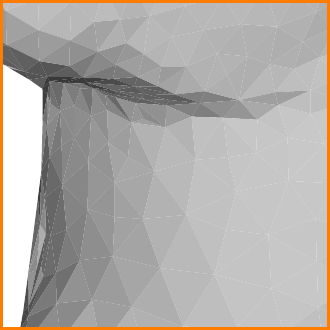} &
        \includegraphics[width=\linewidth]{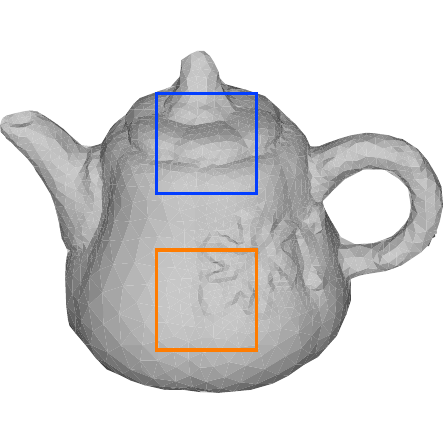}    &
        \includegraphics[width=\linewidth]{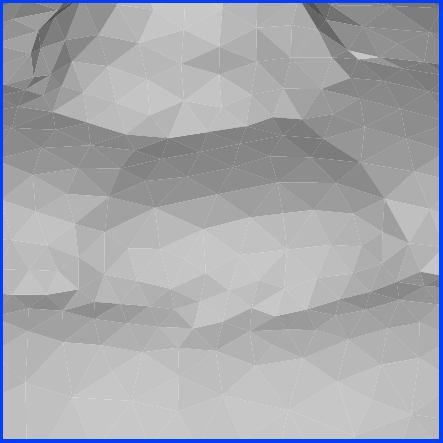} \newline
        \includegraphics[width=\linewidth]{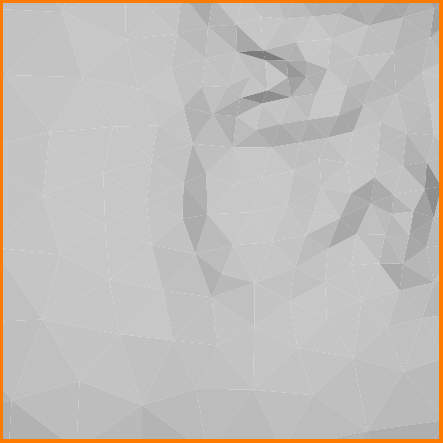} &
        \includegraphics[width=\linewidth]{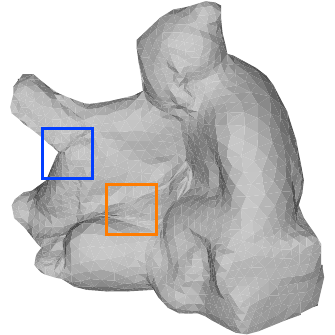} &
        \includegraphics[width=\linewidth]{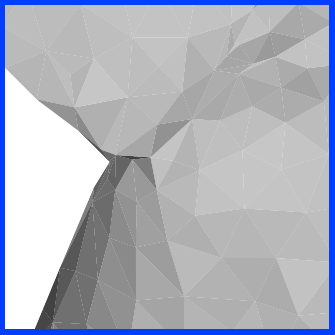} \newline
        \includegraphics[width=\linewidth]{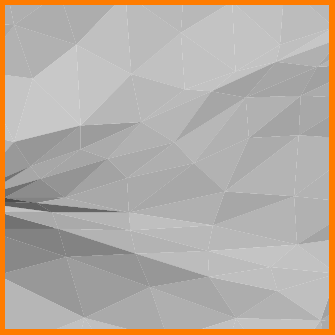} & 
        \SetCell[r=6]{c,m}{{\includegraphics[width=\linewidth]{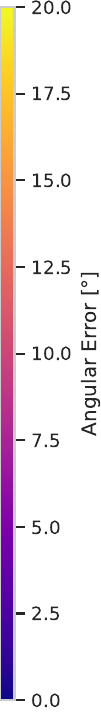}}} \\
        &\includegraphics[width=\linewidth]{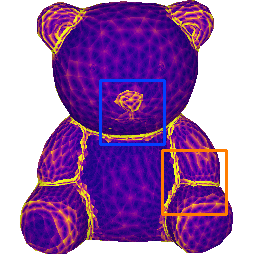}   &
        \includegraphics[width=\linewidth]{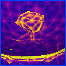} \newline
        \includegraphics[width=\linewidth]{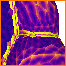} &
        \includegraphics[width=\linewidth]{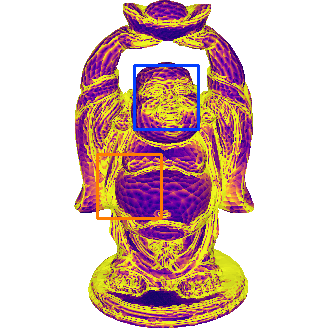}   &
        \includegraphics[width=\linewidth]{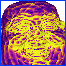} \newline
        \includegraphics[width=\linewidth]{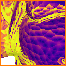} &
        \includegraphics[width=\linewidth]{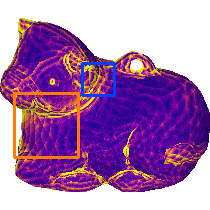}   &
        \includegraphics[width=\linewidth]{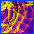} \newline
        \includegraphics[width=\linewidth]{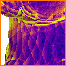} &
        \includegraphics[width=\linewidth]{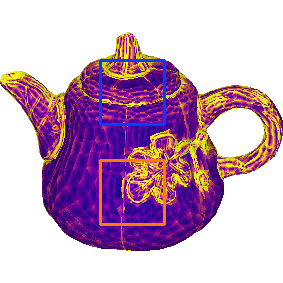}   &
        \includegraphics[width=\linewidth]{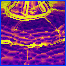} \newline
        \includegraphics[width=\linewidth]{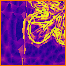} &
        \includegraphics[width=\linewidth]{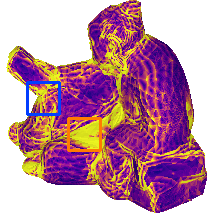}   &
        \includegraphics[width=\linewidth]{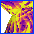} \newline
        \includegraphics[width=\linewidth]{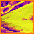} & \\
        &    \SetCell[c=2]{c}{$\mathbf{1.89\,\rm{mm}}$ / ${5.27^\circ}$}    
        &&   \SetCell[c=2]{c}{${5.19\,\rm{mm}}$ / ${15.81^\circ}$}  
        &&   \SetCell[c=2]{c}{$\mathbf{1.65\,\rm{mm}}$ / ${6.26^\circ}$}
        &&   \SetCell[c=2]{c}{${1.61\,\rm{mm}}$ / ${8.99^\circ}$}   
        &&   \SetCell[c=2]{c}{${9.46\,\rm{mm}}$ / ${11.10^\circ}$}   
        & \\
        \cline{2-11}
        \hline[white]
        \SetCell[r=3]{l}{\rotatebox{90}{\textsc{Ours}}}
        &
        \includegraphics[width=\linewidth]{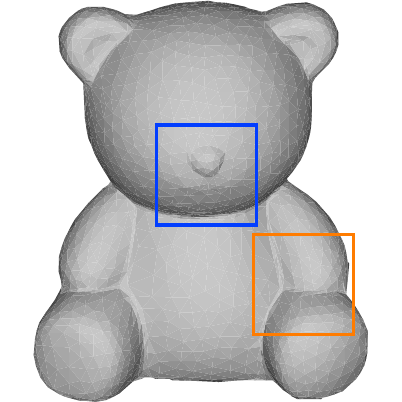}    &
        \includegraphics[width=\linewidth]{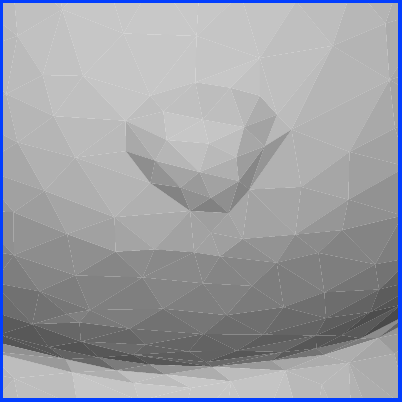} \newline
        \includegraphics[width=\linewidth]{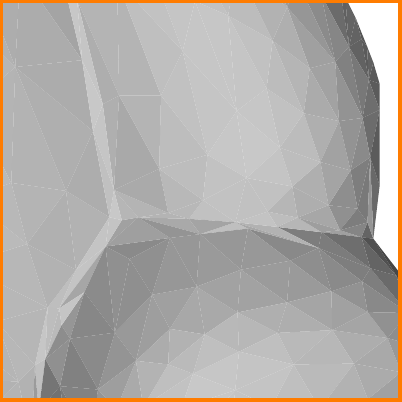} &
        \includegraphics[width=\linewidth]{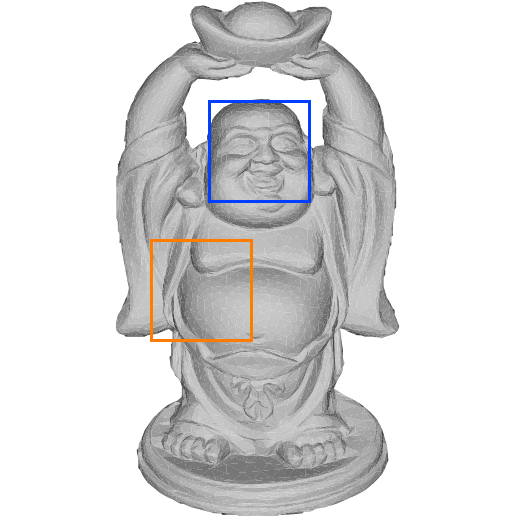}  &
        \includegraphics[width=\linewidth]{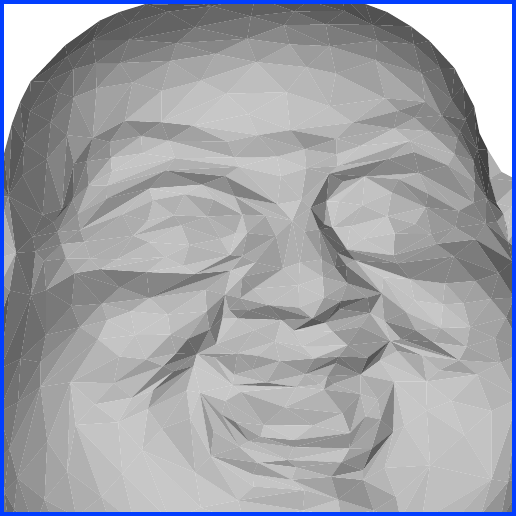} \newline
        \includegraphics[width=\linewidth]{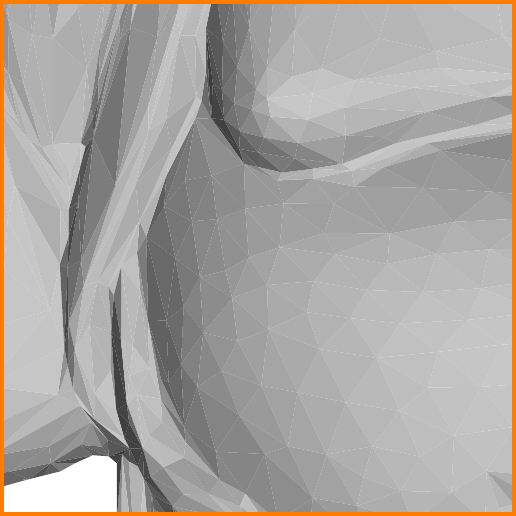} &
        \includegraphics[width=\linewidth]{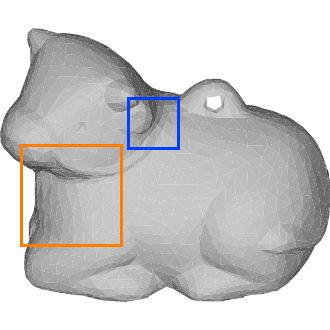} &
        \includegraphics[width=\linewidth]{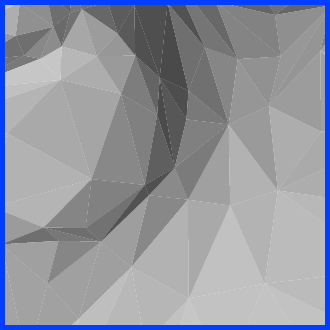} \newline
        \includegraphics[width=\linewidth]{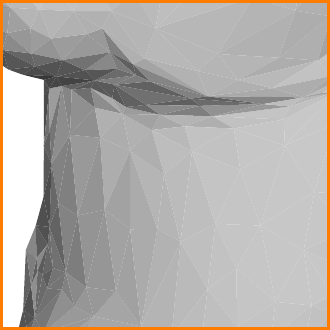} &
        \includegraphics[width=\linewidth]{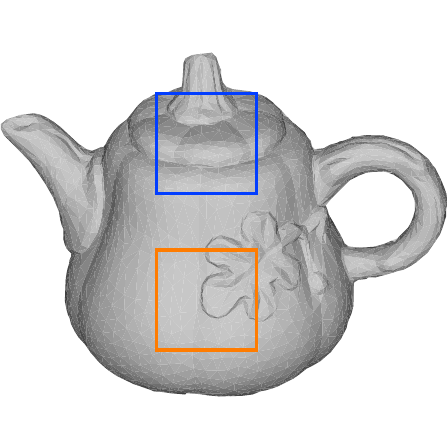}    &
        \includegraphics[width=\linewidth]{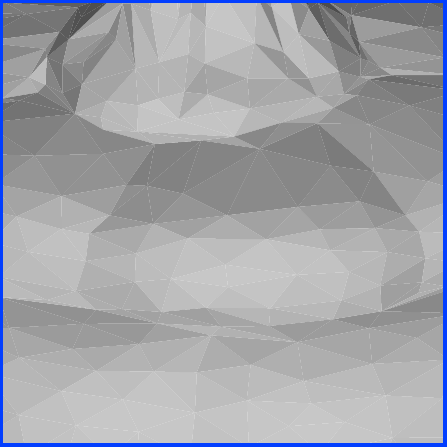} \newline
        \includegraphics[width=\linewidth]{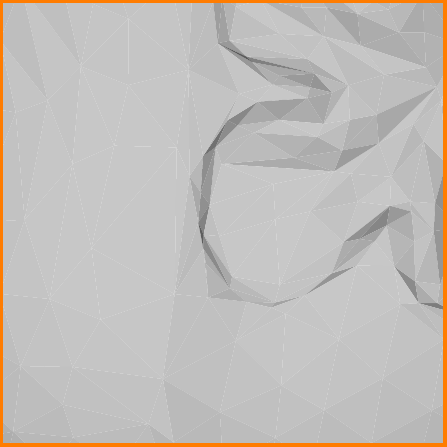} &
        \includegraphics[width=\linewidth]{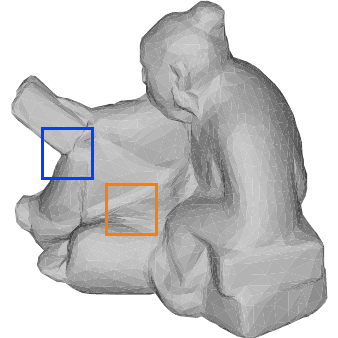} &
        \includegraphics[width=\linewidth]{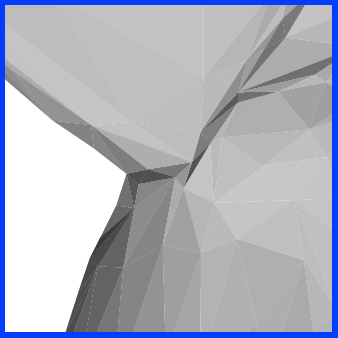} \newline
        \includegraphics[width=\linewidth]{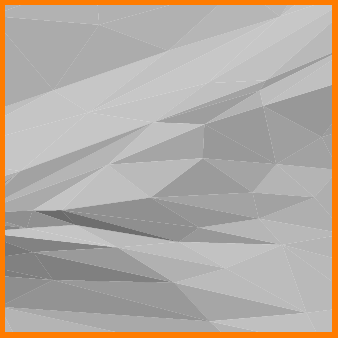} & \\
        &\includegraphics[width=\linewidth]{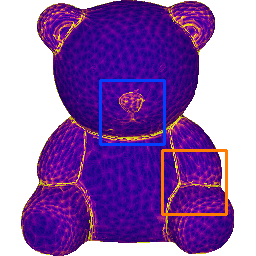}   &
        \includegraphics[width=\linewidth]{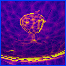} \newline
        \includegraphics[width=\linewidth]{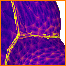} &
        \includegraphics[width=\linewidth]{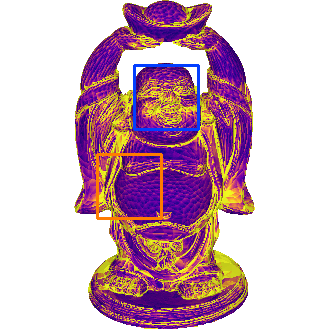}   &
        \includegraphics[width=\linewidth]{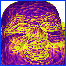} \newline
        \includegraphics[width=\linewidth]{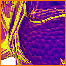} &
        \includegraphics[width=\linewidth]{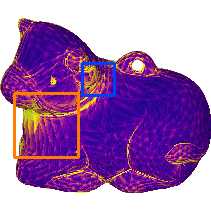}   &
        \includegraphics[width=\linewidth]{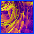} \newline
        \includegraphics[width=\linewidth]{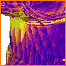} &
        \includegraphics[width=\linewidth]{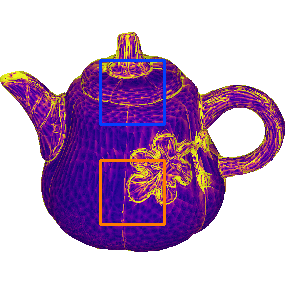}   &
        \includegraphics[width=\linewidth]{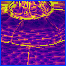} \newline
        \includegraphics[width=\linewidth]{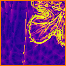} &
        \includegraphics[width=\linewidth]{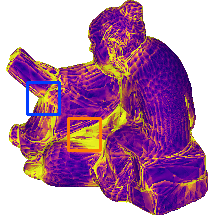}   &
        \includegraphics[width=\linewidth]{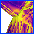} \newline
        \includegraphics[width=\linewidth]{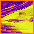} & \\
        &    \SetCell[c=2]{c}{$1.90\,\rm{mm}$ / $\mathbf{3.95^\circ}$}    
        &&   \SetCell[c=2]{c}{$\mathbf{5.00\,\rm{mm}}$ / $\mathbf{11.85^\circ}$}  
        &&   \SetCell[c=2]{c}{${1.76\,\rm{mm}}$ / $\mathbf{5.45^\circ}$}
        &&   \SetCell[c=2]{c}{$\mathbf{1.41\,\rm{mm}}$ / $\mathbf{6.99^\circ}$}   
        &&   \SetCell[c=2]{c}{$\mathbf{9.05\,\rm{mm}}$ / $\mathbf{9.43^\circ}$}
        & \\
        \cline{2-11}
        \hline[white]
        &    \SetCell[c=2]{c}{\normalsize{$1118$ Vertices}}    
        &&   \SetCell[c=2]{c}{\normalsize{$3758$ Vertices}}  
        &&   \SetCell[c=2]{c}{\normalsize{$782$ Vertices}}
        &&   \SetCell[c=2]{c}{\normalsize{$1565$ Vertices}}   
        &&   \SetCell[c=2]{c}{\normalsize{$1118$ Vertices}}
    \end{tblr}
    \caption{Comparison between isotropic~\cite{Heep:2025} (\emph{top}) and our anisotropic meshes (\emph{bottom}), both  in screen space for the DiLiGenT-MV dataset. We show wireframes (as vector graphics), angular error maps and report RMSE and MAE as error metrics. With the same number of vertices, we outperform both locally (MAE) and globally (RMSE) and preserves details better, \eg the bear's nose or the Buddha's face.}
    \label{fig:isotropic_vs_ours}
\end{figure*}
\subsection{Benchmark Comparisons}
As a first step, we compare our results to the only previous mesh-based integration method~\cite{Heep:2025} we are aware of. For an even comparison, we set our vertex target to match the numbers for the low-, mid- and high-resolution settings reported in~\cite{Heep:2025}.
\cref{tab:mesh_comparison} lists a summary of the root mean square error (RMSE) over all 20 views for DiLiGenT-MV. For reference, we also list the results of the pixel-based method in~\cite{Cao:2021} which performed best in our tests. The RMSE is always measured after a non-rigid alignment to absolve the inherent scale ambiguity in normal integration. Our method reliably outperforms the isotropic meshing and even gets within submillimetres of the pixel-based approach for the 'high' setting, where the number of vertices is typically still 10 times smaller than the number of pixels. \Cref{fig:isotropic_vs_ours} displays the visual results at 'mid' resolution together with the RMSE and mean angular error (MAE), which confirm the advantages of our method. Comparing the vertex density images of both methods in \cref{fig:vertex_density}, we identify that our method concentrates vertices directly at ridges and furrows, while the isotropic method~\cite{Heep:2025} creates a much more smoothly varying vertex density. The isotropic approach trades isotropy for accuracy. 
\subsection{Controllability}
For a second evaluation, we use the 14 objects in the LUCES dataset and control the mesh resolution by providing three decimation thresholds, \cf \cref{eqn:collapse_cost}. The results are listed in \cref{tab:luces_comparison}. LUCES is a challenging dataset as objects are placed very close to the camera. Nonetheless, a lower decimation threshold correlates with a lower reconstruction error. As for DiLiGenT-MV, the highest mesh resolution is within the submillimetre range of the pixel-based baseline~\cite{Heep:2022}. Additional results are in the supplementary material.
\subsection{Ablation Studies}
\begin{figure}
    \centering
    \begin{subfigure}[b]{.475\linewidth}
        \centering
        \small{\textbf{Decimation Only}}\\
        \begin{subfigure}[c]{.7\linewidth}
            \includegraphics[width=\linewidth]{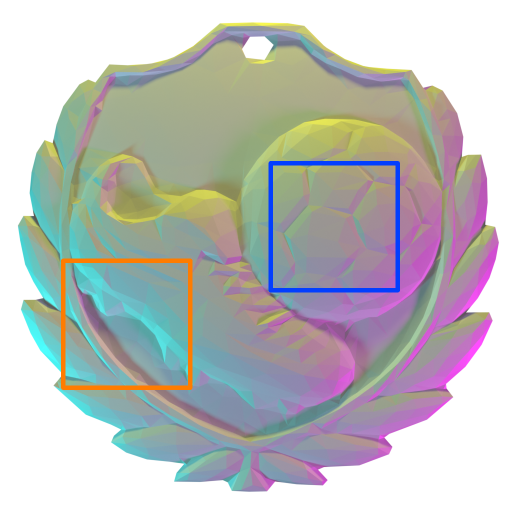}
        \end{subfigure}
        \begin{subfigure}[c]{.25\linewidth}
            \includegraphics[width=\linewidth]{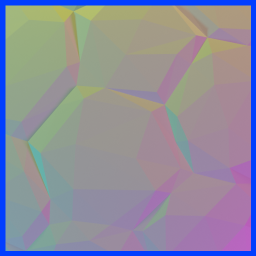}    \\
            \includegraphics[width=\linewidth]{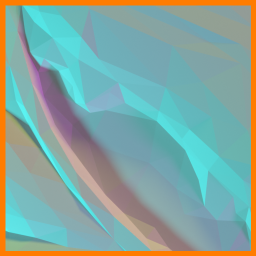}
        \end{subfigure}
        \\
        \hrulefill
        \\
        $1.64$ / $6.18^\circ$
    \end{subfigure}
    \begin{subfigure}[b]{.475\linewidth}
        \centering
        \small{\textbf{Dec + Vertex Alignment}}\\
        \begin{subfigure}[c]{.7\linewidth}
            \includegraphics[width=\linewidth]{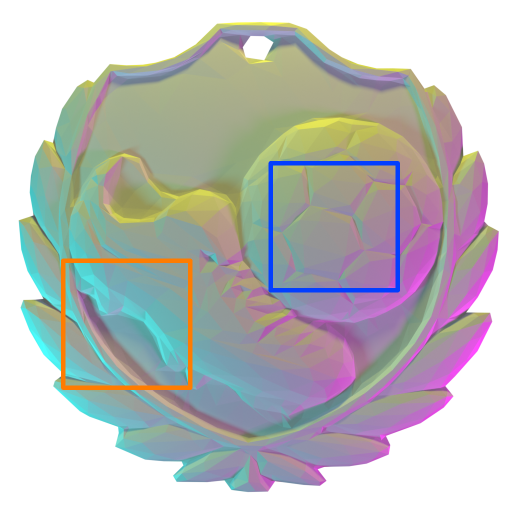}
        \end{subfigure}
        \begin{subfigure}[c]{.25\linewidth}
            \includegraphics[width=\linewidth]{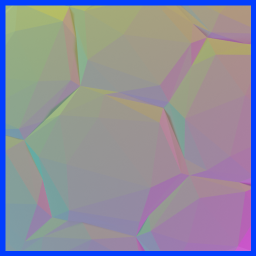}    \\
            \includegraphics[width=\linewidth]{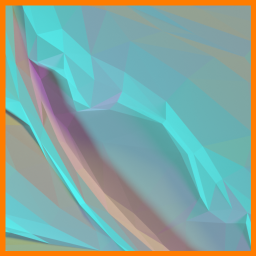}
        \end{subfigure}
        \\
        \hrulefill
        \\
        $1.59$ / $6.03^\circ$
    \end{subfigure}
    \\
    \vspace{1.5mm}
    \begin{subfigure}[b]{.475\linewidth}
        \centering
        \small{\textbf{Dec + Edge Alignment}}\\
        \begin{subfigure}[c]{.7\linewidth}
            \includegraphics[width=\linewidth]{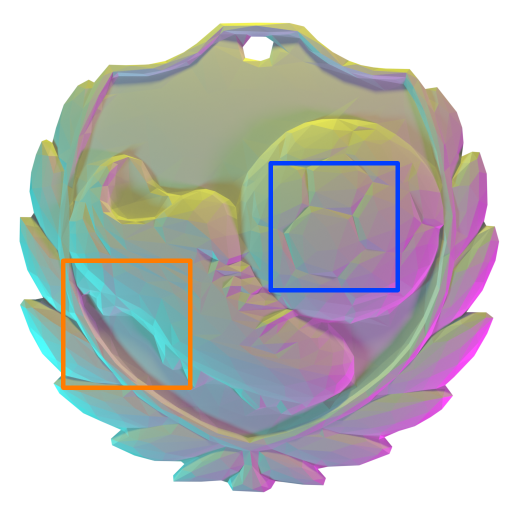}
        \end{subfigure}
        \begin{subfigure}[c]{.25\linewidth}
            \includegraphics[width=\linewidth]{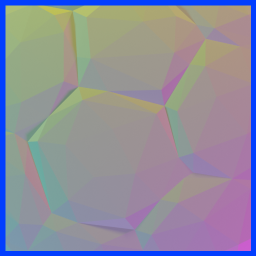}    \\
            \includegraphics[width=\linewidth]{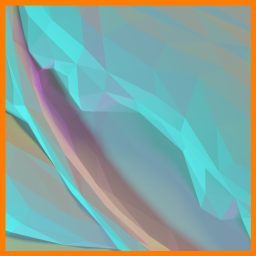}
        \end{subfigure}
        \\
        \hrulefill
        \\
        $1.72$ / $5.90^\circ$
    \end{subfigure}
    \begin{subfigure}[b]{.475\linewidth}
        \centering
        \small{\textbf{Ours}}\\
        \begin{subfigure}[c]{.7\linewidth}
            \includegraphics[width=\linewidth]{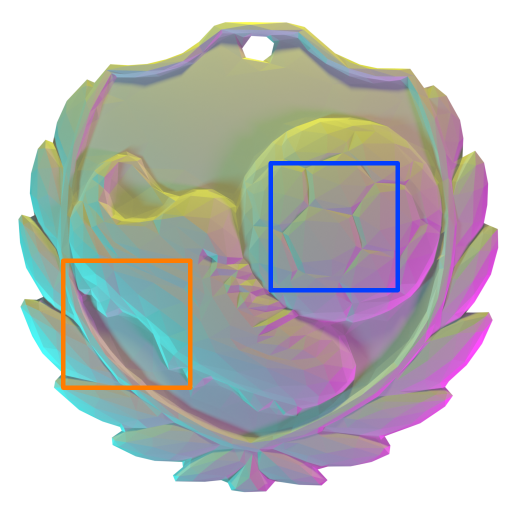}
        \end{subfigure}
        \begin{subfigure}[c]{.25\linewidth}
            \includegraphics[width=\linewidth]{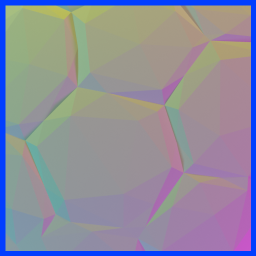}    \\
            \includegraphics[width=\linewidth]{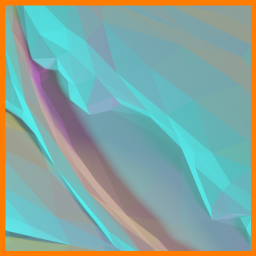}
        \end{subfigure}
        \\
        \hrulefill
        \\
        $\mathbf{1.55}$ / $\mathbf{5.74^\circ}$
    \end{subfigure}
    \caption{Ablation study of the three components of our method: Results for a vertex resolution of 1k using only decimation, decimation and vertex alignment, decimation and edge alignment as well as our full algorithm.}
    \label{fig:ablation_parts}
\end{figure}

For an ablation study, we consider the three operations of our algorithm, see \cref{fig:ablation_parts}: Decimation alone, decimation with either vertex alignment or edge alignment as well as the full algorithm. Decimation alone preserves the rough shape but struggles with high-frequency details. Adding either vertex- or edge-alignment improves RMSE and MAE respectively but only the full algorithm leads to the best results both qualitatively and quantitatively.
\begin{table}[b]
    \centering
    \begin{tabular}{l|S[table-format=1.2]|S[table-format=1.2]S[table-format=1.2]S[table-format=1.2]}
        \toprule
        \multirow{2}{*}{Dataset}    &   \multicolumn{1}{c}{Ref} &   \multicolumn{3}{c}{Threshold}    \\
        \cmidrule{2-5}              &   \cite{Heep:2022}
                                    &   \multicolumn{1}{c}{$2$}
                                    &   \multicolumn{1}{c}{$2^6$}
                                    &   \multicolumn{1}{c}{$2^{11}$} \\
        \midrule
        \textsc{Bell} & 0.30 & 0.33 & 0.49 & 0.72 \\ %
        \textsc{Buddha} & 3.46 & 3.58 & 3.51 & 3.32 \\ %
        \textsc{Bunny} & 3.38 & 3.67 & 3.83 & 4.15 \\ %
        \textsc{Die} & 1.62 & 1.65 & 2.11 & 3.09 \\ %
        \textsc{Hippo} & 2.73 & 2.85 & 2.91 & 3.09 \\ %
        \textsc{Jar} & 0.50 & 0.45 & 0.45 & 0.46 \\ %
        \textsc{Owl} & 4.89 & 5.41 & 5.63 & 5.94 \\ %
        \textsc{Queen} & 3.74 & 4.02 & 4.41 & 4.98 \\ %
        \textsc{Squirrel} & 1.91 & 2.07 & 2.37 & 3.00 \\ %
        \textsc{Tool} & 0.91 & 0.98 & 0.98 & 1.04 \\ %
        \bottomrule
    \end{tabular}
    \caption{RMSE in mm for some of the objects in LUCES~\cite{Mecca:2021} with increasing decimation threshold, see \cref{eqn:collapse_cost}. The chosen thresholds should lead to a constant reduction rate of the RMSE.}
    \label{tab:luces_comparison}
\end{table}
\subsection{Runtime} 
\begin{figure}
    \centering
    \includegraphics[width=\linewidth]{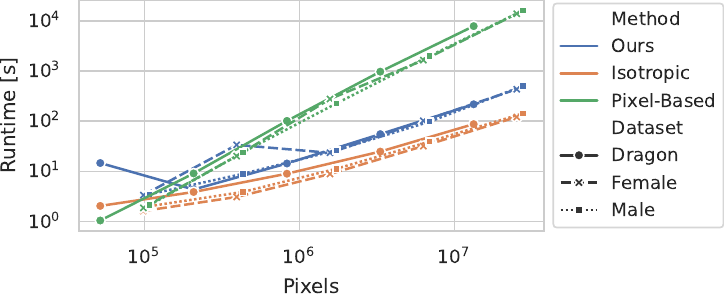}
    \\
    \vspace{-3mm}
    \caption{Total runtime (meshing and surface integration) as a function of the number of foreground pixels. For the pixel-based method it is only integration time. Please note the log-log scale.}
    \label{fig:runtime_graph}
\end{figure}

To evaluate the runtime at higher resolutions, we created synthetic normals maps from high-resolution 3D scans with resolutions ranging from $512^2$ to $8192^2$ and applied the method in~\cite{Heep:2025} and our method with a constant quality setting. Compared to the simpler isotropic remeshing~\cite{Heep:2025}, the improved accuracy of our anisotropic approach has the downside of slightly higher runtimes. Nevertheless, our method is still considerably faster than pixel-wise integration, especially for increasing resolutions. 
According to \cref{fig:runtime_graph}, the overhead of converting pixels to meshes \emph{before} the integration already pays off at resolutions as low as 1MP. At resolutions beyond 10MP, mesh-based methods are up to 100 times faster than traditional pixel-based integration.
\subsection{Robustness}
\begin{figure}
    \centering \small
    \begin{subfigure}[t]{.30\linewidth}
        \centering
        \normalsize{${\sigma=0^\circ}$}
        \includegraphics[trim={11 11 22 8}, clip, width=\linewidth]{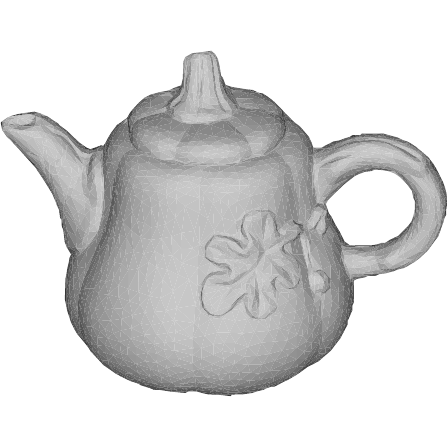}
        $1.41\,\text{mm}$ / $6.56^\circ$ 
    \end{subfigure}
    \hspace{.02\linewidth}
    \begin{subfigure}[t]{.30\linewidth}
        \centering
        \normalsize{${\sigma=3^\circ}$}
        \includegraphics[trim={11 11 22 8}, clip, width=\linewidth]{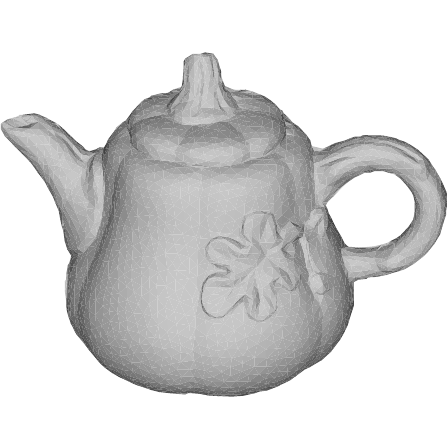}
        $1.57\,\text{mm}$ / $7.06^\circ$ 
    \end{subfigure}
    \hspace{.02\linewidth}
    \begin{subfigure}[t]{.30\linewidth}
        \centering
        \normalsize{${\sigma=10^\circ}$}
        \includegraphics[trim={11 11 22 8}, clip, width=\linewidth]{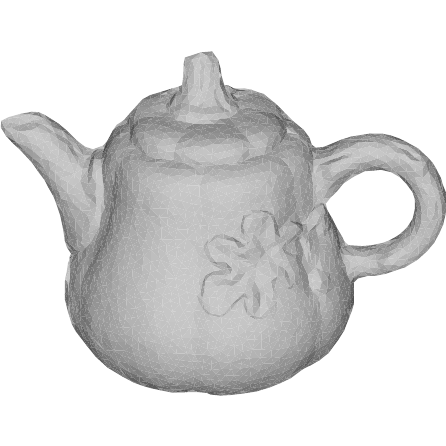}
        $2.28\,\text{mm}$ / $9.04^\circ$ 
    \end{subfigure}
    \caption{Robustness against increasing levels of normally distributed noise: Numbers are RMSE and MAE respectively. All meshes have a 2k vertices and are vector graphics to allow close examination in the digital version.}
    \label{fig:robustness_noise}
\end{figure}

Given that noise might be present in experimentally obtained normal maps, we tested the robustness of our method towards imperfect inputs by distorting the ground-truth normal maps with Gaussian noise of various amplitudes, see \cref{fig:robustness_noise}. As expected, the quality of the input is reflected in the quality of the output. However, our vertex and edge alignment to ridges and furrows remains consistent even at high levels of noise. 
\section{Conclusion and Limitations} %
We proposed a screen space decimation approach for mesh-based integration. Our results show that careful alignment of vertices and edges to ridges and furrows of the underlying surface is key to surpassing the quality of previous methods and maintaining high geometric faithfulness even at high compression ratios. Conversely, we achieve comparable results to pixel-based methods at moderate compression ratios. Unlike previous approaches, our method offers greater control over the output. This can be achieved by either setting an error threshold to reach a desired quality or by defining a fixed number of vertices to stay within limitations like GPU memory. All things considered, we presented a versatile approach that allows balancing runtime and quality and can be adjusted to the needs of almost any photometric stereo pipeline. 
\par
As with all single-view reconstruction approaches, normal integration suffers from an inherent scale ambiguity for the final geometry. Combining our approach with multi-view reconstruction could overcome this limitation and is an interesting direction for future research. Furthermore, our method is currently designed for continuous surfaces. The major challenge in introducing discontinuities into mesh-based integration so far has been aligning edges with discontinuities. It seems that our alignment strategies to geometric details are a stepping stone in this direction.
\paragraph{Acknowledgements}
The "David Head" by \href{https://skfb.ly/oKvzW}{1d\_inc} and the "Football Medal2 - PhotoCatch" by \href{https://skfb.ly/oGXrA}{Moshe Caine} were licensed under CC BY 4.0.
This work has been funded by the Deutsche Forschungsgemeinschaft (DFG, German Research Foundation) under Germany's Excellence Strategy, EXC-2070 -- 390732324 (PhenoRob).
\FloatBarrier
\clearpage
{
    \small
    \bibliographystyle{ieeenat_fullname}
    \bibliography{cvpr2025}
}
\end{document}


\maketitle
\section*{Overview}
This supplementary material, provides additional technical details on the normal integration, the final algorithm and additional evaluation results that have been removed from the main document due to page constraints. For easier navigation and cross-referencing, we follow the section titles of the submitted paper.
For a quick overview, we briefly list the content of the Appendix:
\begin{itemize}
    \item \cref{sec:integration_details} provides additional details on the integration, improving further the surface reconstruction accuracy compared to previous work\cite{Heep:2025}.
    \item \cref{sec:quadrics_details} describes additional details on the computation of quadrics and the linear system to solve. 
    \item \cref{sec:evaluation_details} provides additional comparisons to previous work, quantitative analysis on the reconstruction accuracy and qualitative examples.
    \item \cref{sec:datasets} is a list of all the datasets we used together with the urls to find them. 
\end{itemize}
Finally, we included samples of the reconstructed surfaces using our method (\texttt{ours.obj}) and using \cite{Heep:2025} (\texttt{adaptive.obj}) for
\begin{itemize}
    \item Michelangelo's David (Fig.\ 1 of the manuscript) and
    \item all five objects in the DiLiGenT-MV \cite{Li:2020} dataset (reconstructed from the first view each).
\end{itemize}
These samples can be found in the respectives subdirectories of \texttt{meshes/} of this supplementary material.
\par
A reference implementation of our method is available under \url{https://moritzheep.github.io/anisotropic-screen-meshing}.
\appendix
\section{Screen Space Mesh Decimation}\label{sec:integration_details}
In the following, we will provide more detail on our mesh-based integration and the underlying sparse linear system, that slightly differs from \cite{Heep:2025} but improves the reconstruction accuracy even further.
\subsection{Details on Mesh-Based Normal Integration}
\begin{table*}
    \centering
    \begin{tabular}{l|l|S[table-format=1.2]|S[table-format=1.2]S[table-format=1.2]S[table-format=1.2]|S[table-format=1.2]S[table-format=1.2]S[table-format=1.2]|S[table-format=1.2]S[table-format=1.2]S[table-format=1.2]}
        \toprule
        & 
        &&&&& \multicolumn{3}{c|}{Our Decimation,}  \\
        &&  \cite{Cao:2021} 
        &   \multicolumn{3}{c|}{Isotropic \cite{Heep:2025}}   
        &   \multicolumn{3}{c|}{\cite{Heep:2025}'s Integration}
        &   \multicolumn{3}{c}{Ours}    \\
        \midrule
        &   Dataset &   {Ref}   &   {low}   &   {mid}   &   {high}  &   {low}   &   {mid}   &   {high}  &   {low}   &   {mid}   &   {high}  \\ 
        \midrule
        \multirow{5}{*}{\rotatebox{90}{Orthographic}}
        &   Bear    &   2.97    &   3.95    &   3.65    &   3.37    &   3.67    &   3.33    &   3.19    &   3.84    &   3.38    &   3.04    \\
        &   Buddha  &   6.74    &   7.74    &   7.54    &   7.33    &   7.30    &   7.10    &   7.08    &   6.86    &   6.68    &   6.61    \\
        &   Cow     &   2.45    &   3.42    &   3.12    &   2.96    &   3.23    &   3.00    &   2.86    &   3.07    &   2.85    &   2.74    \\
        &   Pot2    &   5.15    &   5.89    &   5.77    &   5.65    &   5.72    &   5.59    &   5.48    &   5.63    &   5.47    &   5.29    \\
        &   Reading &   6.34    &   7.08    &   6.93    &   6.83    &   6.88    &   6.76    &   6.64    &   6.82    &   6.67    &   6.50    \\
        \midrule
        \multirow{5}{*}{\rotatebox{90}{Perspective}}
        &   Bear    &   2.91    &   3.94    &   3.72    &   3.47    &   3.64    &   3.48    &   3.33    &   3.69    &   3.22    &   2.90    \\
        &   Buddha  &   6.75    &   7.74    &   7.53    &   7.40    &   7.31    &   7.13    &   7.09    &   6.83    &   6.68    &   6.62    \\
        &   Cow     &   2.35    &   3.49    &   3.24    &   3.07    &   3.29    &   3.09    &   2.99    &   2.97    &   2.77    &   2.63    \\
        &   Pot2    &   4.99    &   6.04    &   5.86    &   5.76    &   5.81    &   5.69    &   5.61    &   5.48    &   5.32    &   5.16    \\
        &   Reading &   6.28    &   7.19    &   6.94    &   6.85    &   6.91    &   6.77    &   6.69    &   6.74    &   6.52    &   6.45    \\
        \bottomrule
    \end{tabular}
    \caption{Comparison of the average RMSE over all 20 views of DiLiGenT-MV: Isotropic remeshing and integration from \cite{Heep:2025}, our decimation combined with the integrator of \cite{Heep:2025} and our decimation with our integrator. Our finer approximation of the integration parameters yields to a tighter approximation of the underlying surface.}
    \label{tab:old_vs_new_integration}
\end{table*}
The unified functional for normal integration is
\begin{align}
    E_\text{Int}=\int_\Omega\left\|\langle\vec{n},\vec{r}\rangle\cdot
    \begin{pmatrix}
        \partial_u z    \\  \partial_v z
    \end{pmatrix}
    +D\cdot
    \begin{pmatrix}
        n_x \\ n_y
    \end{pmatrix}
    \right\|^2\exterior^2u\,.
    \label{eqn:unified_integration}
\end{align}
with the respective choices for $\vec{r}$ and $D$ depending on the projection type -- orthographic or perspective. In the case of pixel-based integration, the partial derivatives are discretised by using each pixel and its immediate neighbourhood. In the case of mesh-based integration, the pixel neighbourhood is replaced by adjacent vertices. 
\par
Previous work \cite{Heep:2025} showed that the minimiser of $E_\text{Int}$ is found by solving the sparse linear system
\begin{align}
    \sum_{w\in\Vertices_v}\sum_{f\in\Faces_{(v,w)}}\omega_{f,vw}\cdot(m_f\cdot\delta z_{vw}+\langle\vec{b}_f,\delta\vec{u}_{vw}\rangle)=0
\end{align}
for each vertex $v\in\Vertices$. All vertices $w$ that are connected by an edge with vertex $v$ have an impact on depth $z_v$. We denote these vertices by $\Vertices_v$. The faces adjacent to the edge $(v,w)$, denoted as $\Faces_{u,v}$, define the strength of the influence of $w$. The edge weights $\omega_{f,vw}:=\cot(\alpha_{f,vw})$ are given by the cotangent of the angle in $f$ that is opposite to $(v,w)$. The weights are identical to those of the Cotan-Laplacian \cite{Pinkall:1993}. Together with the two parameters \cite{Heep:2025}
\begin{align}
    \label{eqn:integration_parameters}
    m_f 	   &= \int_f\langle\vec{r},\vec{n}\rangle^2\exterior\Omega \\
    \vec{b}_f  &= D\cdot\int_f\langle\vec{r},\vec{n}\rangle\cdot\begin{pmatrix}
        n_x \\ n_y
    \end{pmatrix}\exterior\Omega
\end{align}
they ensure that the integration yields the same results for different triangulations. Previously, these parameters have been calculated by assuming a constant face normal $\vec{n}_f$. In contrast, we found that 
\begin{align}
    m_f 	   &= \frac{1}{|\Pixels_f|}\sum_{p\in\Pixels_f}\langle\vec{r}_p,\vec{n}_p\rangle^2 \\
    \vec{b}_f  &= \frac{D}{|\Pixels_f|}\langle\vec{r}_p,\vec{n}_p\rangle\cdot
    \begin{pmatrix}
        n_x \\ n_y
    \end{pmatrix}_p
\end{align}
is a tighter approximation that considers variations of the normals within the faces and yields more accurate surface integrations, see \cref{tab:old_vs_new_integration}.
\section{Algorithm}\label{sec:quadrics_details}
In this section, we want to express the quadrics in a more familiar form of a quadratic function and derive the linear system to solve during vertex alignment.
\subsection{Explicit Form of the Quadrics}
In Eq.\ (9) of the main paper, we defined the quadric for the continuous case as follows:  
\begin{align}
     Q_v(\delta\vec{x}_v):=\sum_{f\in\Faces_v}\int_f\|J_f(\vec{u}_v-\vec{u})+\delta\vec{x}_v\|^2_{M(\vec{x})}\exterior\Omega\,,
\end{align}
where both energy terms can be unified into one term by the norm induced by
\begin{align}
    M(\vec{x})=\vec{n}(\vec{x})\cdot\vec{n}(\vec{x})^t+\lambda\cdot\1\,,
\end{align}
\cf Eq.\ (13) of the main work. To obtain the known form of a quadratic problem,
\begin{align}
    \label{eqn:default_form}
    Q_v(\delta\vec{x}_v)=\langle\delta\vec{x}_v,A_v\delta\vec{x}_v\rangle-2\langle\vec{b}_v,\delta\vec{x}_v\rangle+c_v
\end{align}
we have to apply the binomial formula to the integrand and rearrange the addends. In the end, we get
\begin{align}
    A_v=\sum_{f\in\Faces_v}\int_f M(\vec{x})\exterior\Omega
\end{align}
for the quadratic part,
\begin{align*}
    \vec{b}_v=\sum_{f\in\Faces_v}\int_f M(\vec{x})\cdot J_f(\vec{u}_v-\vec{u})\exterior\Omega
\end{align*}
for the linear part and
\begin{align}
    c_v=\sum_{f\in\Faces_v}\int_f \left\|J_f(\vec{u}_v-\vec{u})\right\|_{M(\vec{x})}\exterior\Omega
\end{align}
for the constant part. 
\par
In the discretised version (Eq. 12 of the main document), we replace the integral by a sum:
\begin{align}
    Q_v(\delta\vec{x})=\sum_{f\in\Faces_v}\frac{A^{(3)}_f}{|\Pixels_f|}\sum_{p\in\Pixels_f}\left\|J_f\delta\vec{u}_{vp}+\delta\vec{x}\right\|_{M_p}^2\,.
\end{align}
Similarly, the integral is replaced by a sum in the coefficients of the linear system:
\begin{align}
    A_v         &=  \sum_{f\in\Faces_v}\frac{A^{(3)}_f}{|\Pixels_f|}M_p\,, \\
    \vec{b}_f   &=  \sum_{f\in\Faces_v}\frac{A^{(3)}_f}{|\Pixels_f|}\cdot M_p\cdot J_f(\vec{u}_v-\vec{u})\,,   \\
    c_f         &=  \sum_{f\in\Faces_v}\frac{A^{(3)}_f}{|\Pixels_f|}\left\|J_f(\vec{u}_v-\vec{u})\right\|_{M_p}\,,
\end{align}
where $M_p=\vec{n}_p\cdot\vec{n}_p^t+\lambda\1$ from the pixel normals $\vec{n}_p$.
However, when solving for the optimal vertex positions, we consider the following quadric: 
\begin{align}
    \tilde{Q}_v(\delta\vec{u}_v):=Q_v(J_f\cdot\delta\vec{x}_v)
\end{align}
which is now in $\R^2$ since $J_f:\R^2\rightarrow\R^3$. Hence, we must replace $\delta\vec{x}\mapsto J_f\cdot\delta\vec{x}_v$ in \cref{eqn:default_form}. As a result, the coefficients of this quadratic function are:
\begin{align}
    \tilde{A}_v         &=  J_f^tA_vJ_f\,, \\
    \tilde{\vec{b}}_f   &=  J_f^t\vec{b}_v\,,   \\
    \tilde{c}_v         &=  c_v\,.
\end{align}
To find the final displacement $\delta\vec{u}_v$ that moves vertex $v$ into its optimal position, we solve
\begin{align}
    \tilde{A}_v\delta\vec{u}=\tilde{\vec{b}}_v\,.
\end{align}
By doing so, we neglect the influence of a vertex displacement on the adjacent vertices and their quadrics, which is a common approximation in mesh-processing \cite{Chen:2004,Xu:2024}.
\begin{table*}[h]
    \centering
    \begin{tabular}{l|S[table-format=2.2]|S[table-format=2.2]S[table-format=2.2]S[table-format=2.2]|S[table-format=2.2]S[table-format=2.2]S[table-format=2.2]}
        \toprule
        &   \multicolumn{1}{c|}{\cite{Cao:2021}} 
        &   \multicolumn{3}{c|}{{Isotropic \cite{Heep:2025}}}
        &   \multicolumn{3}{c}{Ours}    \\
        \midrule
        Dataset  &  \multicolumn{1}{c|}{Ref} & \multicolumn{1}{c}{low} & \multicolumn{1}{c}{mid} & \multicolumn{1}{c|}{high} & \multicolumn{1}{c}{low} & \multicolumn{1}{c}{mid} & \multicolumn{1}{c}{high}   \\
        \midrule
            \textsc{Ball}       &       0.40    &       0.56    &       0.48    &       0.47    &       0.54    &       0.51    &       0.49    \\
            \textsc{Bell}       &       0.30    &       0.82    &       0.62    &       0.54    &       0.54    &       0.51    &       0.47    \\
            \textsc{Bowl}       &       0.08    &       0.35    &       0.22    &       0.15    &       0.15    &       0.14    &       0.12    \\
            \textsc{Buddha}     &       3.46    &       3.59    &       3.68    &       3.73    &       3.46    &       3.55    &       3.56    \\
            \textsc{Bunny}      &       3.38    &       4.03    &       3.93    &       3.83    &       3.90    &       3.80    &       3.74    \\
            \textsc{Cup}        &       0.01    &       0.36    &       0.20    &       0.08    &       0.06    &       0.03    &       0.02    \\
            \textsc{Die}        &       1.62    &       2.98    &       2.67    &       2.46    &       2.83    &       2.57    &       2.63    \\
            \textsc{Hippo}      &       2.73    &       3.13    &       2.96    &       2.91    &       3.04    &       2.88    &       2.86    \\
            \textsc{House}      &       11.08   &       11.08   &       11.30   &       11.35   &       11.38   &       11.49   &       11.32   \\
            \textsc{Jar}        &       0.50    &       0.55    &       0.46    &       0.43    &       0.43    &       0.43    &       0.43    \\
            \textsc{Owl}        &       4.89    &       6.22    &       5.86    &       5.69    &       6.00    &       5.47    &       5.34    \\
            \textsc{Queen}      &       3.74    &       5.17    &       4.43    &       4.30    &       5.05    &       4.08    &       3.98    \\
            \textsc{Squirrel}   &       1.91    &       2.87    &       2.47    &       2.34    &       2.62    &       2.15    &       2.03    \\
            \textsc{Tool}       &       0.91    &       1.04    &       0.96    &       0.93    &       0.91    &       0.90    &       0.90    \\
        \bottomrule
    \end{tabular}
    \caption{RMSE in mm for all objects of the LUCES dataset using the perspective projection. The vertex count is given by the respective low, mid and high settings of \cite{Heep:2025} and matched by our decimation pipeline.}
    \label{tab:luces_comparison}
\end{table*}
\section{Evaluation}\label{sec:evaluation_details}
In this section, we provide additional benchmark results and insights on how the compression ratio and reconstruction error depend on the user-set decimation threshold. Further error maps are found in \cref{sec:error_maps}. We also discuss two interesting outliers: One, where our anisotropic decimation method is more accurate than the pixel-based reference and one, where a higher decimation threshold, \ie lower resolution mesh, surpasses higher resolutions. At the end of this section, we depict reconstructions of all objects of the LUCES \cite{Mecca:2021}, RGBN \cite{Toler-Franklin:2007} and PS \cite{Frankot:1988} dataset for various values of the decimation threshold and report vertex count and compression ratios.
\subsection{Benchmark Comparison}
To evaluate our method against previous work on mesh-based integration \cite{Heep:2025}, we listed only the orthographic projection for the DiLiGenT-MV dataset \cite{Li:2020} in the main document. We complement this comparison with the results of the perspective projection in \cref{tab:old_vs_new_integration}.
Furthermore, we perform the same evaluation on the LUCES dataset \cite{Mecca:2021}, but only for the perspective projection. Since this dataset is dedicated to near-field photometric stereo, \ie objects are very close to the camera, the orthographic projection is an unsuitable approximation. As in previous tests, we match the vertex count to the low, mid and high settings of previous work \cite{Heep:2025}. Results are listed in \cref{tab:luces_comparison}. Again, our method generates tighter approximations of the underlying surface.
\subsubsection{Inspecting the Buddha Results in DiLiGenT-MV}
\begin{figure}
    \centering
    \begin{subfigure}[c]{.4\linewidth}
        \centering
        \large{Pixel-Based} \\  \vspace{2mm}
        \includegraphics[trim={18mm 0 18mm 0},clip,width=\linewidth]{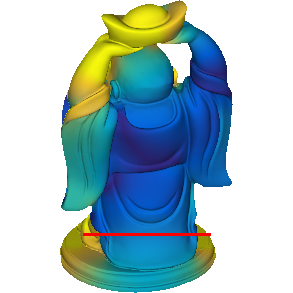}
        \\
        5.17mm
    \end{subfigure}
    \begin{subfigure}[c]{.4\linewidth}
        \centering
        \large{Ours}    \\  \vspace{2mm}
        \includegraphics[trim={18mm 0 18mm 0},clip,width=\linewidth]{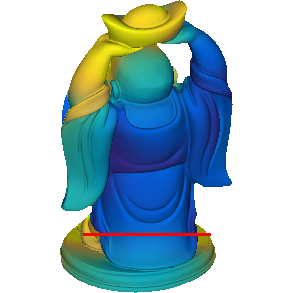}
        \\
        4.72mm
    \end{subfigure}
    \begin{subfigure}[c]{.15\linewidth}
        \includegraphics[width=\linewidth]{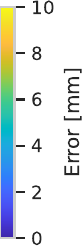}
    \end{subfigure}
    \\
    \includegraphics[width=\linewidth]{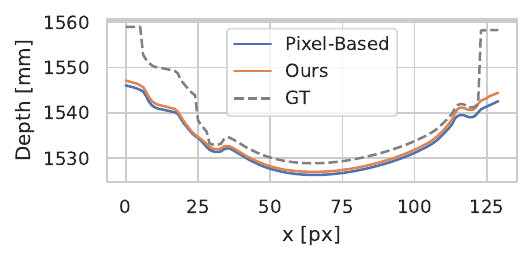}
    \caption{Top: Error map and RMSE for pixel-based \cite{Cao:2021} (\emph{left}) and our (\emph{right}) integration. Bottom: Slice of the aligned depth maps (along the red line indicated above).}
    \label{fig:diligent_buddha}
\end{figure}

We noticed that our decimation-based method at the 'high' accuracy setting outperforms the pixel-based baseline in the case of the \textsc{Buddha} figurine in the DiLiGenT-MV dataset. This is surprising as our method uses fewer vertices than the pixel-based method. To investigate the origin of this result, we examined the differences in each view. \Cref{fig:diligent_buddha} illustrates the error map for the tenth view. While our decimation is generally lossy -- except for perfectly flat regions -- discontinuities and highly slanted (near-discontinuous) surfaces are problematic and error prone regions for normal integration. For the Buddha statue, such surfaces are present around the base and the lower part of the garment. It seems that our method performs slightly better in these situations.
We believe this is due to differences between our integrator and the integrator in \cite{Cao:2021}. In \cite{Cao:2021} all normals are weighted equally while our method compensates for foreshortening assigning a higher weight to normals in slanted regions, as their real surface is bigger than it appears in screen space.
\subsubsection{Inspecting the Buddha Results in LUCES}
\begin{figure}
    \centering
    \begin{subfigure}[t]{.49\linewidth}
        \centering
        \includegraphics[trim={170mm 420mm 80mm 30mm},clip,width=\linewidth]{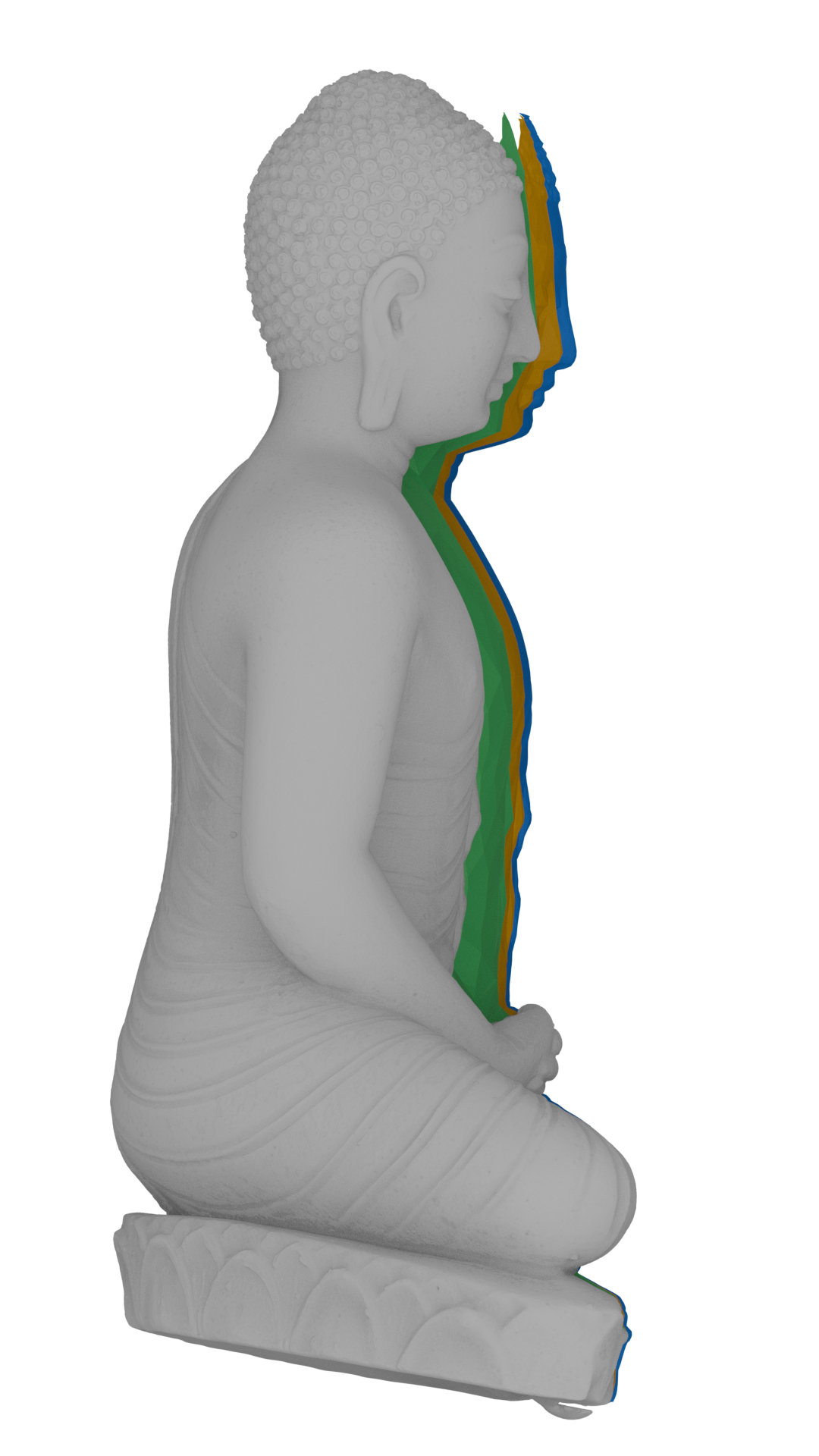}
    \end{subfigure}
    \begin{subfigure}[t]{.49\linewidth}
        \centering
        \includegraphics[trim={170mm 420mm 80mm 30mm},clip,width=\linewidth]{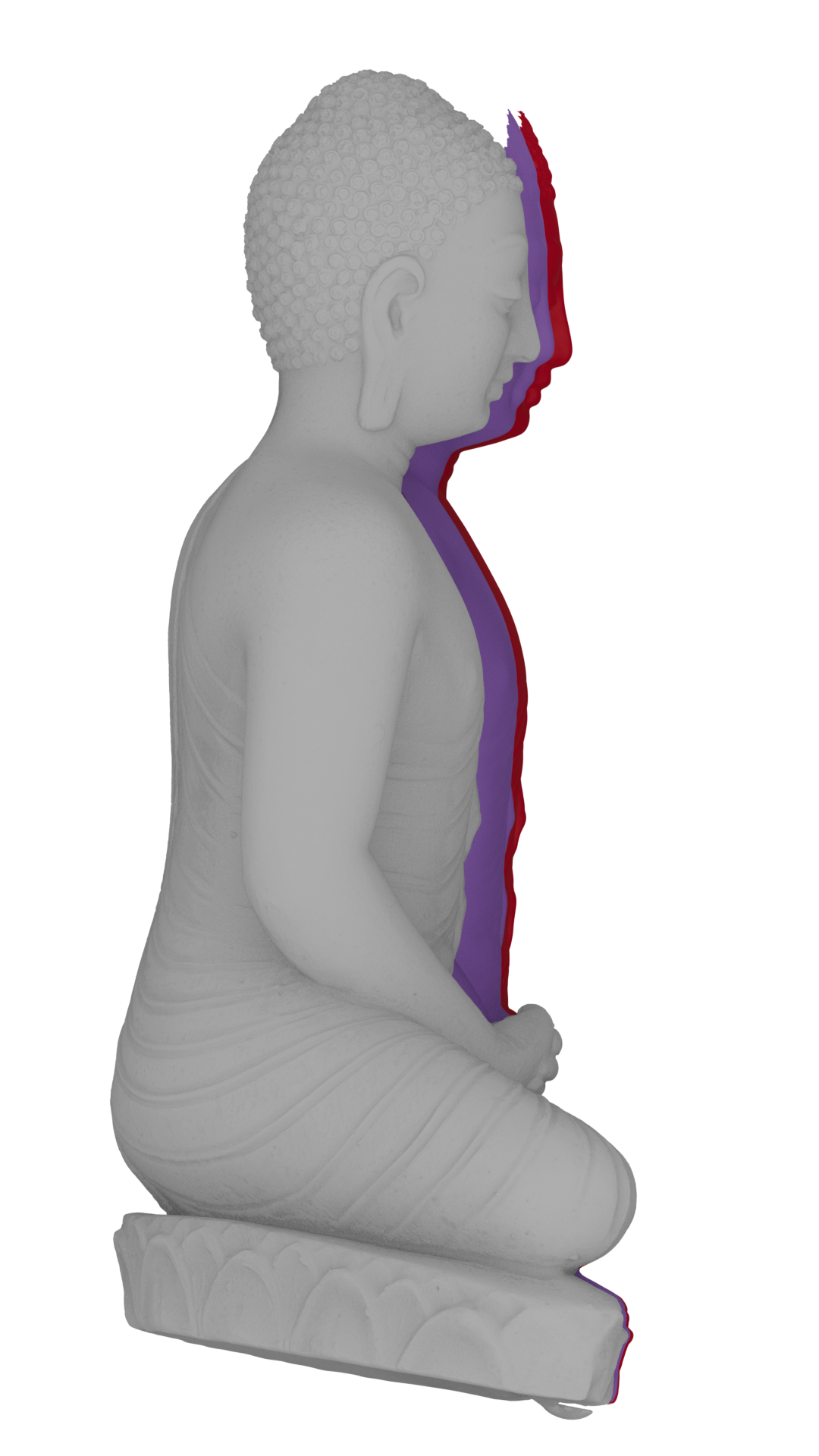}
    \end{subfigure}
    \caption{Side profiles of the \textsc{Buddha} figurine in LUCES \cite{Mecca:2021}. Left: Our decimation with thresholds $2048$ (green), $64$ (yellow) and $2$ (blue). Right: Pixel-based integration \cite{Heep:2022} with ground-truth normals (red) and smoothed normals (purple). All examples were aligned to the ground-truth surface (grey) at the base of the figurine.}
    \label{fig:luces_buddha}
\end{figure}
The \textsc{Buddha} figurine in the LUCES dataset shows an atypical behaviour: Lower mesh resolutions yield a better surface approximation. This inverted connection between resolution and reconstruction quality is also present for the previous isotropic normal integration \cite{Heep:2025}, \cf \cref{tab:luces_comparison}. A visual inspection revealed that the normal integration places the Buddha's face too far in the front compared to the figurine's base, see \cref{fig:luces_buddha}. This is true for both pixel- and mesh-based integration. At lower mesh resolutions, there are more pixels per triangle which implicitly smoothens the surface normals and flattens the integrated surface. This flattening coincidentally reduces the constant offset to the ground-truth surface. This hypothesis is supported by the fact that we can recreate this behaviour in pixel-based integration by applying a Gaussian kernel to the normal map, see \cref{fig:luces_buddha}.
\subsection{Controllability}
\begin{figure*}
    \centering
    \includegraphics[,width=\linewidth]{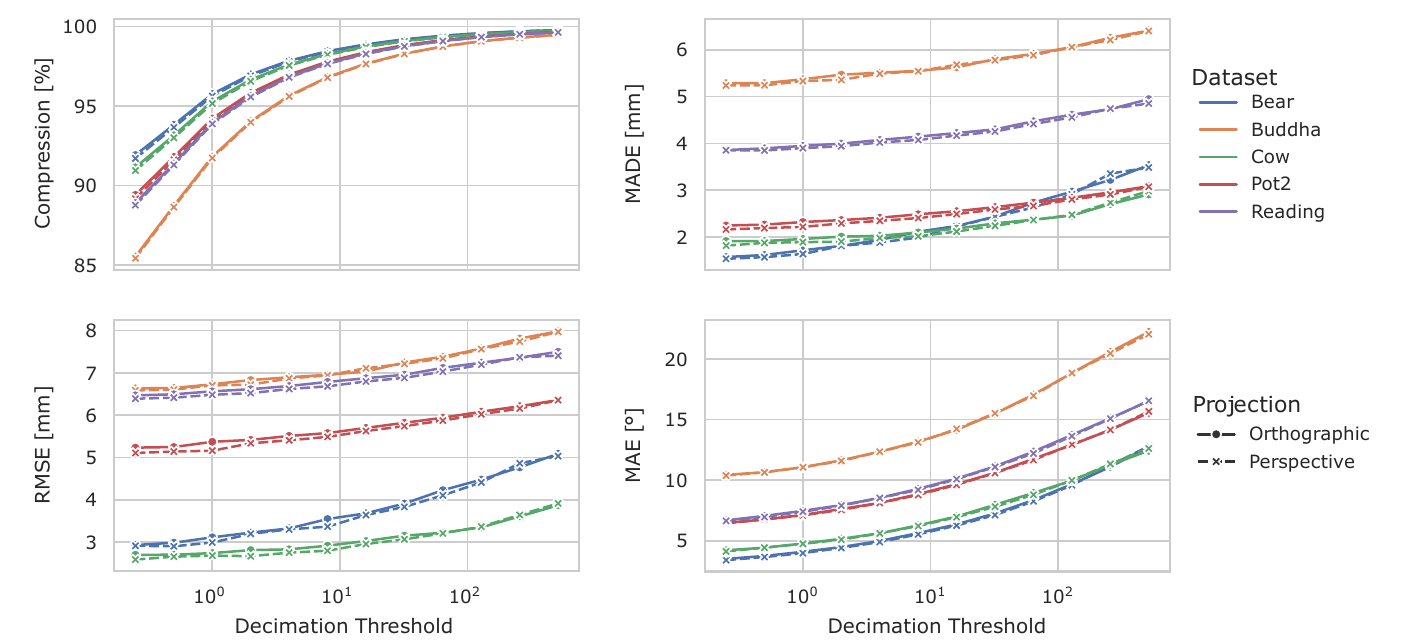}
    \caption{Influence of the decimation threshold on compression ratio, RMSE, MADE and MAE. All numbers are averages over the 20 views for each object. We investigate both orthographic and perspective projection. Please note the logarithmic $x$-axis.}
    \label{fig:diligent_controllability}
\end{figure*}

In the main paper, we studied the controllability using the LUCES \cite{Mecca:2021} dataset. For completeness, we list all results in ~\cref{tab:luces}. For a more extensive study of the influence of the decimation threshold on the final mesh quality, we tested all normal maps of the DiLiGenT-MV dataset \cite{Li:2020} for threshold values ranging from $0.25$ to $512$ and evaluated both root mean square error (RMSE) and mean absolute deviation (MADE) --  both after the appropriate rigid-alignment to absolve the scale ambiguity -- as well as mean angular error (MAE) and vertex count. The results are depicted in \cref{fig:diligent_controllability}. As expected, there is virtually no difference between the results for orthographic and perspective projection. All objects are far away from the camera, \ie the orthographic projection is a good approximation of the true perspective projection. The compression ratios reflect how our algorithm adapts to the complexity of the datasets: The \textsc{Bear} and \textsc{Cow} mostly consist of smooth featureless surfaces and achieve higher compression ratios than the more complex \textsc{Buddha} dataset, especially for lower thresholds.
\begin{table}[h]
    \centering
    \begin{tabular}{l|S[table-format=2.2]|S[table-format=2.2]S[table-format=2.2]S[table-format=2.2]}
        \toprule
        \multirow{2}{*}{Dataset}    &   \multicolumn{1}{c}{Ref} &   \multicolumn{3}{c}{Threshold}    \\
        \cmidrule{2-5}              &   \cite{Heep:2022}
                                    &   \multicolumn{1}{c}{$2$}
                                    &   \multicolumn{1}{c}{$2^6$}
                                    &   \multicolumn{1}{c}{$2^{11}$} \\
        \midrule
        \textsc{Ball} & 0.40 & 0.48 & 0.54 & 0.60 \\ %
        \textsc{Bell} & 0.30 & 0.33 & 0.49 & 0.72 \\ %
        \textsc{Bowl} & 0.08 & 0.10 & 0.13 & 0.17 \\ %
        \textsc{Buddha} & 3.46 & 3.58 & 3.51 & 3.32 \\ %
        \textsc{Bunny} & 3.38 & 3.67 & 3.83 & 4.15 \\ %
        \textsc{Cup} & 0.01 & 0.01 & 0.04 & 0.14 \\ %
        \textsc{Die} & 1.62 & 1.65 & 2.11 & 3.09 \\ %
        \textsc{Hippo} & 2.73 & 2.85 & 2.91 & 3.09 \\ %
        \textsc{House} & 11.08 & 11.30 & 11.56 & 11.39 \\ %
        \textsc{Jar} & 0.50 & 0.45 & 0.45 & 0.46 \\ %
        \textsc{Owl} & 4.89 & 5.41 & 5.63 & 5.94 \\ %
        \textsc{Queen} & 3.74 & 4.02 & 4.41 & 4.98 \\ %
        \textsc{Squirrel} & 1.91 & 2.07 & 2.37 & 3.00 \\ %
        \textsc{Tool} & 0.91 & 0.98 & 0.98 & 1.04 \\ %
        \bottomrule
    \end{tabular}
    \caption{RMSE in mm for all of the objects in LUCES \cite{Mecca:2021} with increasing decimation threshold. The chosen thresholds should lead to a constant reduction rate of the RMSE.}
    \label{tab:luces}
\end{table}

\clearpage
\onecolumn
\subsection{Error Maps for the Benchmark Comparisons}
\label{sec:error_maps}
\begin{figure*}[!h]
    \centering
    \LARGE{LUCES Dataset (1 of 3)} \\
    \begin{subfigure}[b]{.48\linewidth}
        \centering
		\large{\textsc{\capitalize{Ball}}}\\ \vspace{2mm}
        \includegraphics[width=\linewidth]{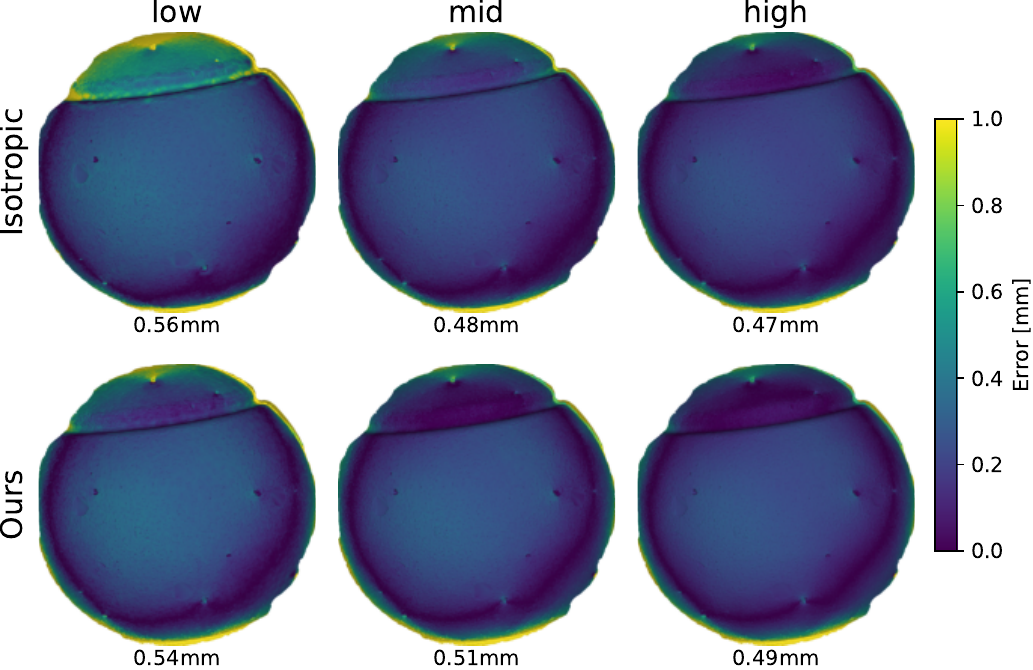}
    \end{subfigure}\hspace{2mm}
    \begin{subfigure}[b]{.48\linewidth}
        \centering
		\large{\textsc{\capitalize{Bell}}}\\ \vspace{2mm}
        \includegraphics[width=\linewidth]{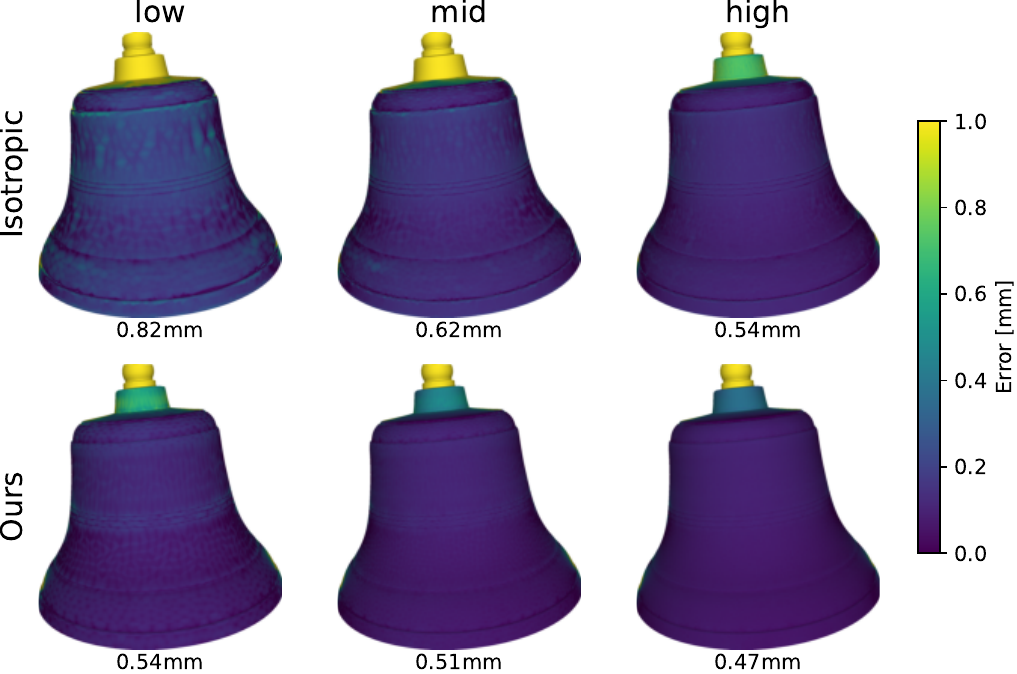}
    \end{subfigure}
    \\\vspace{5mm}
    \begin{subfigure}[b]{.48\linewidth}
        \centering
		\large{\textsc{\capitalize{Bowl}}}\\ \vspace{2mm}
        \includegraphics[width=\linewidth]{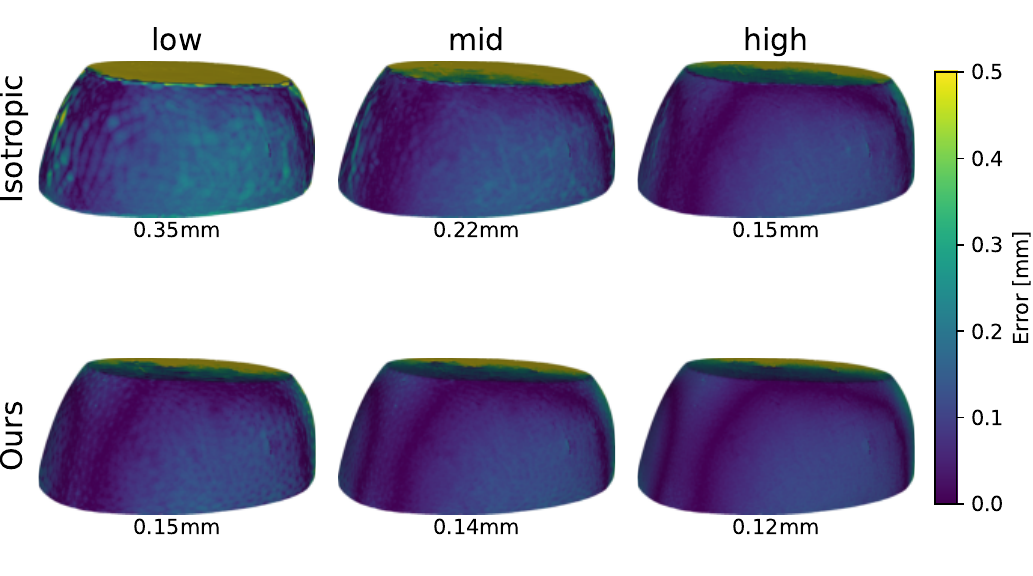}
    \end{subfigure}\hspace{2mm}
    \begin{subfigure}[b]{.48\linewidth}
        \centering
		\large{\textsc{\capitalize{Buddha}}}\\ \vspace{2mm}
        \includegraphics[width=\linewidth]{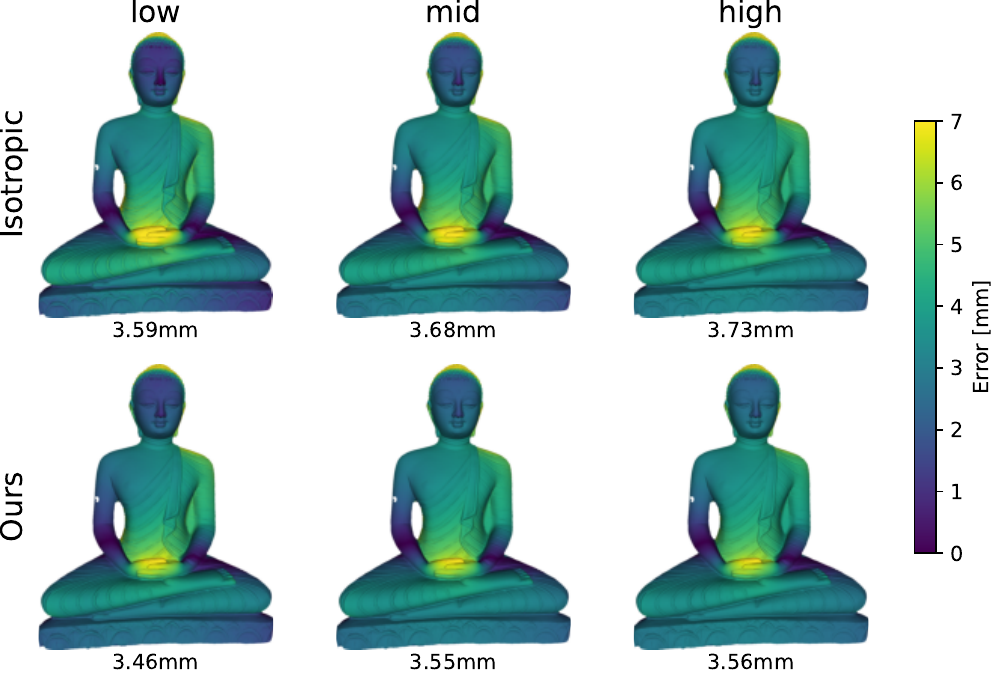}
    \end{subfigure}
    \caption{Error maps of the LUCES dataset after rigid alignment. We show results for all three quality settings in \cite{Heep:2025} and match the respective vertex number for our method. Pictured are the results of the perspective projection.}
\end{figure*}
\FloatBarrier
\begin{figure*}
    \centering
    \LARGE{LUCES Dataset (2 of 3)} \\
    \vspace{5mm}
	\begin{subfigure}[b]{.48\linewidth}
        \centering
		\large{\textsc{\capitalize{Bunny}}}\\ \vspace{2mm}
        \includegraphics[width=.95\linewidth]{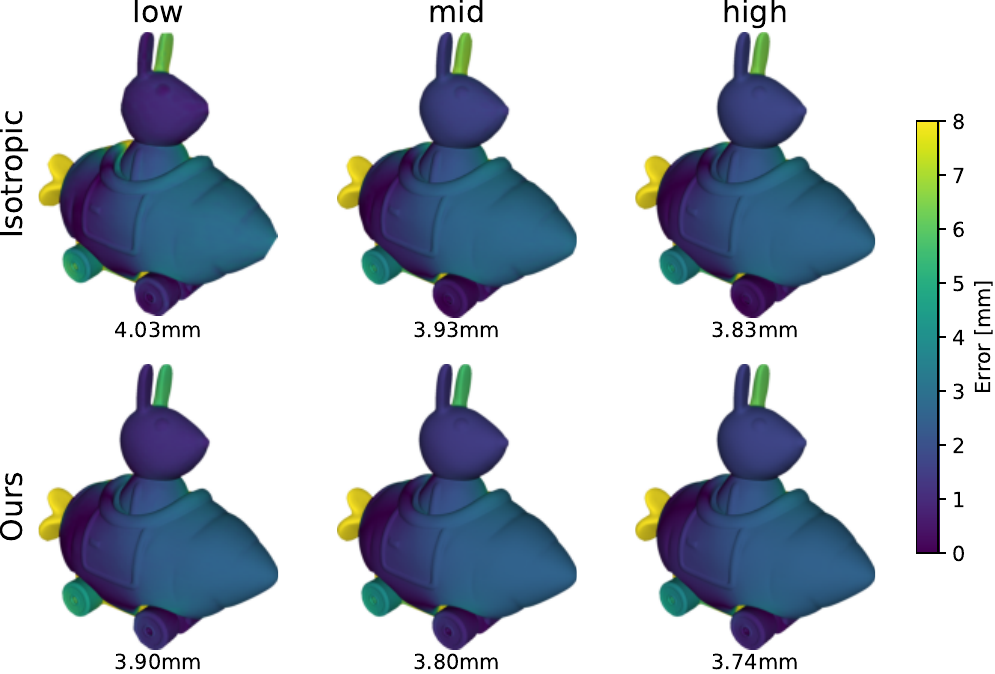}
    \end{subfigure}\hspace{2mm}
    \begin{subfigure}[b]{.48\linewidth}
        \centering
		\large{\textsc{\capitalize{Cup}}}\\ \vspace{2mm}
        \includegraphics[width=\linewidth]{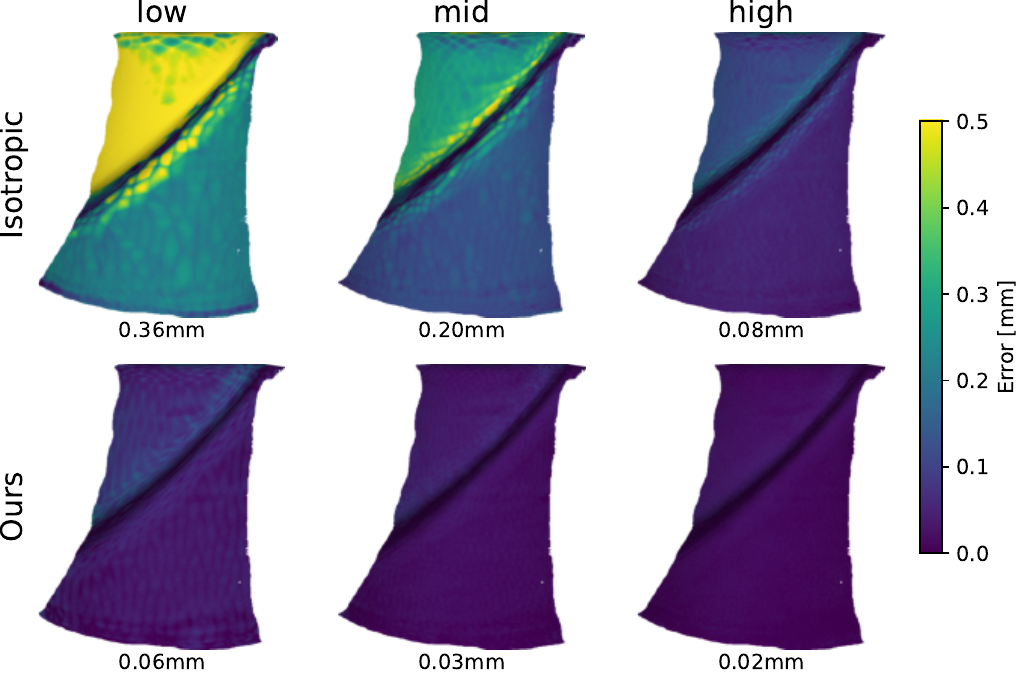}
    \end{subfigure}
	\begin{subfigure}[b]{.48\linewidth}
        \centering
		\large{\textsc{\capitalize{Die}}}\\ \vspace{2mm}
        \includegraphics[width=\linewidth]{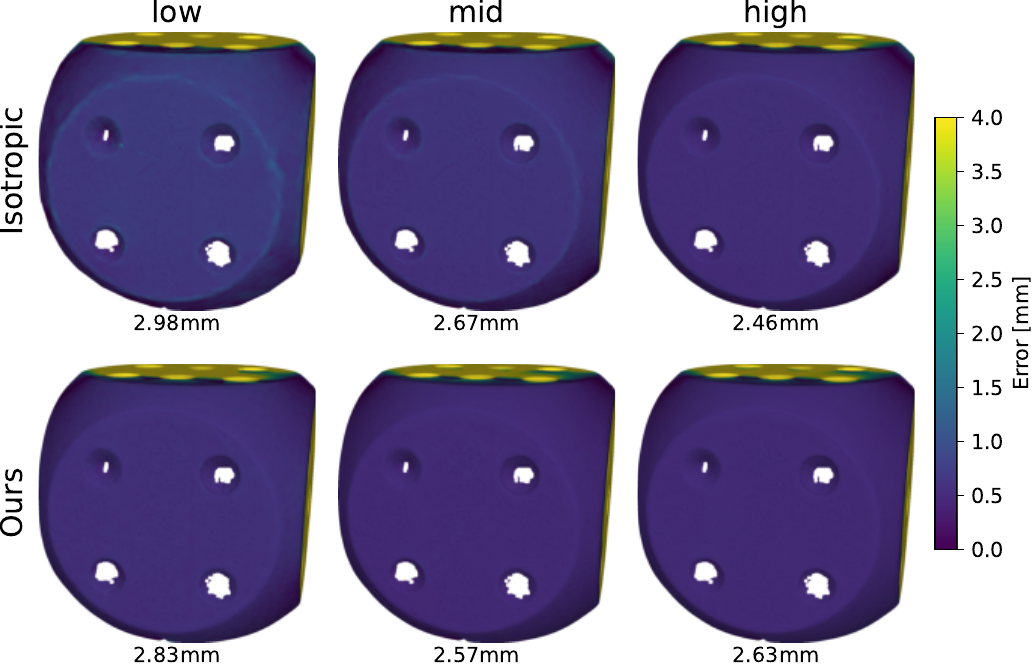}
    \end{subfigure}\hspace{2mm}
	\begin{subfigure}[b]{.48\linewidth}
        \centering
		\large{\textsc{\capitalize{Hippo}}}\\ \vspace{2mm}
        \includegraphics[width=\linewidth]{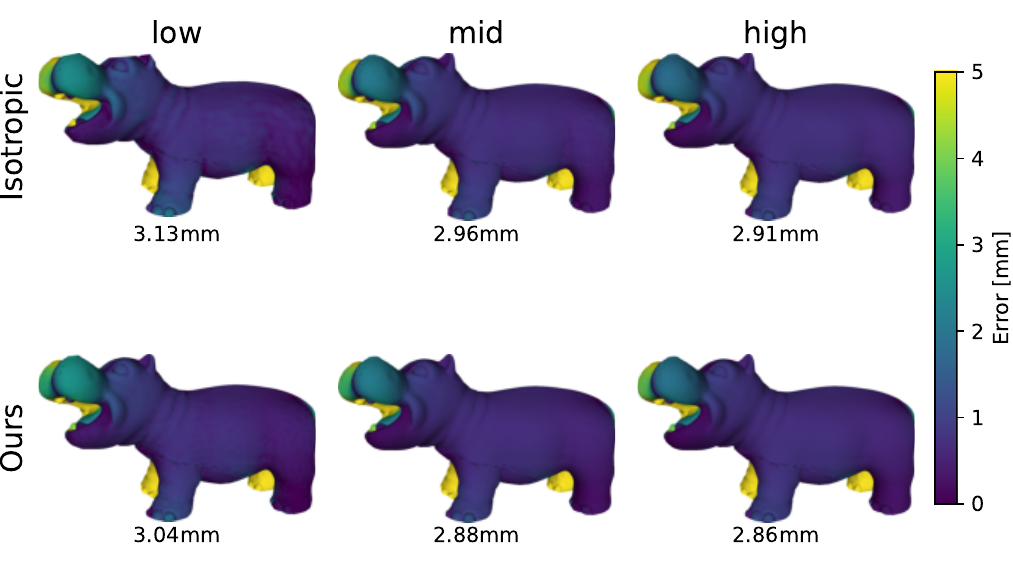}
    \end{subfigure}
    \\\vspace{5mm}
    \begin{subfigure}[b]{.48\linewidth}
        \centering
		\large{\textsc{\capitalize{House}}}\\ \vspace{2mm}
        \includegraphics[width=.95\linewidth]{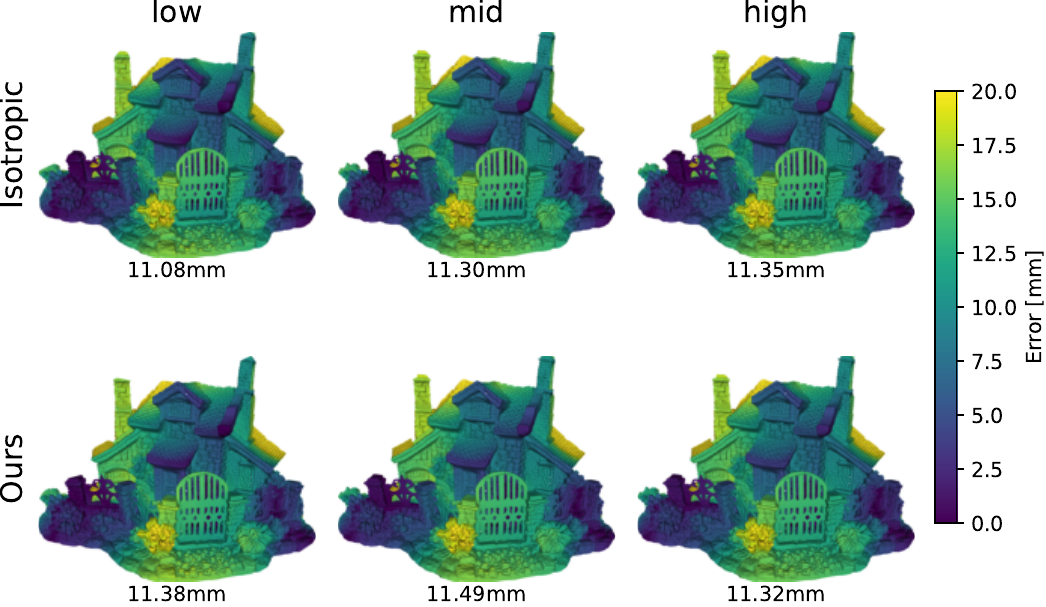}
    \end{subfigure}\hspace{2mm}
    \begin{subfigure}[b]{.48\linewidth}
        \centering
		\large{\textsc{\capitalize{Jar}}}\\ \vspace{2mm}
        \includegraphics[width=\linewidth]{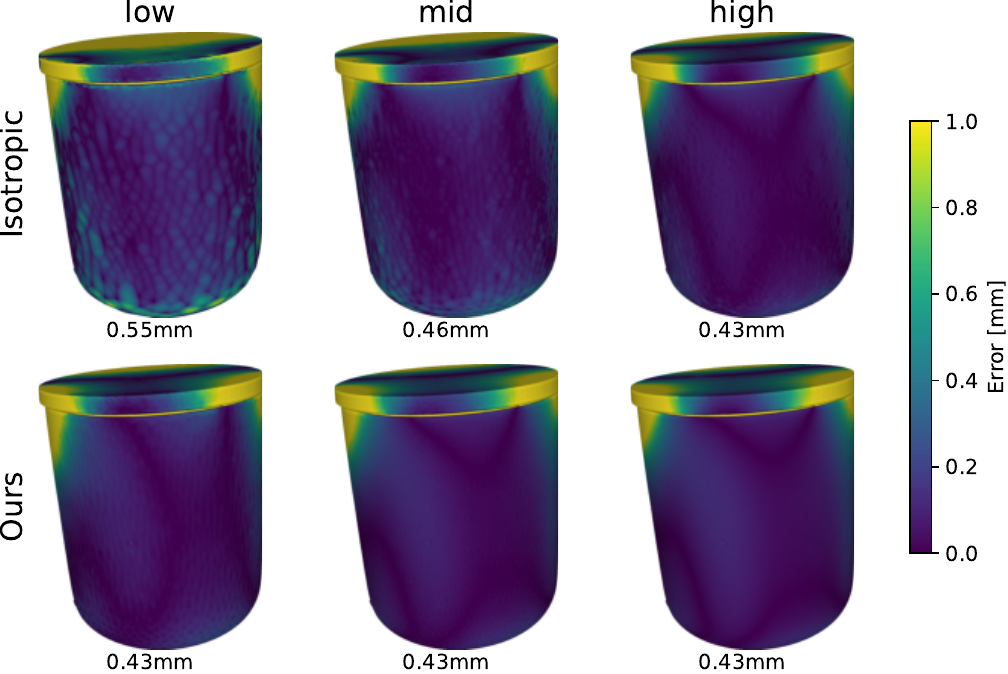}
    \end{subfigure}
    \caption{Error maps of the LUCES dataset after rigid alignment. We show results for all three quality settings in \cite{Heep:2025} and match the respective vertex number for our method. Pictured are the results of the perspective projection.}
\end{figure*}

\begin{figure*}
    \centering
    \LARGE{LUCES Dataset (3 of 3)} \\
    \begin{subfigure}[b]{.48\linewidth}
        \centering
		\large{\textsc{\capitalize{Owl}}}\\ \vspace{2mm}
        \includegraphics[width=\linewidth]{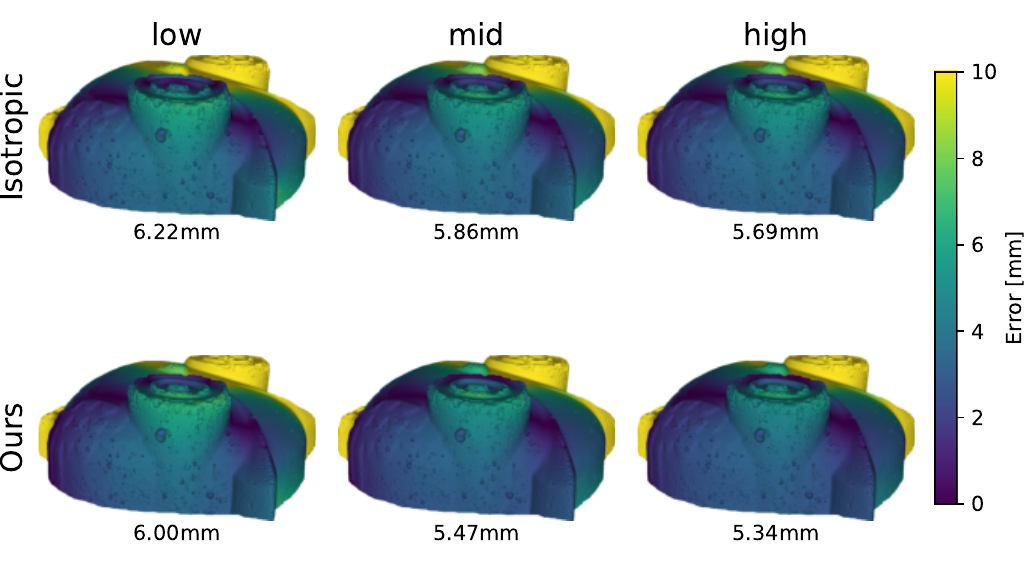}
    \end{subfigure}\hspace{2mm}
    \begin{subfigure}[b]{.48\linewidth}
        \centering
		\large{\textsc{\capitalize{Queen}}}\\ \vspace{2mm}
        \includegraphics[width=\linewidth]{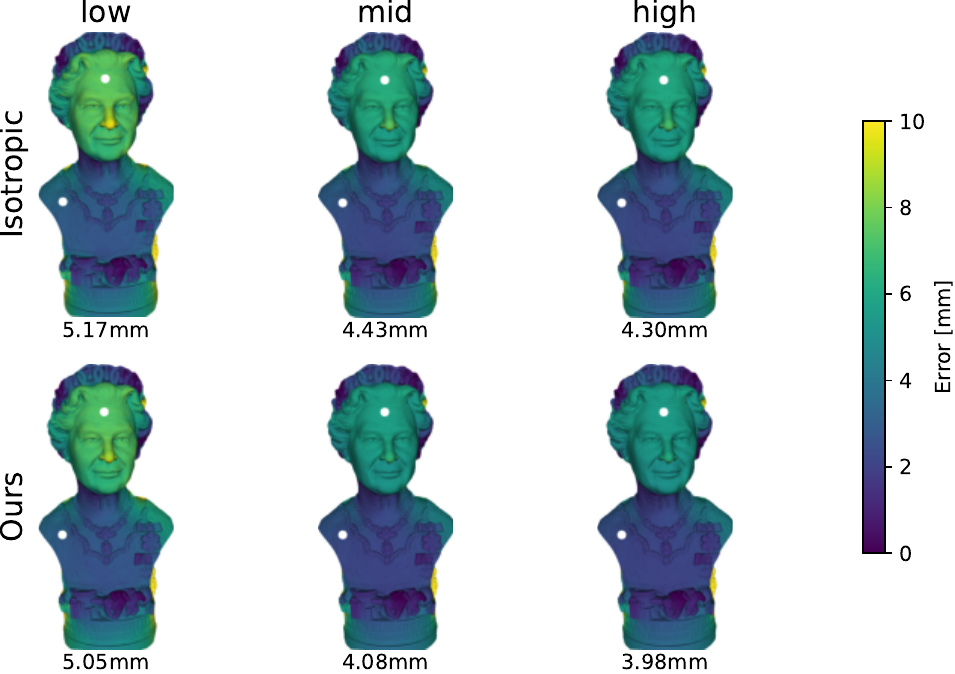}
    \end{subfigure}
    \\\vspace{5mm}
	\begin{subfigure}[b]{.48\linewidth}
        \centering
		\large{\textsc{\capitalize{Squirrel}}}\\ \vspace{2mm}
        \includegraphics[width=\linewidth]{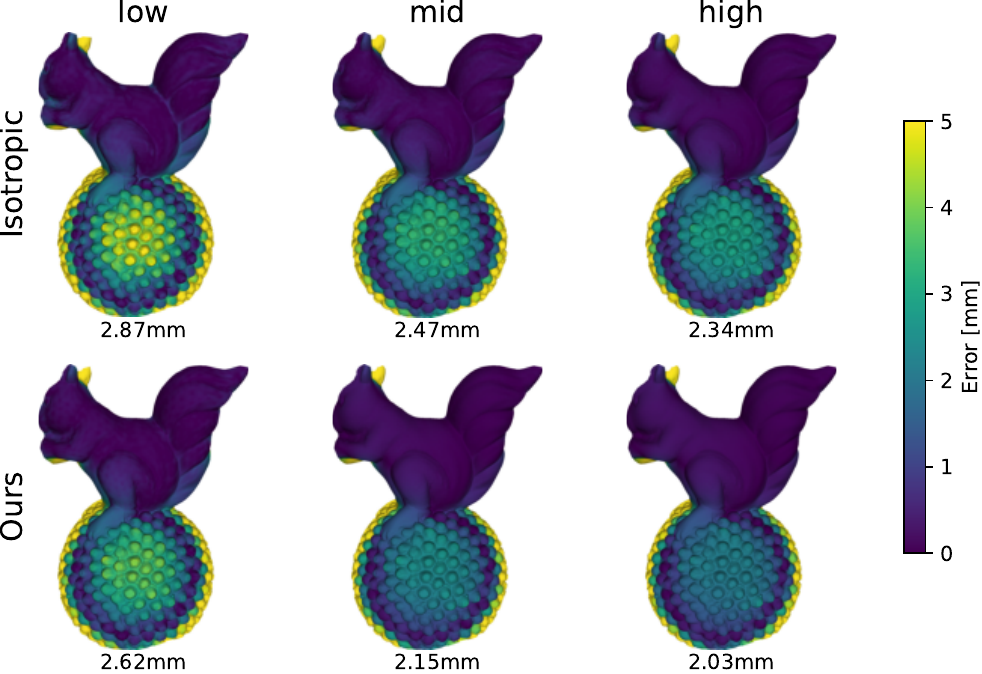}
    \end{subfigure}\hspace{2mm}
	\begin{subfigure}[b]{.48\linewidth}
        \centering
		\large{\textsc{\capitalize{Tool}}}\\ \vspace{2mm}
        \includegraphics[width=\linewidth]{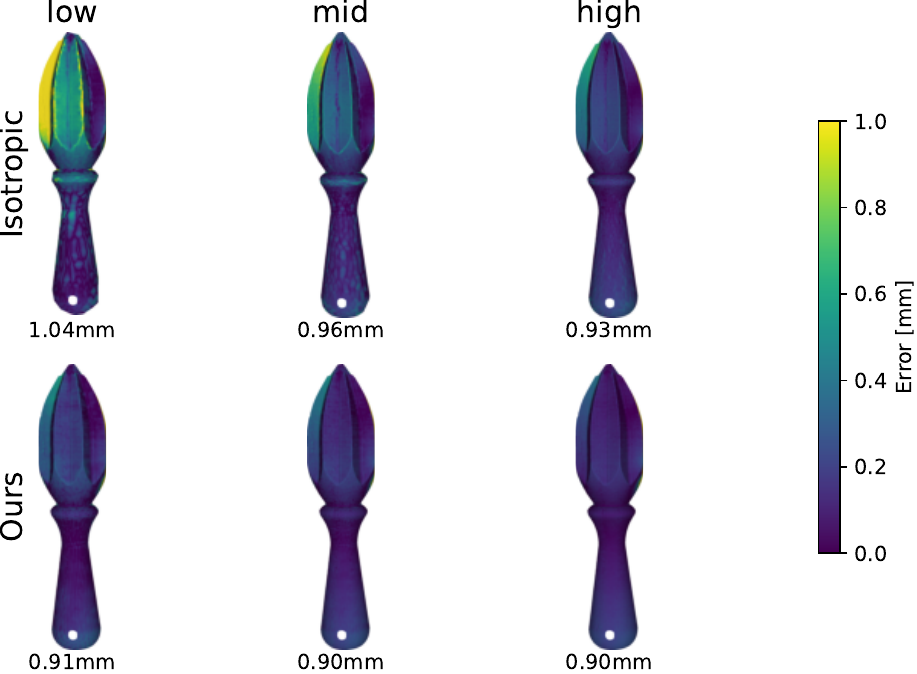}
    \end{subfigure}
    \caption{Error maps of the LUCES dataset after rigid alignment. We show results for all three quality settings in \cite{Heep:2025} and match the respective vertex number for our method. Pictured are the results of the perspective projection.}
\end{figure*}
\begin{figure}[h]
    \centering
    \LARGE{DiLiGenT-MV Dataset} \\
    \begin{subfigure}[b]{.48\linewidth}
        \centering
		\large{\textsc{\capitalize{bear}}}\\ \vspace{2mm}
        \includegraphics[width=\linewidth]{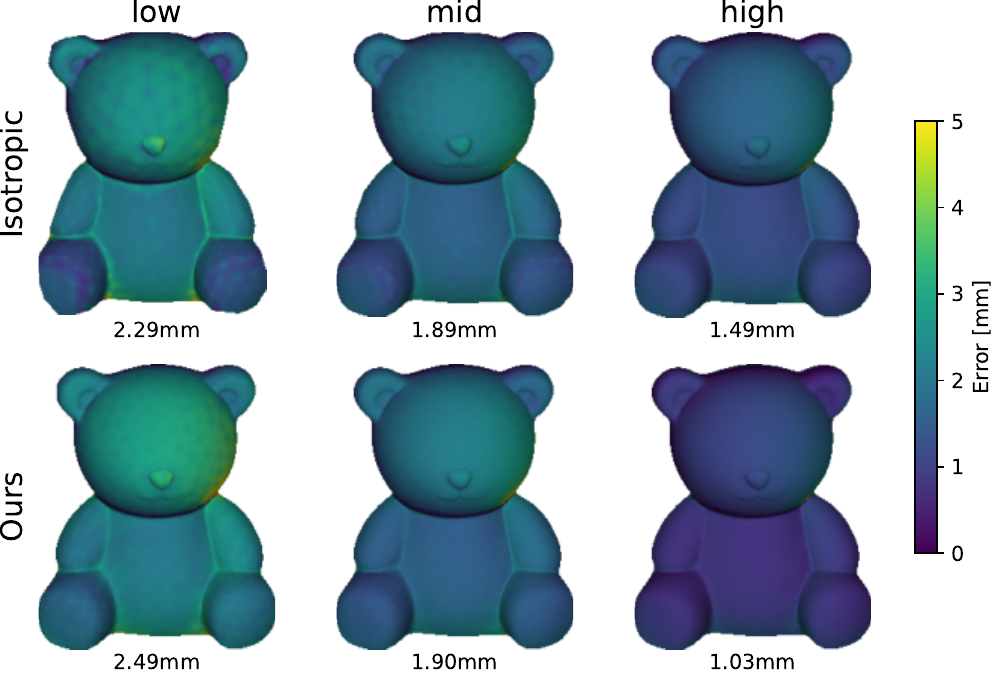}
    \end{subfigure}\hspace{2mm}
    \begin{subfigure}[b]{.48\linewidth}
        \centering
		\large{\textsc{\capitalize{buddha}}}\\ \vspace{2mm}
        \includegraphics[width=\linewidth]{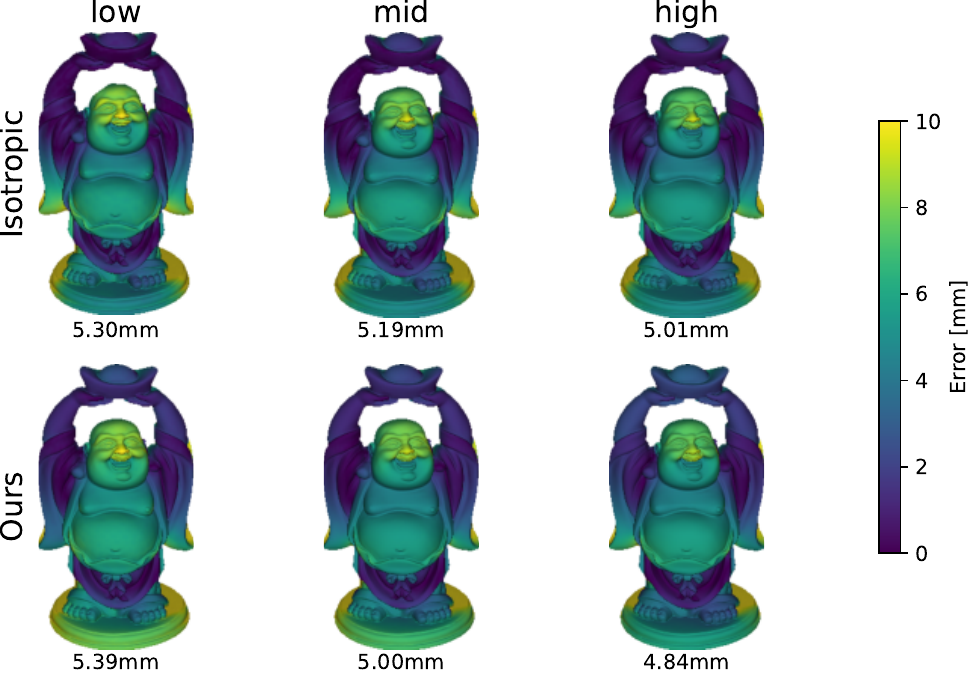}
    \end{subfigure}
    \\\vspace{5mm}
	    \begin{subfigure}[b]{.48\linewidth}
        \centering
		\large{\textsc{\capitalize{cow}}}\\ \vspace{2mm}
        \includegraphics[width=\linewidth]{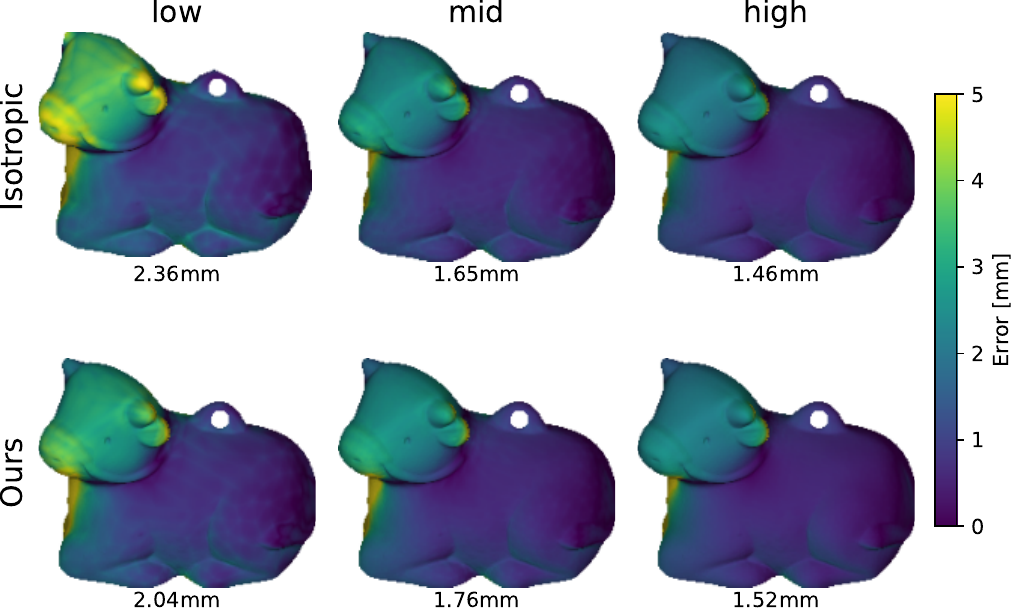}
    \end{subfigure}\hspace{2mm}
    \begin{subfigure}[b]{.48\linewidth}
        \centering
		\large{\textsc{\capitalize{pot2}}}\\ \vspace{2mm}
        \includegraphics[width=\linewidth]{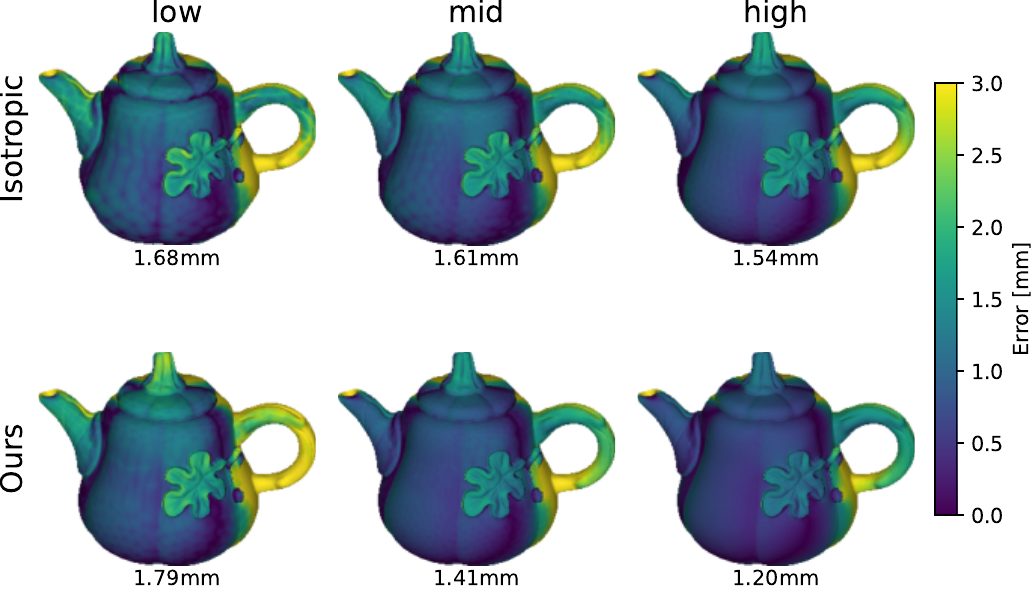}
    \end{subfigure}
    \\\vspace{5mm}
	    \begin{subfigure}[b]{.48\linewidth}
        \centering
		\large{\textsc{\capitalize{reading}}}\\ \vspace{2mm}
        \includegraphics[width=\linewidth]{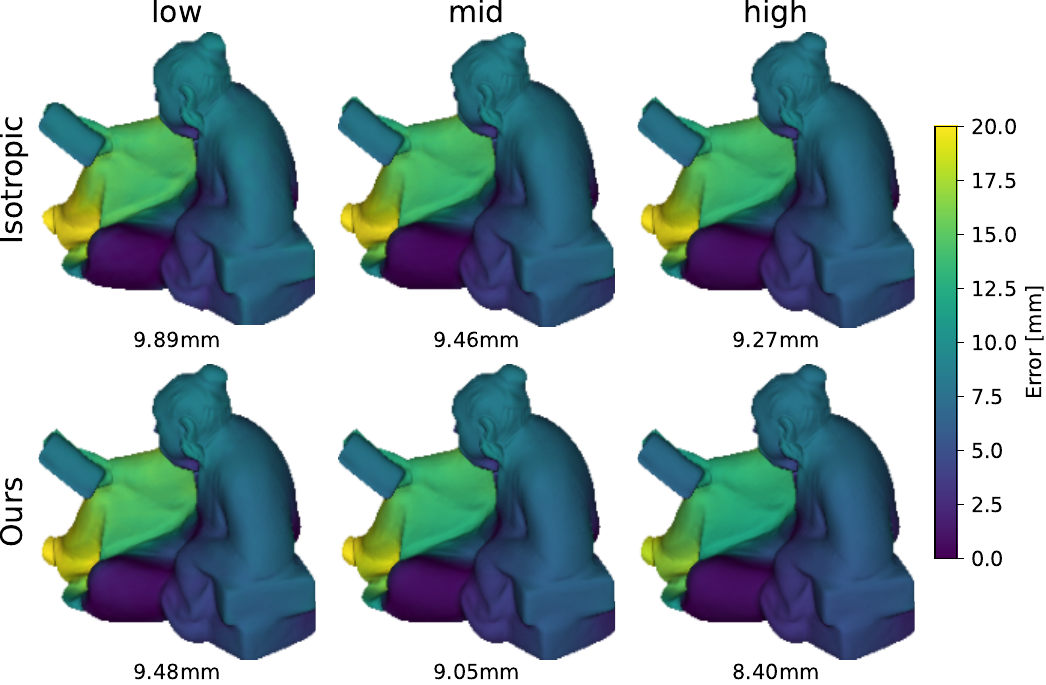}
    \end{subfigure}
    \\
    \caption{Error map for the first view in the DiLiGenT-MV dataset after rigid alignment. We show results for all three quality settings in \cite{Heep:2025} and match the respective vertex number for our method. Pictured are the results of the orthographic projection.}
\end{figure}

\FloatBarrier
\subsection{Additional Datasets}
\begin{figure*}[b]
    \centering
    \LARGE{LUCES Dataset (1 of 3)} \\ \vspace{1mm}
    \begin{tabular}{cccccc}
    &\textsc{{Ball}}&\textsc{{Bell}} &\textsc{{Bowl}}&\textsc{{Buddha}}&\textsc{{Bunny}}\\ \vspace{1mm}
	\rotatebox{90}{\textsc{high-res}} & \includegraphics[width=0.15\linewidth]{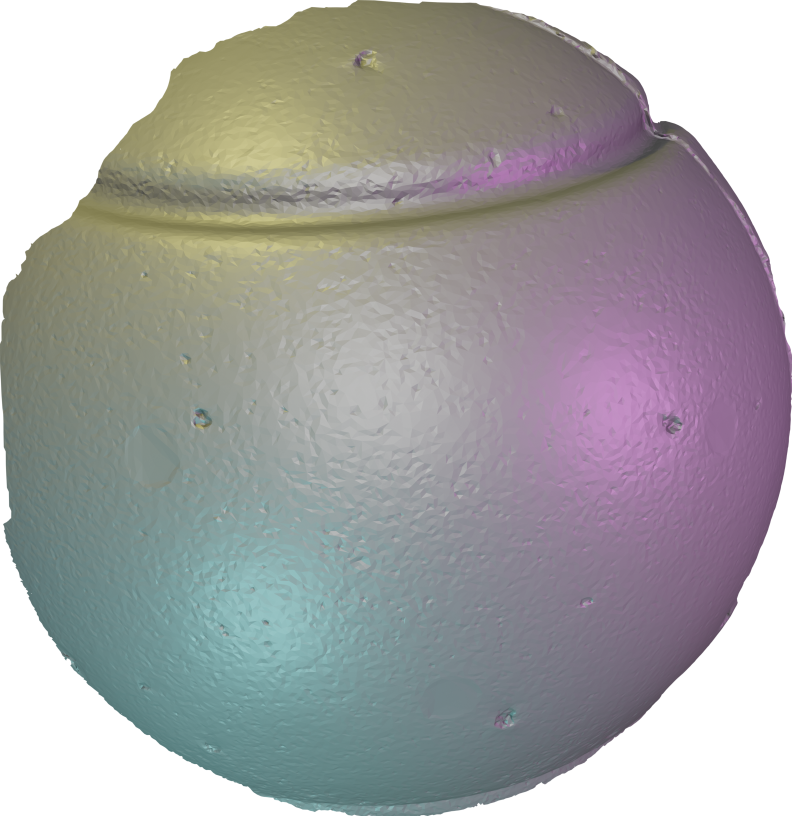}
        &\includegraphics[width=0.15\linewidth]{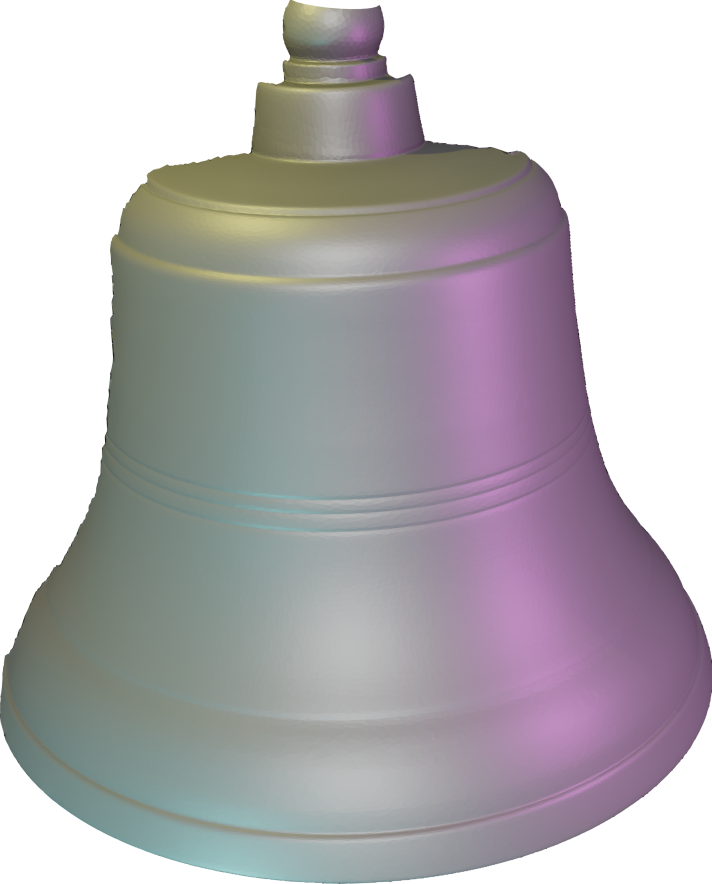}
		&\includegraphics[width=0.17\linewidth]{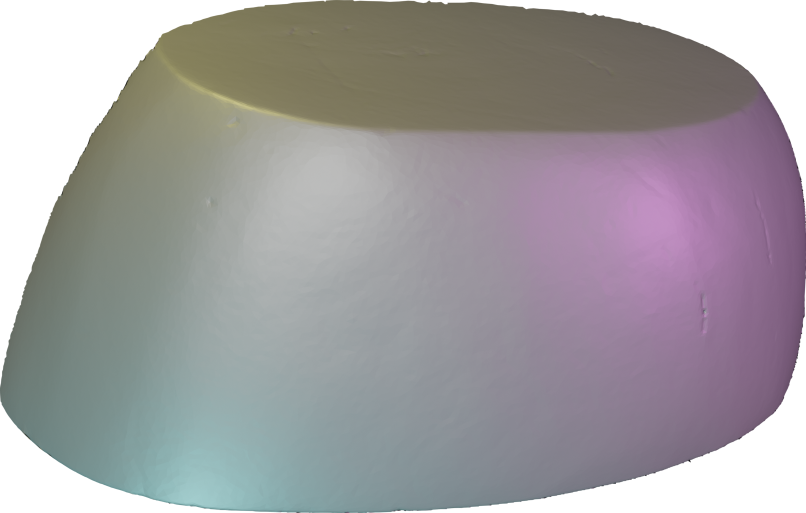}
        &\includegraphics[width=0.17\linewidth]{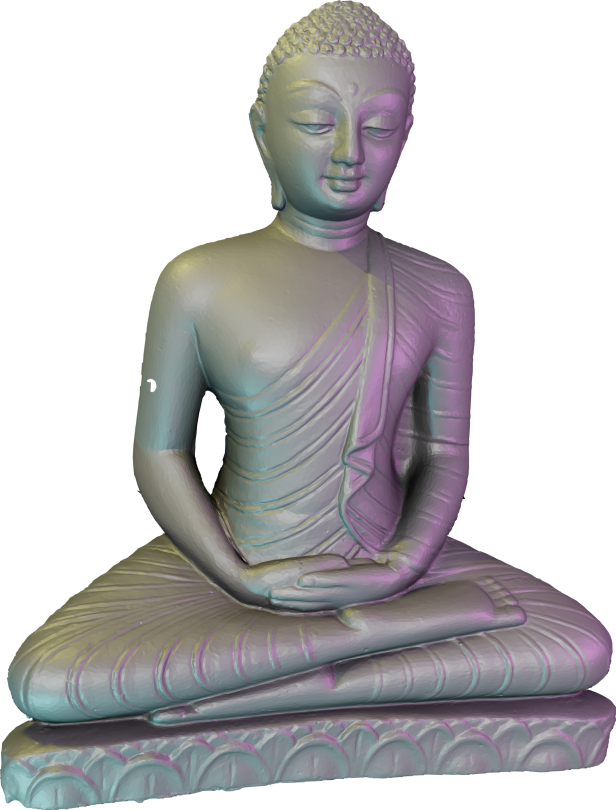} 
        &\includegraphics[width=0.17\linewidth]{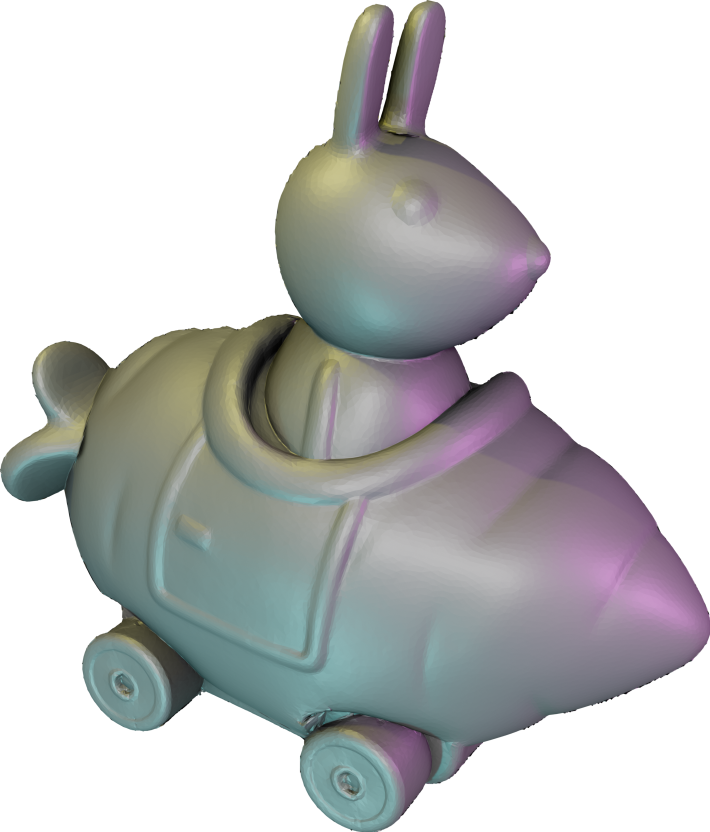} \\
        &       \small{29693 (96.8\%)}  &       \small{21856 (98.3\%)}  &       \small{17710 (98.3\%)}  &       \small{29142 (96.7\%)}  &       \small{16308 (97.8\%)}  \\
		&\includegraphics[width=0.15\linewidth]{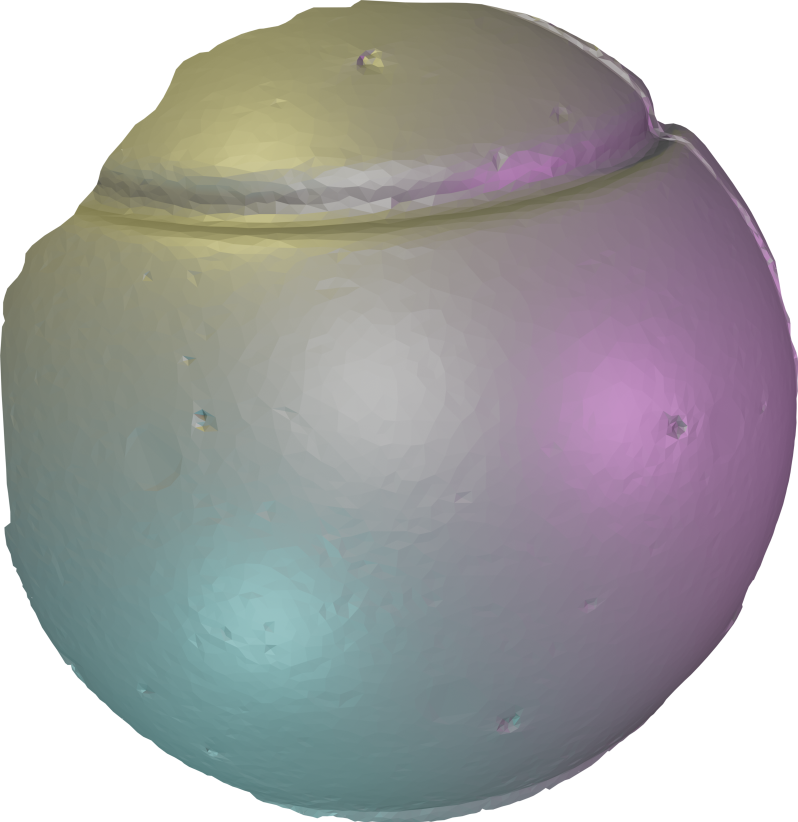}
		&\includegraphics[width=0.15\linewidth]{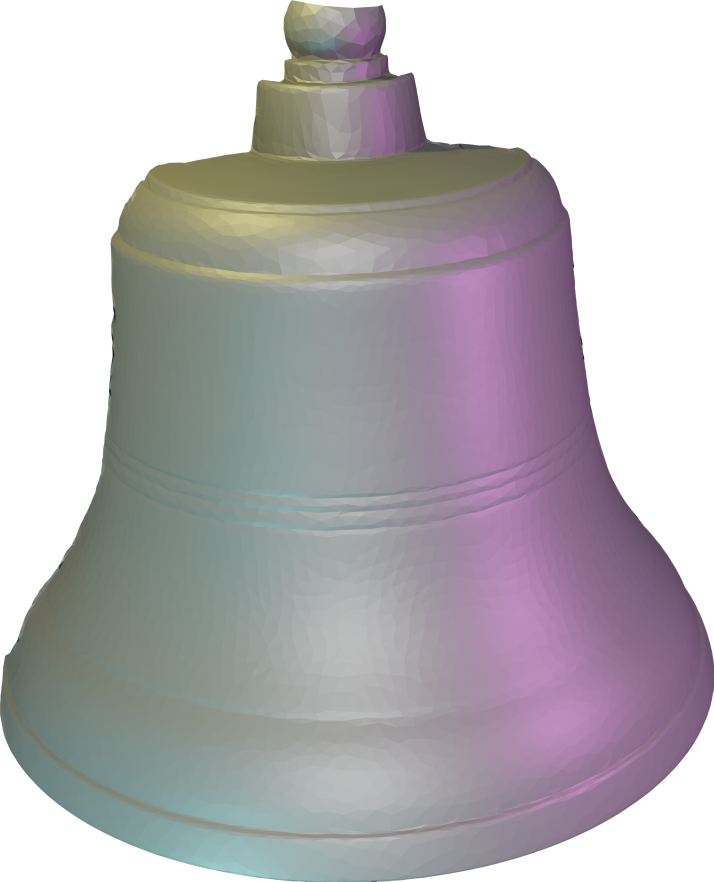}
        &\includegraphics[width=0.17\linewidth]{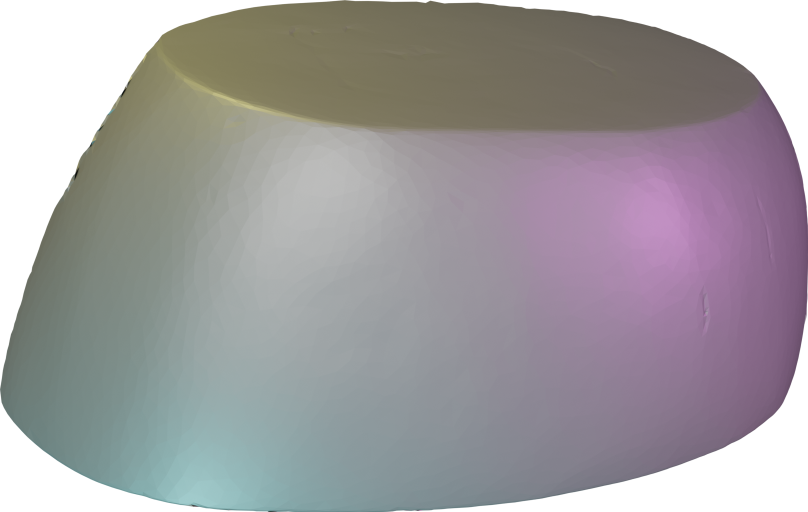}
        &\includegraphics[width=0.17\linewidth]{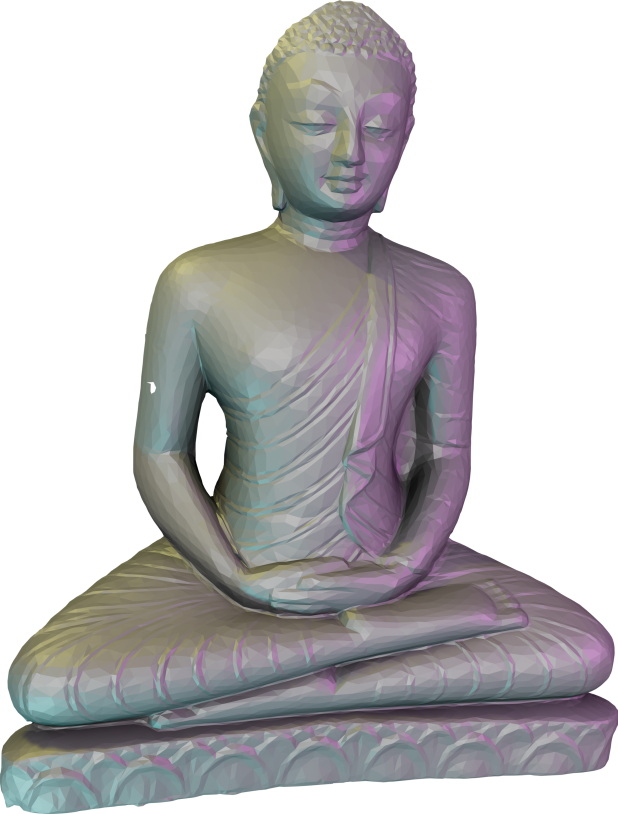}
        &\includegraphics[width=0.17\linewidth]{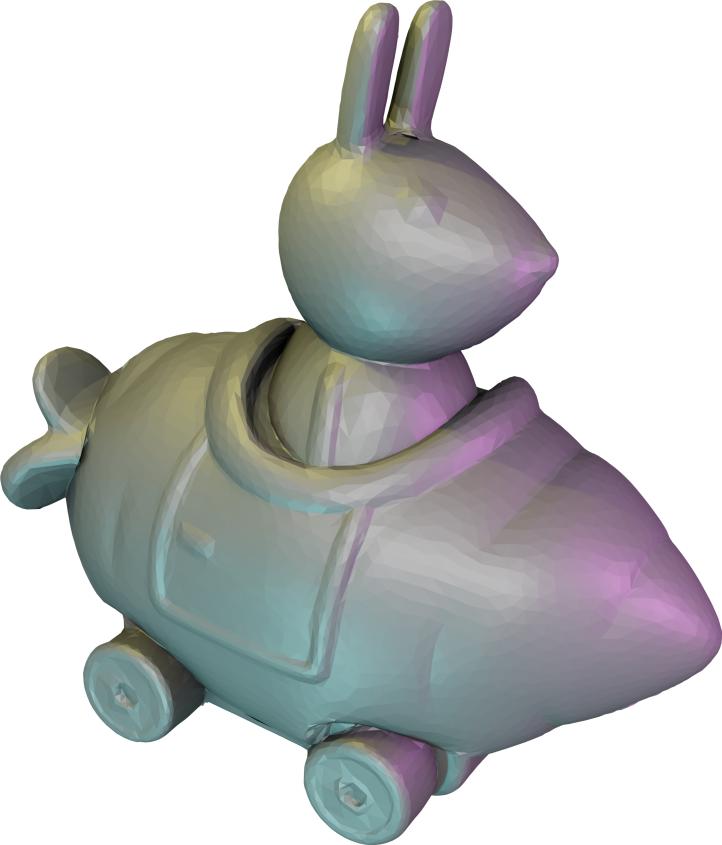}\\
        &       \small{5849 (99.4\%)}   &       \small{4089 (99.7\%)}   &       \small{3248 (99.7\%)}   &       \small{5881 (99.3\%)}   &       \small{3206 (99.6\%)}   \\
	\rotatebox{90}{\textsc{low-res}}	&\includegraphics[width=0.15\linewidth]{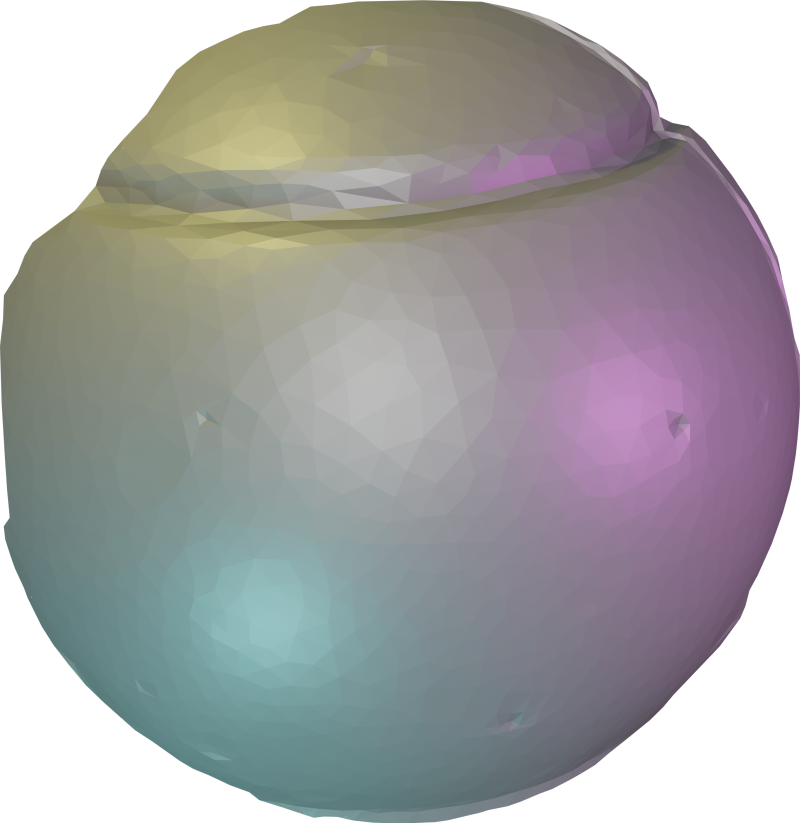}
        &\includegraphics[width=0.15\linewidth]{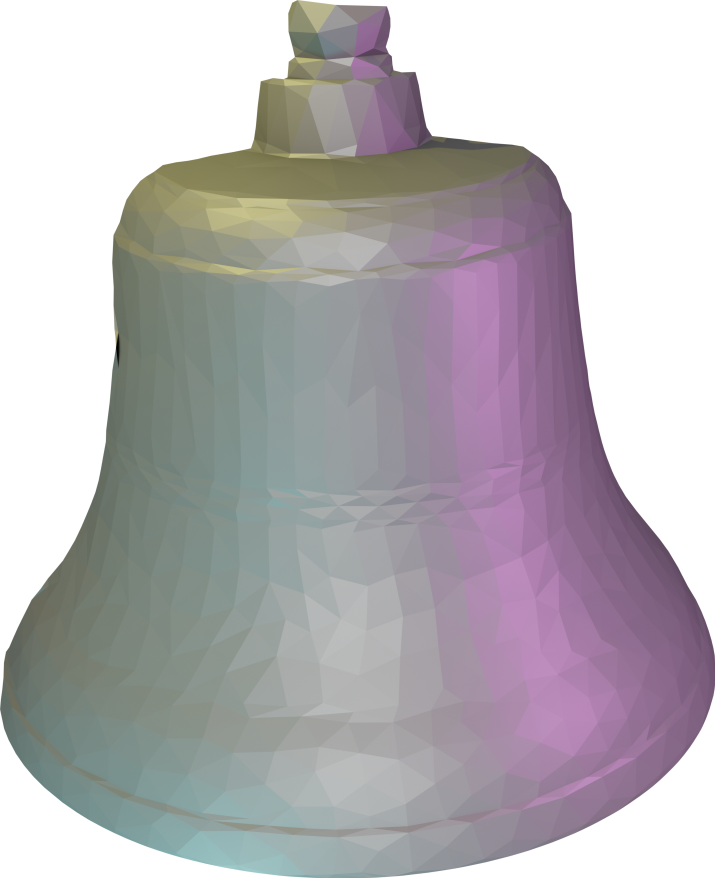}
	&\includegraphics[width=0.17\linewidth]{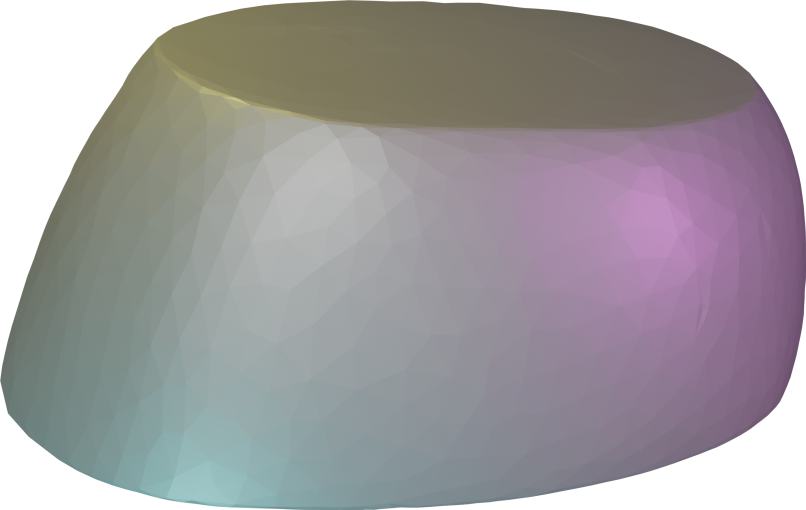}
        &\includegraphics[width=0.17\linewidth]{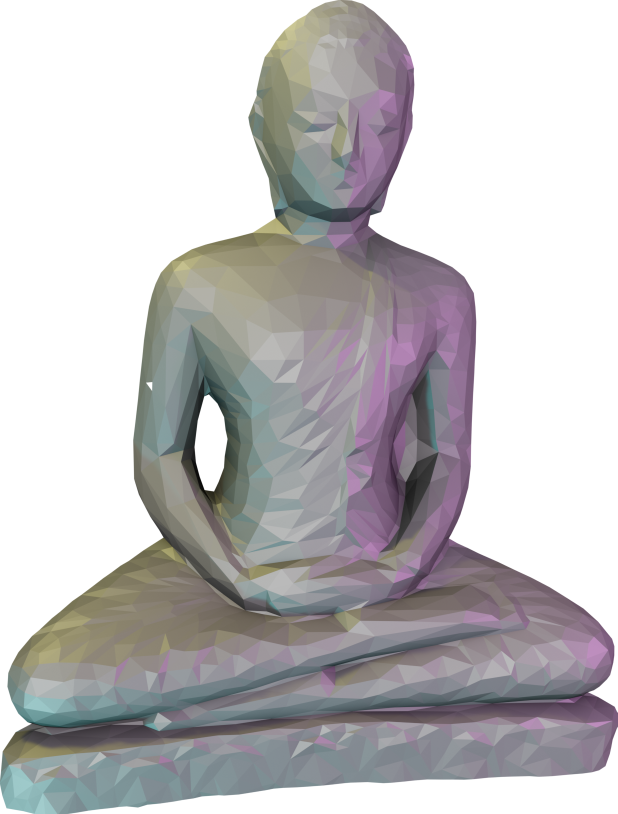}
        &\includegraphics[width=0.17\linewidth]{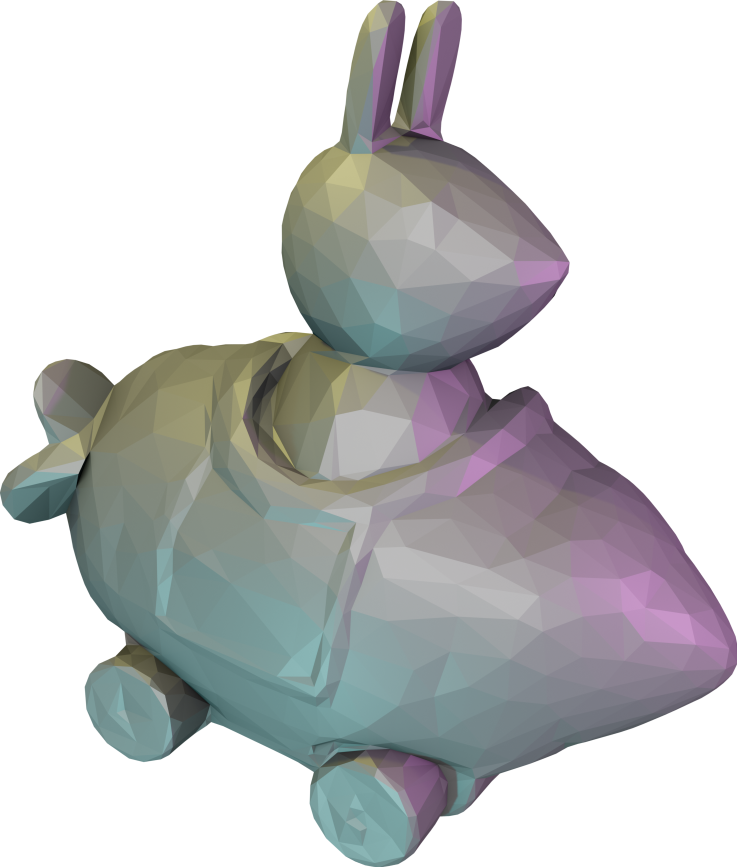} \\
        &       \small{1199 (99.9\%)}   &       \small{799 (99.9\%)}    &       \small{634 (99.9\%)}    &       \small{1175 (99.9\%)}   &       \small{664 (99.9\%)}    \\
	\end{tabular}
    \caption{Reconstruction results for the LUCES dataset \cite{Mecca:2021} for decimation thresholds of 2, 64 and 2048. These reconstructions correspond to the numbers reported in Tab.\ 2 of the manuscript. Any holes in the mesh surface are part of the provided foreground mask.}
    \label{fig:LUCES_first_figure}
\end{figure*}

\begin{figure*}
    \centering
    \LARGE{LUCES Dataset (2 of 3)} \\ \vspace{1mm}
    \begin{tabular}{cccccc}
    &\textsc{{Cup}}&\textsc{{Die}} &\textsc{{Hippo}}&\textsc{{House}}&\textsc{{Jar}}\\ \vspace{1mm}
		\rotatebox{90}{\textsc{high-res}}&\includegraphics[width=0.13\linewidth]{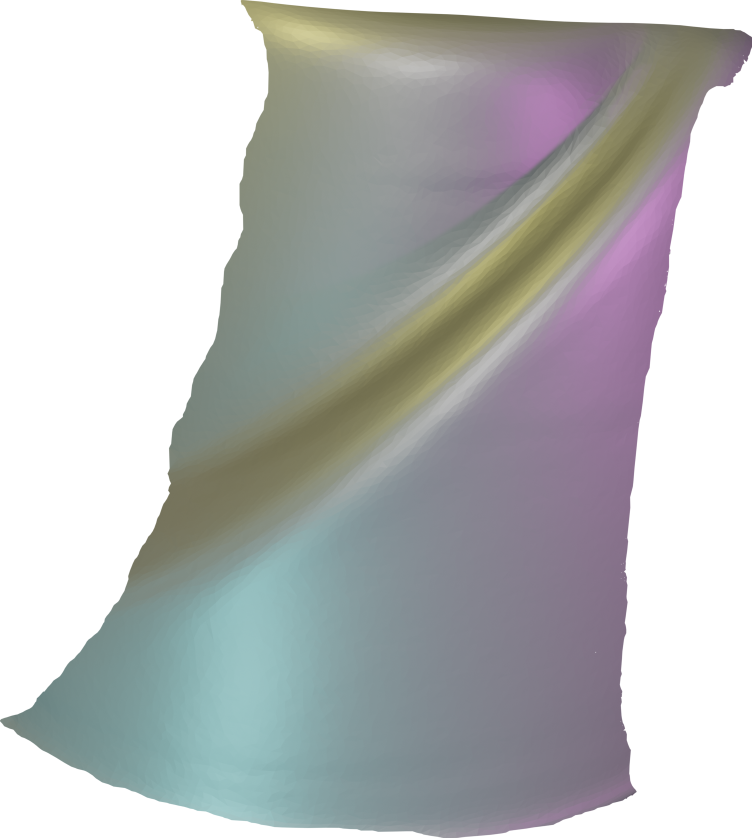}
        &\includegraphics[width=0.14\linewidth]{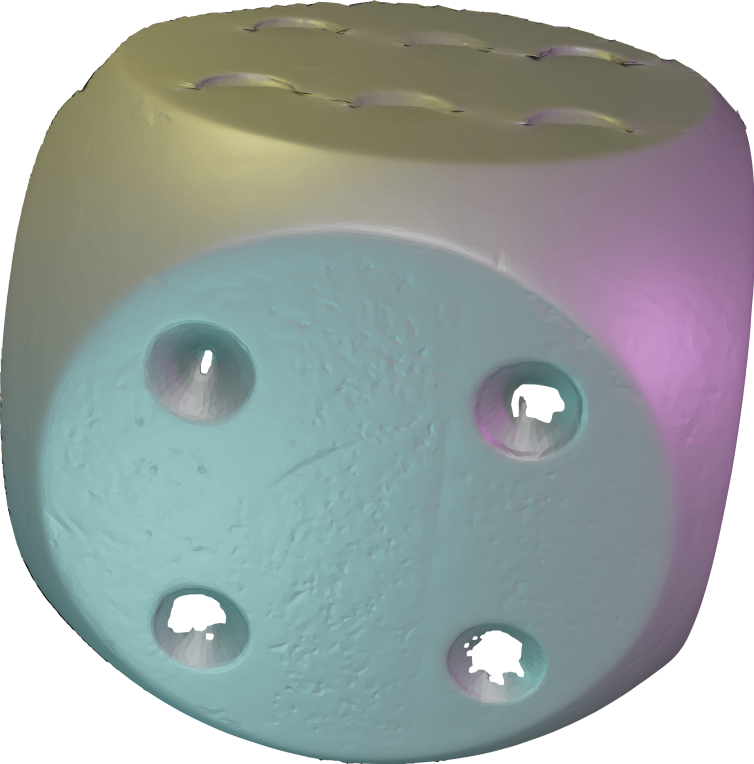}
		&\includegraphics[width=0.23\linewidth]{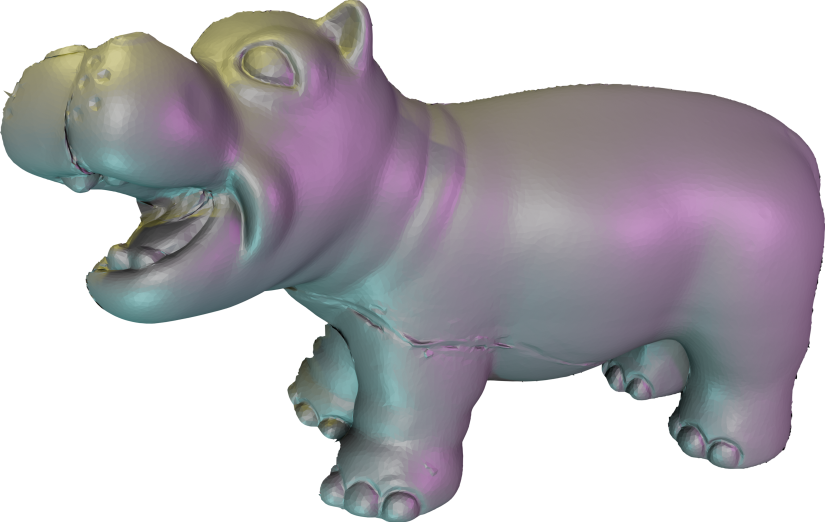}
        &\includegraphics[width=0.22\linewidth]{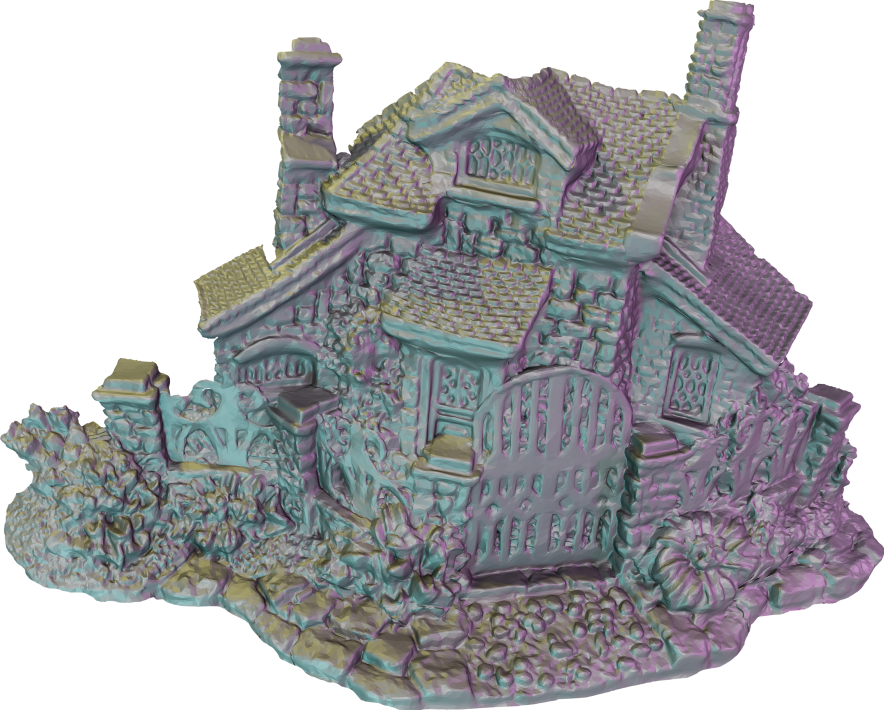} 
        &\includegraphics[width=0.14\linewidth]{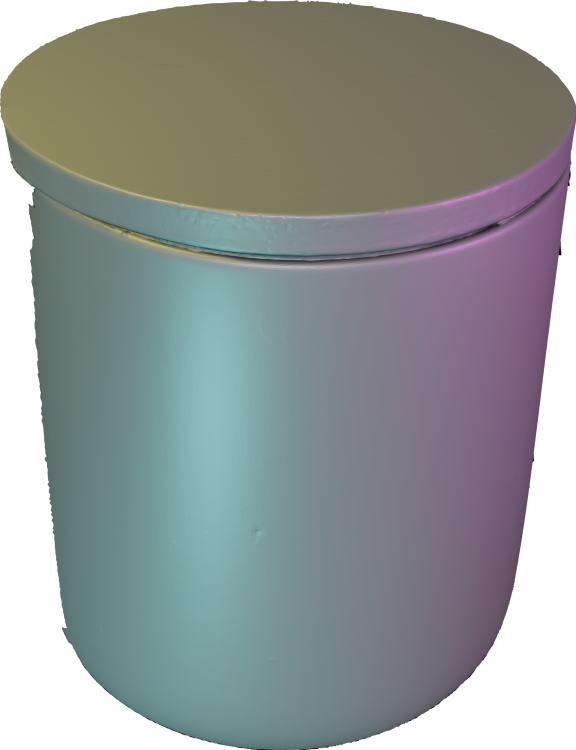} \\
        &       \small{7268 (98.8\%)}   &       \small{19391 (97.8\%)}  &       \small{15951 (97.9\%)}  &       \small{68216 (95.0\%)}  &       \small{28761 (97.9\%)}  \\
		&\includegraphics[width=0.13\linewidth]{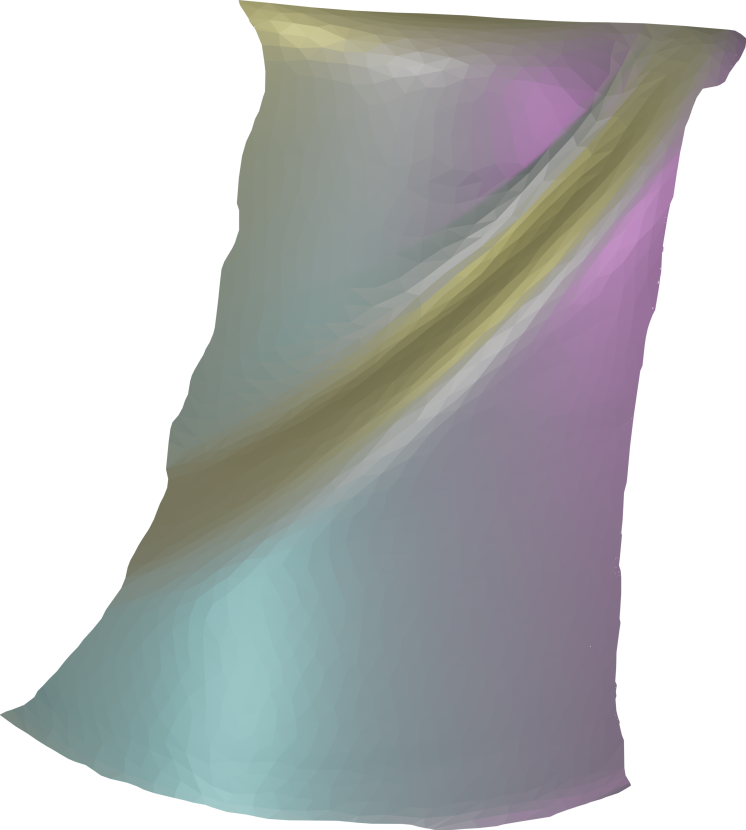}
		&\includegraphics[width=0.14\linewidth]{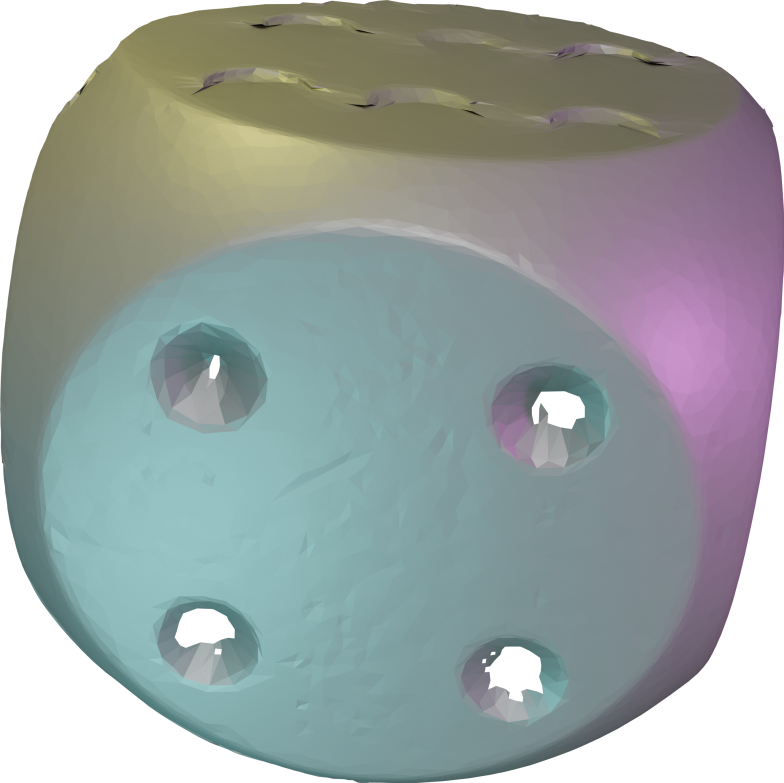}
        &\includegraphics[width=0.23\linewidth]{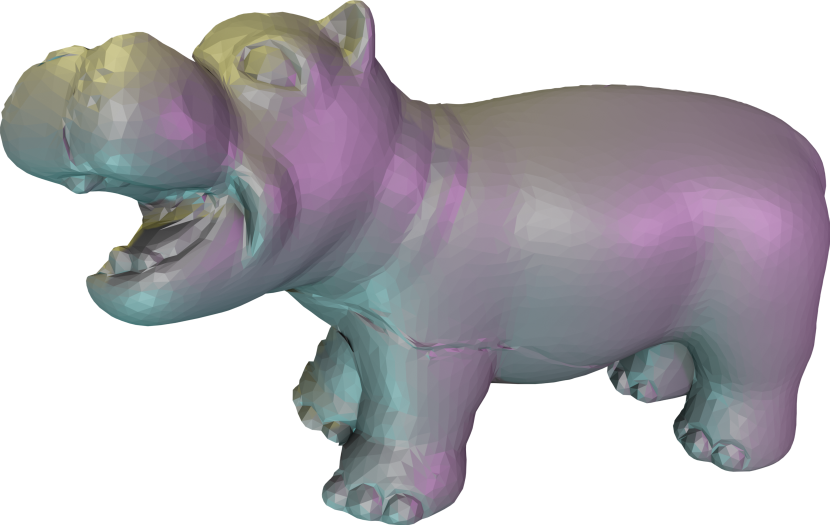}
        &\includegraphics[width=0.22\linewidth]{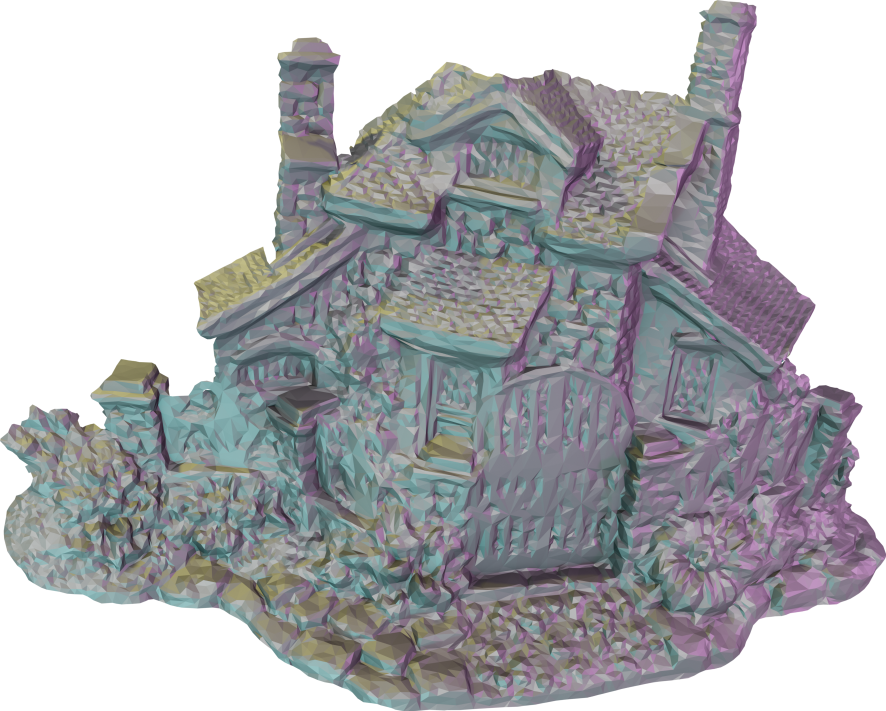}
        &\includegraphics[width=0.14\linewidth]{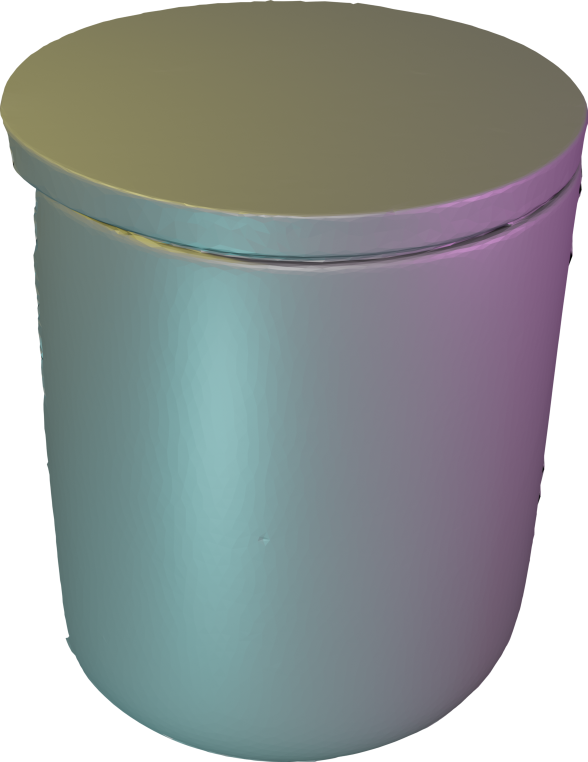}\\
        &       \small{1426 (99.8\%)}   &       \small{3472 (99.6\%)}   &       \small{3110 (99.6\%)}   &       \small{13520 (99.0\%)}  &       \small{5180 (99.6\%)}   \\
	\rotatebox{90}{\textsc{low-res}}	&\includegraphics[width=0.13\linewidth]{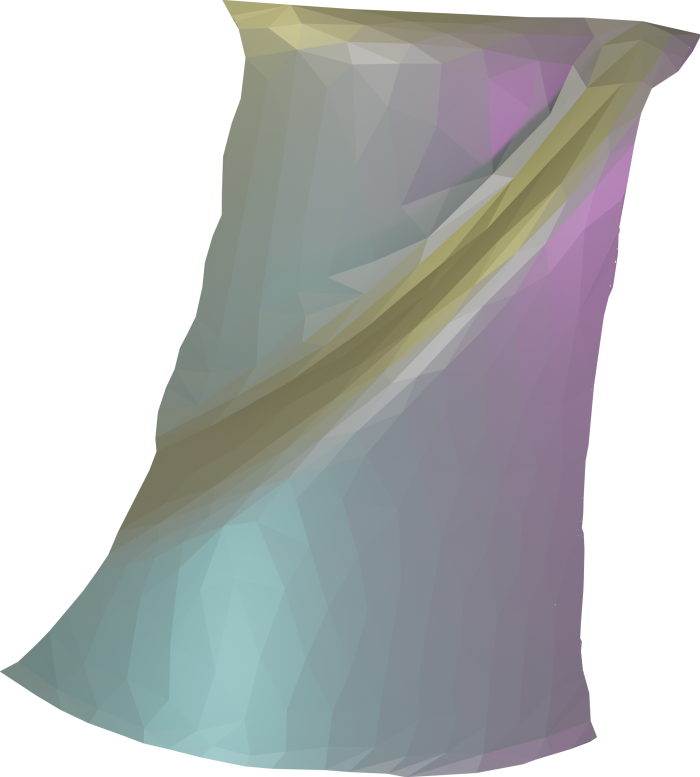}
        &\includegraphics[width=0.14\linewidth]{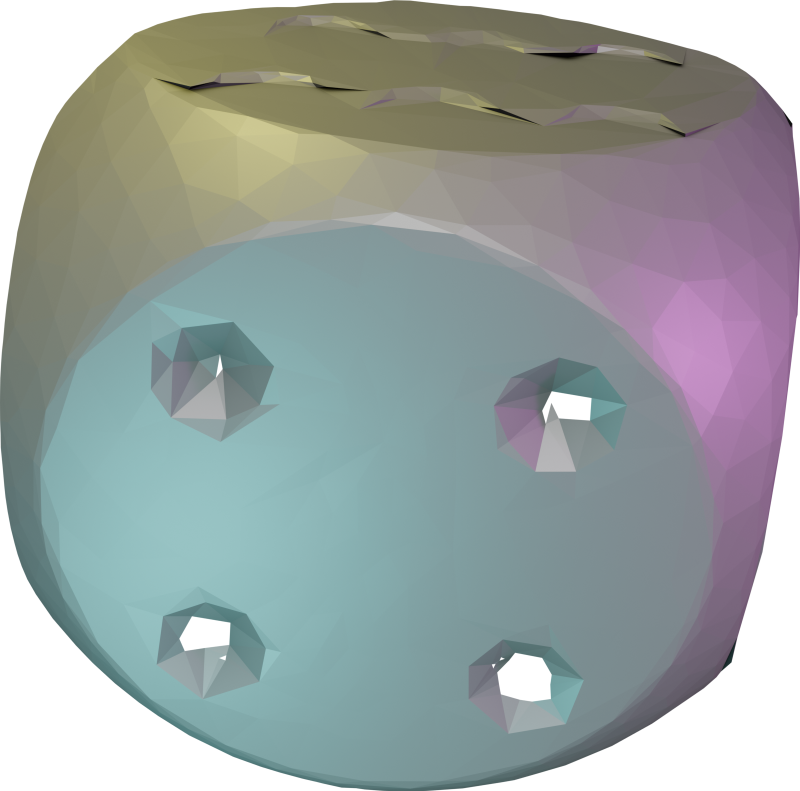}
		&\includegraphics[width=0.23\linewidth]{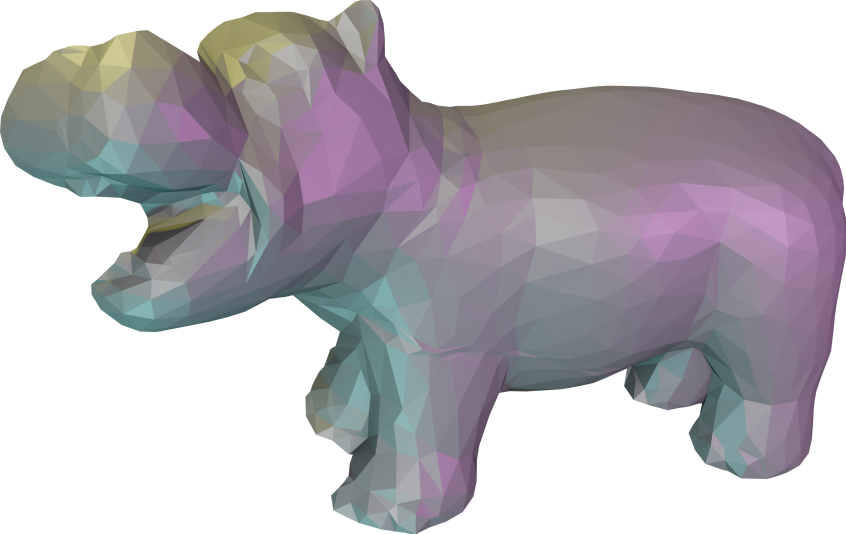}
        &\includegraphics[width=0.22\linewidth]{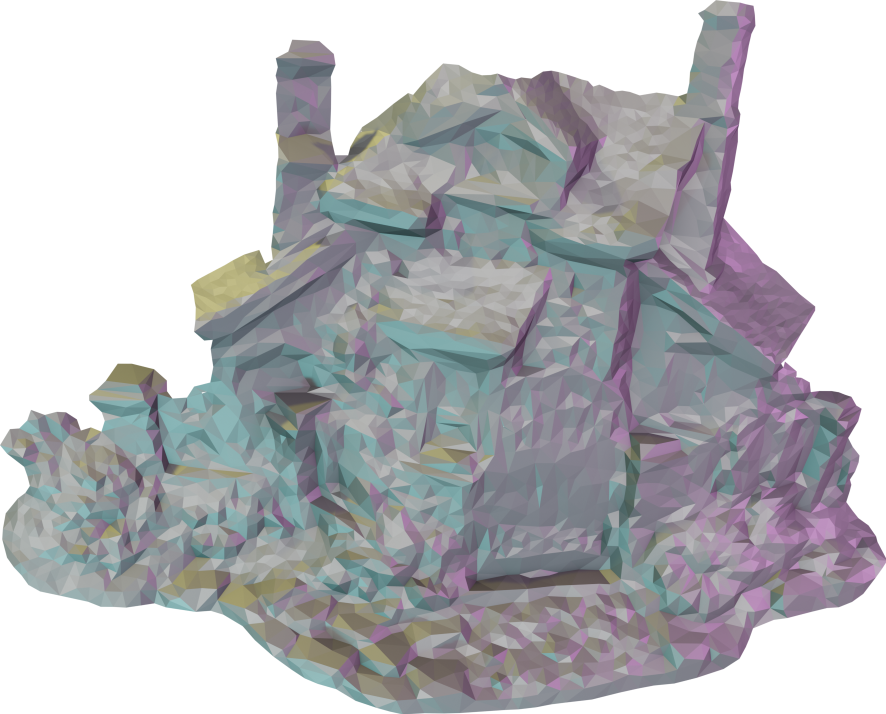}
        &\includegraphics[width=0.14\linewidth]{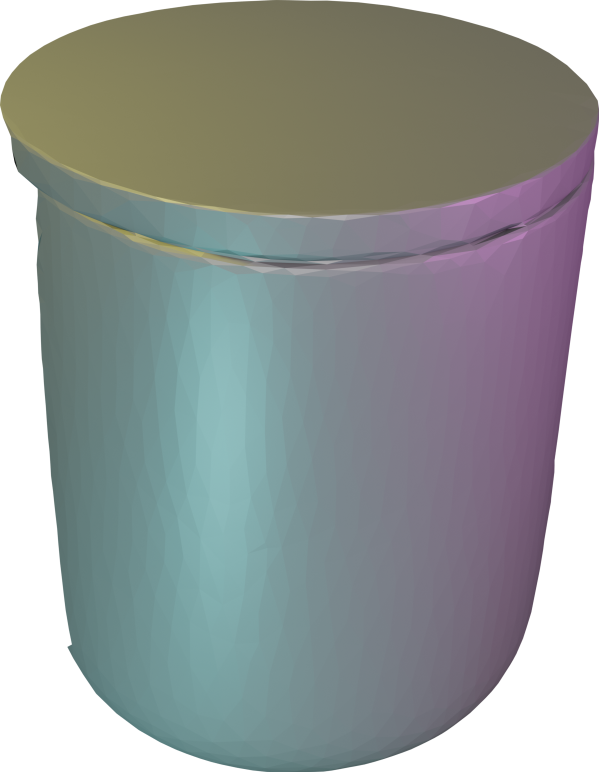} \\
        &       \small{343 (99.9\%)}    &       \small{692 (99.9\%)}    &       \small{631 (99.9\%)}    &       \small{3043 (99.8\%)}   &       \small{957 (99.9\%)}    \\
    \end{tabular}
    \caption{Reconstruction results for the LUCES dataset \cite{Mecca:2021} for decimation thresholds of 2, 64 and 2048. These reconstructions correspond to the numbers reported in Tab.\ 2 of the main paper. Any holes in the mesh surface are part of the provided foreground mask.}
    \label{fig:LUCES_second_figure}
\end{figure*}

\begin{figure*}
    \centering
    \LARGE{LUCES Dataset (3 of 3)} \\ \vspace{3mm}
    \begin{tabular}{ccccc}
    &\textsc{{Owl}}&\textsc{{Queen}} &\textsc{{Squirrel}}&\textsc{{Tool}}\\ \vspace{1mm}
		\rotatebox{90}{\textsc{high-res}}
        &\includegraphics[width=0.25\linewidth]{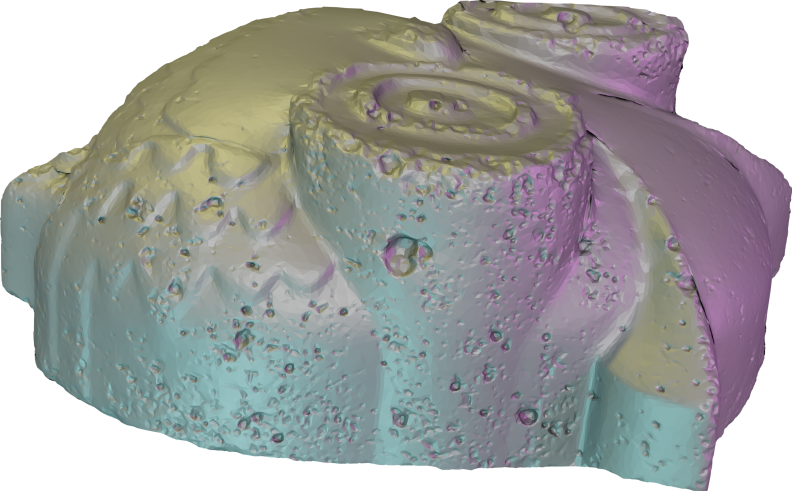}
        &\includegraphics[width=0.15\linewidth]{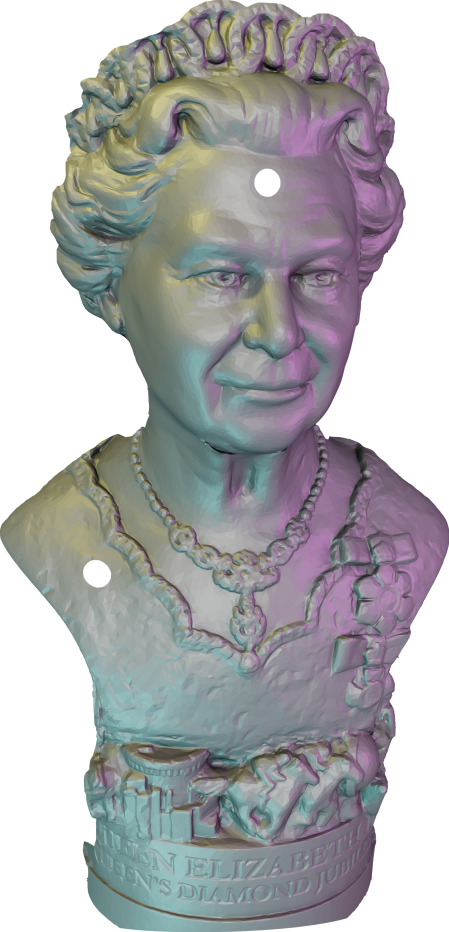}
		&\includegraphics[width=0.24\linewidth]{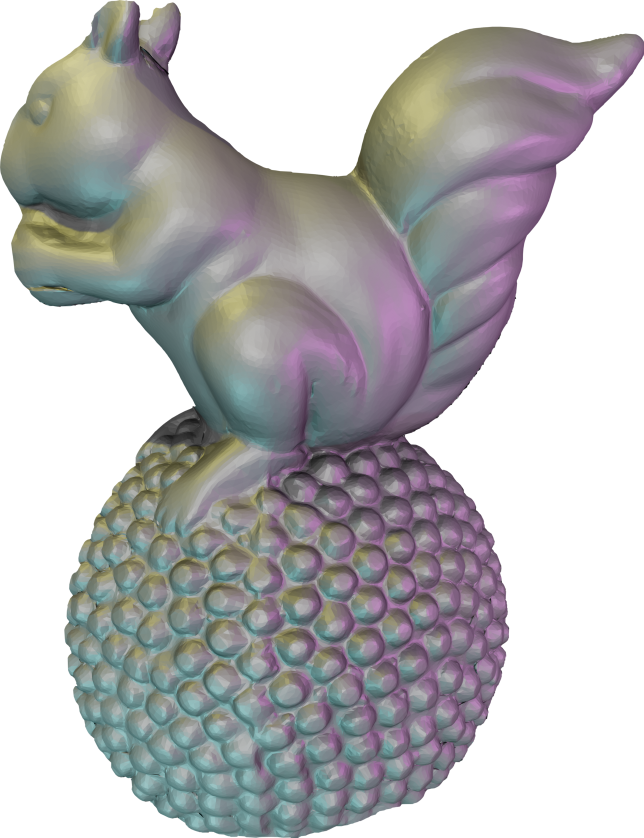}
        &\includegraphics[width=0.07\linewidth]{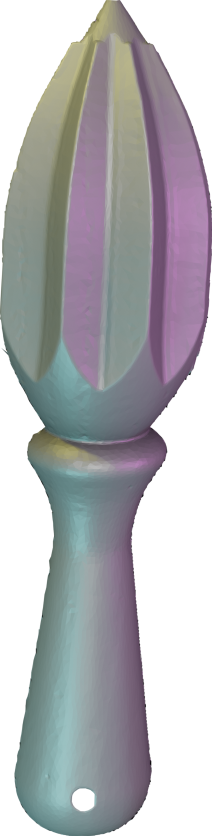} \\
        &       \small{25233 (96.3\%)}  &       \small{24315 (96.7\%)}  &       \small{25777 (97.3\%)}  &       \small{7391 (97.9\%)}   \\
		&\includegraphics[width=0.25\linewidth]{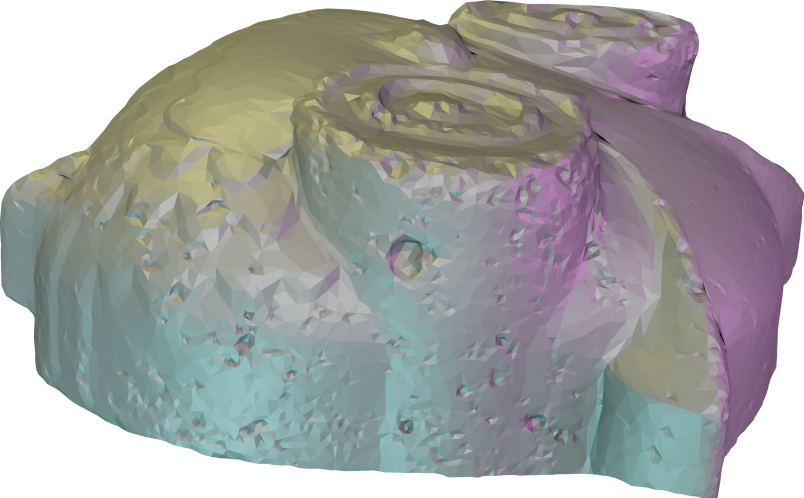}
		&\includegraphics[width=0.15\linewidth]{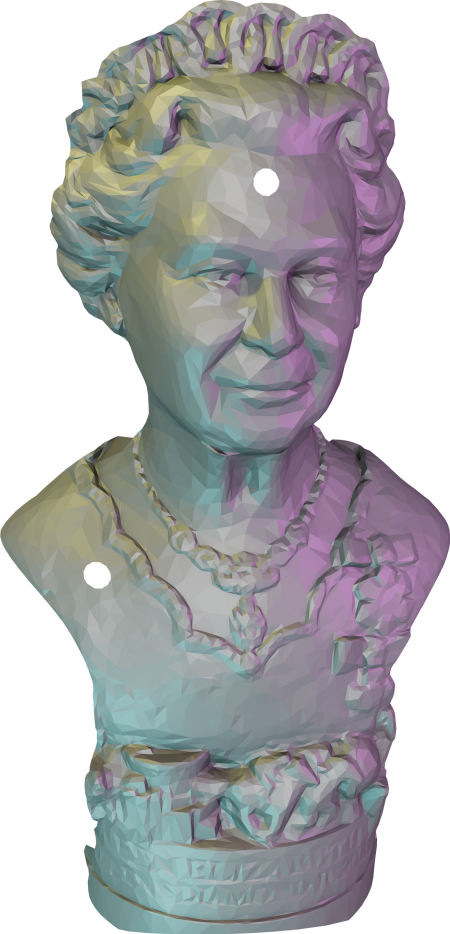}
        &\includegraphics[width=0.24\linewidth]{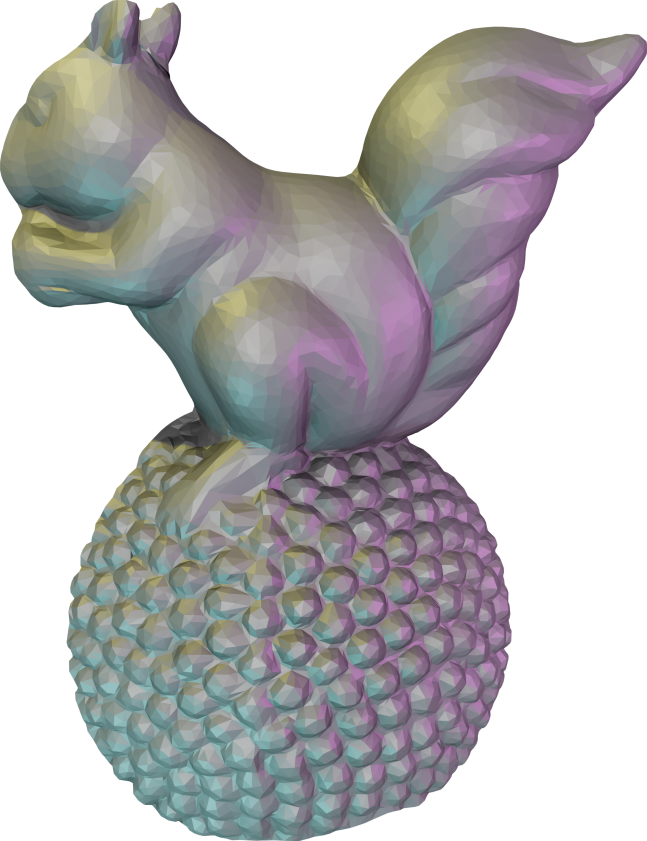}
        &\includegraphics[width=0.07\linewidth]{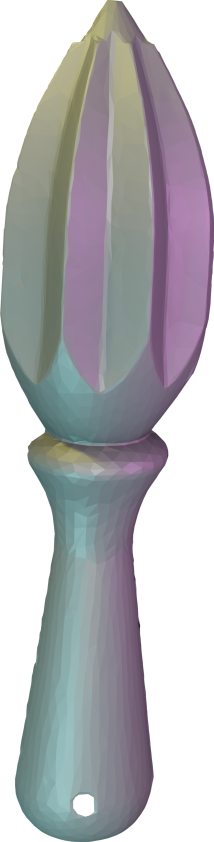}\\
        &       \small{5019 (99.3\%)}   &       \small{4954 (99.3\%)}   &       \small{4941 (99.5\%)}   &       \small{1559 (99.6\%)}   \\
		\rotatebox{90}{\textsc{low-res}}
        &\includegraphics[width=0.25\linewidth]{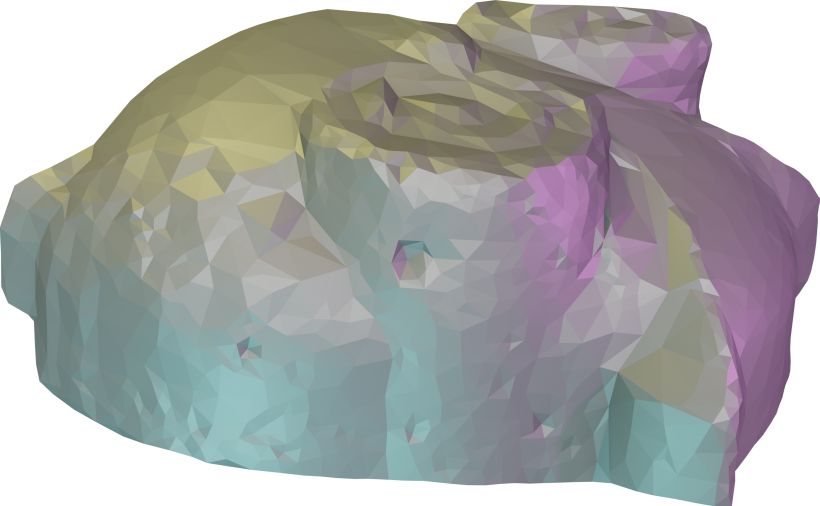}
        &\includegraphics[width=0.15\linewidth]{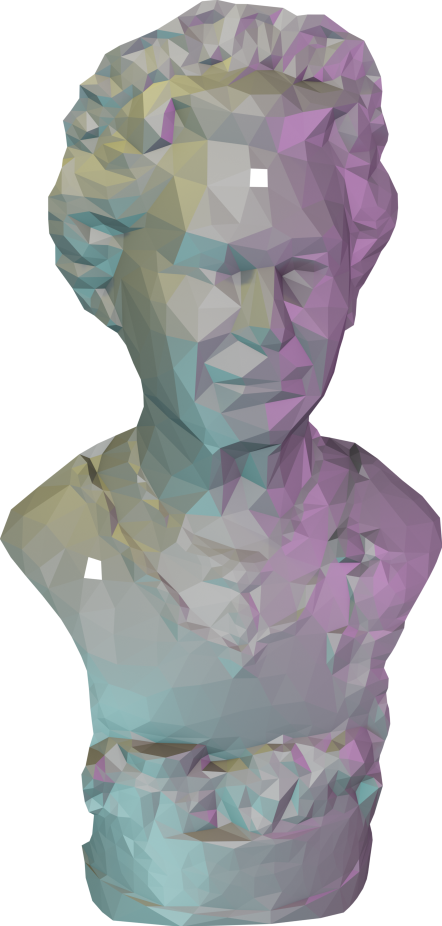}
		&\includegraphics[width=0.24\linewidth]{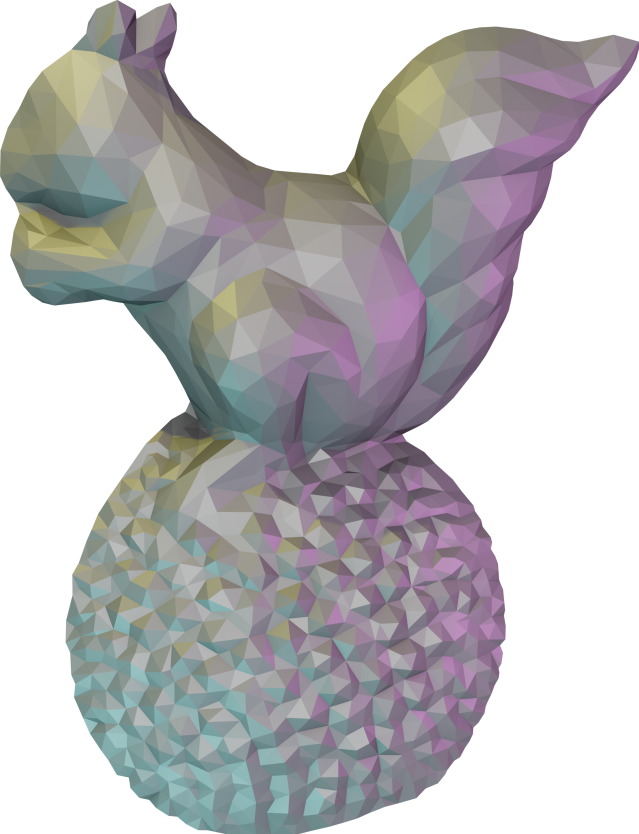}
        &\includegraphics[width=0.07\linewidth]{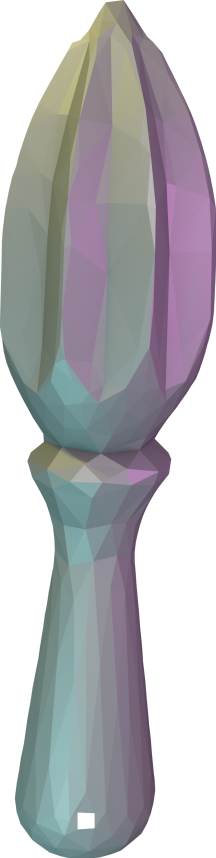} \\
        &       \small{1071 (99.8\%)}   &       \small{1081 (99.9\%)}   &       \small{997 (99.9\%)}    &       \small{344 (99.9\%)}    \\
    \end{tabular}
    \caption{Reconstruction results for the LUCES dataset \cite{Mecca:2021} for decimation thresholds of 2, 64 and 2048. These reconstructions correspond to the numbers reported in Tab.\ 2 of the main paper. Any holes in the mesh surface are part of the provided foreground mask.}
    \label{fig:LUCES_third_figure}
\end{figure*}

\begin{figure*}
    \centering
    \LARGE{RGBN Dataset (1 of 2)} \\
    \Large{High-Res} \hfill \large{Low-Res}\\
    \vspace{-3mm}
    \hrulefill
    \\    
    \LARGE{\textsc{Chard}} \\ \vspace{1mm}
    \begin{subfigure}[b]{.22\linewidth}
        \centering
        \includegraphics[width=\linewidth]{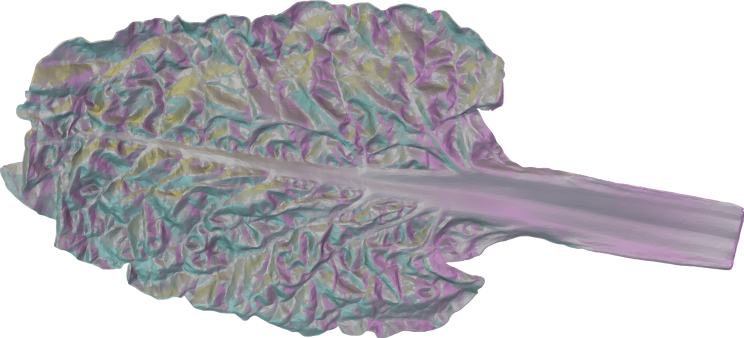}  \\
        \small{28812 (98.4\%)}
    \end{subfigure}
    \begin{subfigure}[b]{.22\linewidth}
        \centering
        \includegraphics[width=\linewidth]{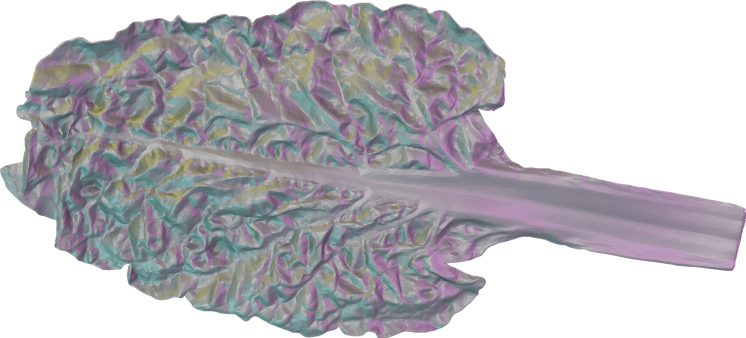}  \\
        \small{20816 (98.9\%)}
    \end{subfigure}
    \begin{subfigure}[b]{.22\linewidth}
        \centering
        \includegraphics[width=\linewidth]{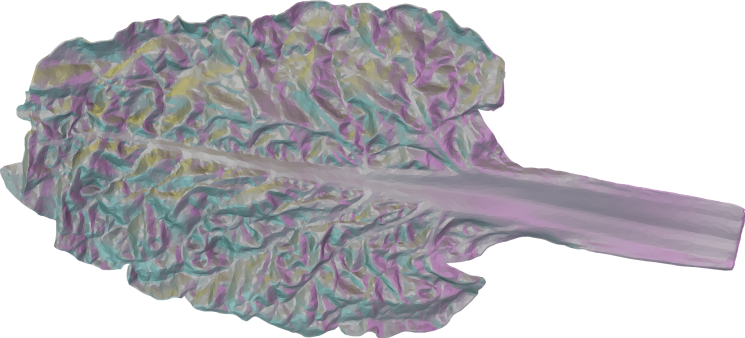}  \\
        \small{15088 (99.2\%)}
    \end{subfigure}
    \begin{subfigure}[b]{.22\linewidth}
        \centering
        \includegraphics[width=\linewidth]{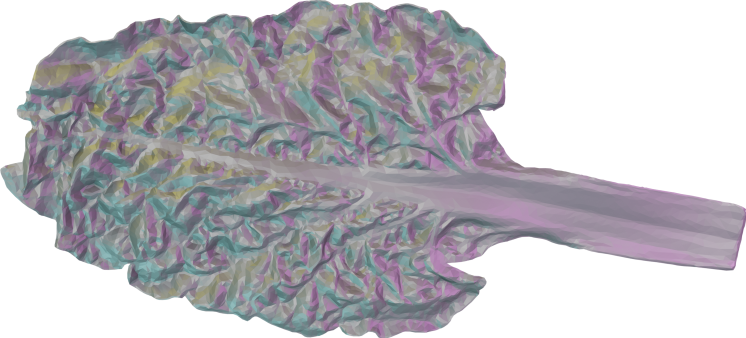}  \\
        \small{10906 (99.4\%)}
    \end{subfigure}
    \\
    \LARGE{\textsc{Chard2}} \\ \vspace{1mm}
    \begin{subfigure}[b]{.22\linewidth}
        \centering
        \includegraphics[width=\linewidth]{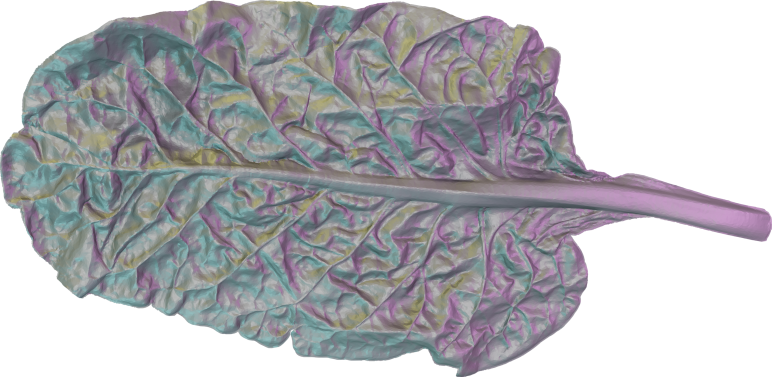}  \\
        \small{59857 (98.2\%)}
    \end{subfigure}
    \begin{subfigure}[b]{.22\linewidth}
        \centering
        \includegraphics[width=\linewidth]{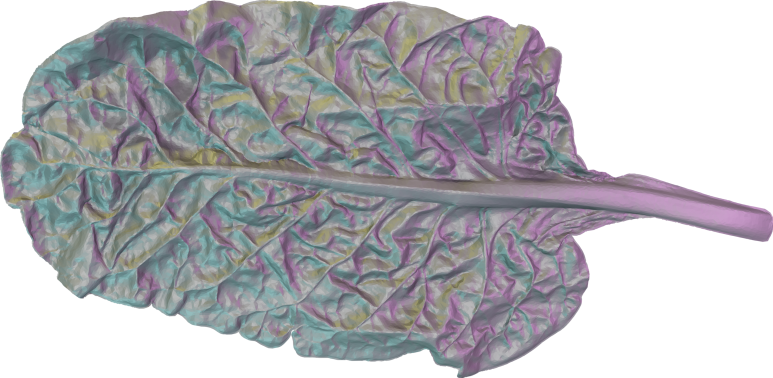}  \\
        \small{43442 (98.7\%)}
    \end{subfigure}
    \begin{subfigure}[b]{.22\linewidth}
        \centering
        \includegraphics[width=\linewidth]{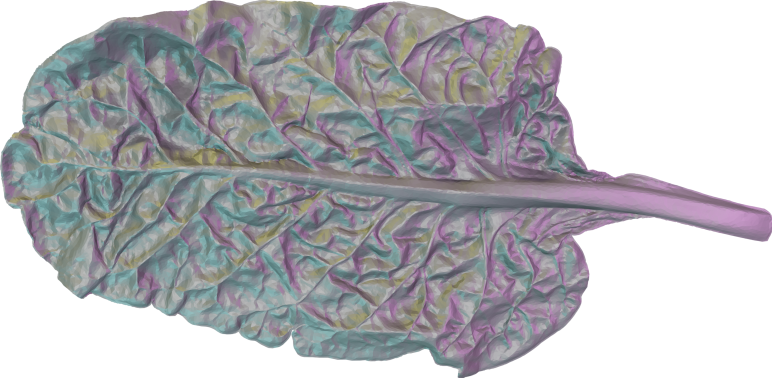}  \\
        \small{31312 (99.1\%)}
    \end{subfigure}
    \begin{subfigure}[b]{.22\linewidth}
        \centering
        \includegraphics[width=\linewidth]{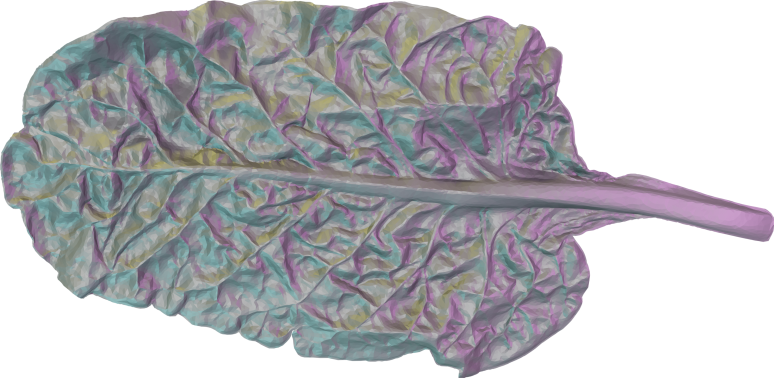}  \\
        \small{22710 (99.3\%)}
    \end{subfigure}
    \\
    \LARGE{\textsc{Food}} \\ \vspace{1mm}
    \begin{subfigure}[b]{.22\linewidth}
        \centering
        \includegraphics[width=\linewidth]{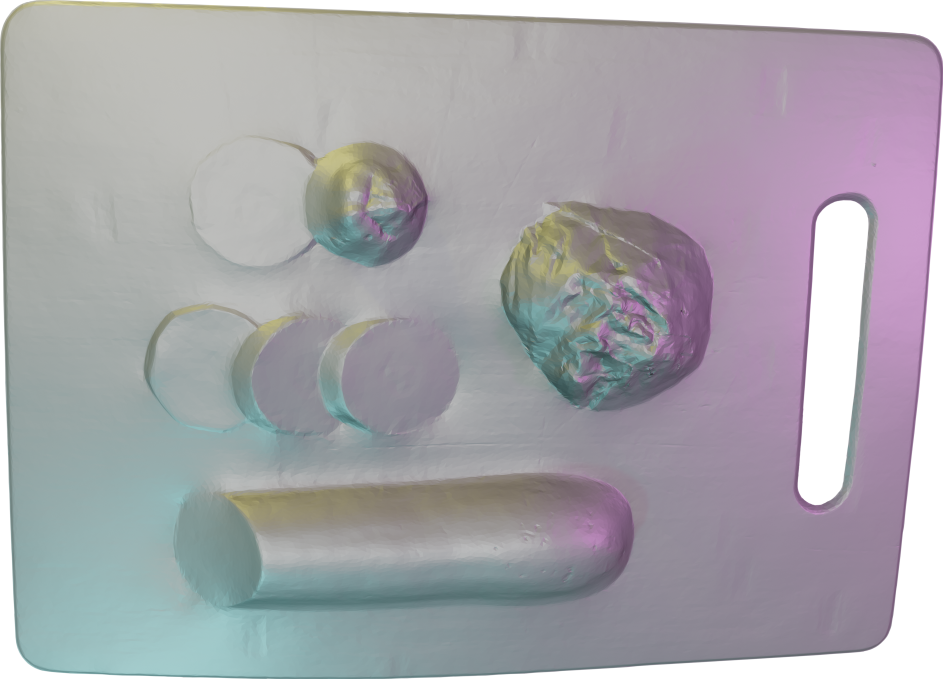}  \\
        \small{35258 (99.2\%)}
    \end{subfigure}
    \begin{subfigure}[b]{.22\linewidth}
        \centering
        \includegraphics[width=\linewidth]{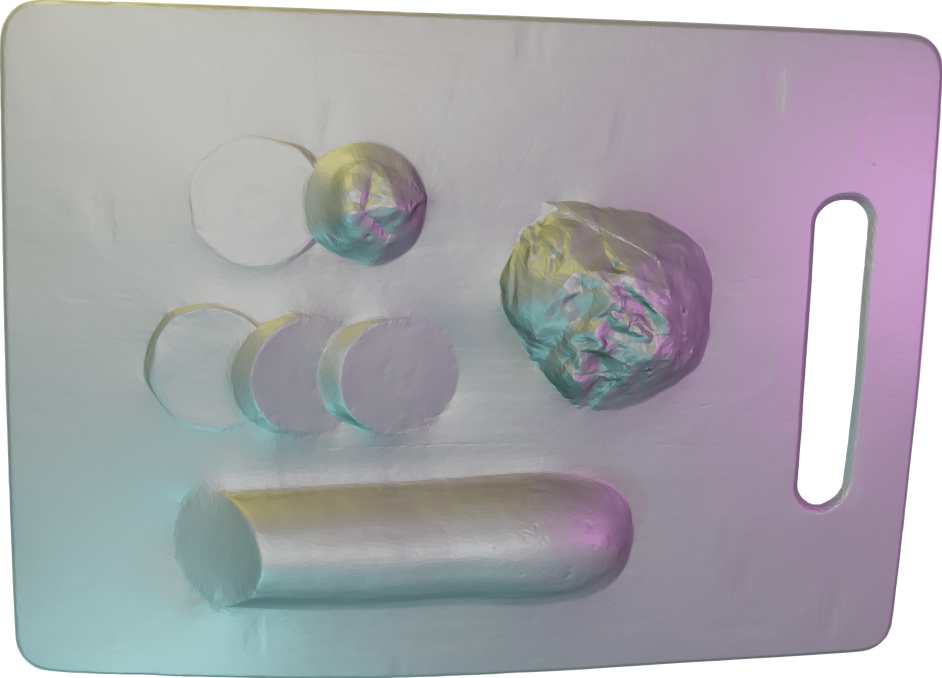}  \\
        \small{25215 (99.4\%)}
    \end{subfigure}
    \begin{subfigure}[b]{.22\linewidth}
        \centering
        \includegraphics[width=\linewidth]{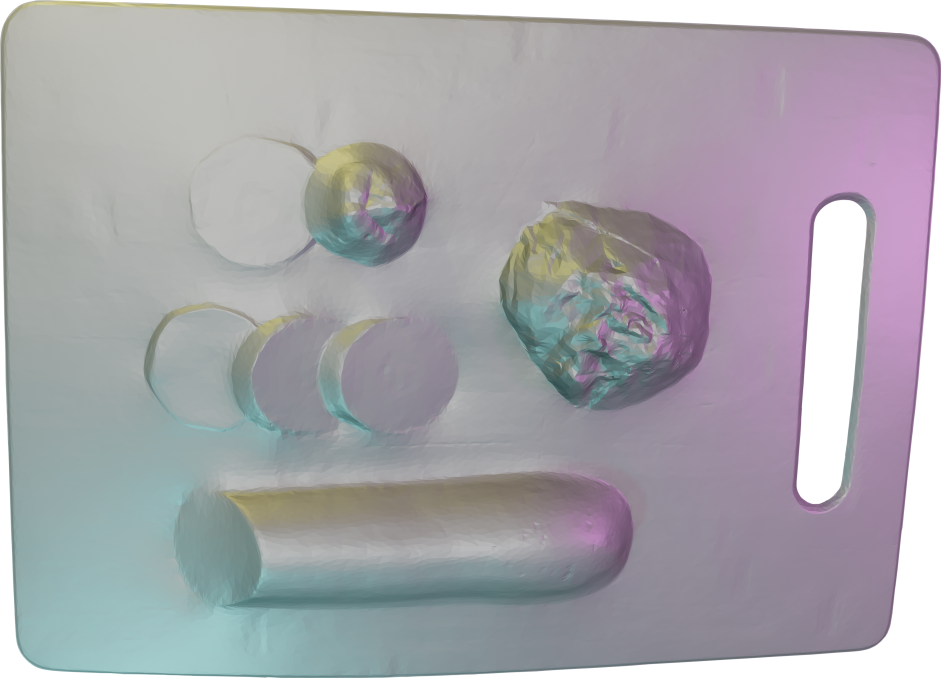}  \\
        \small{18070 (99.6\%)}
    \end{subfigure}
    \begin{subfigure}[b]{.22\linewidth}
        \centering
        \includegraphics[width=\linewidth]{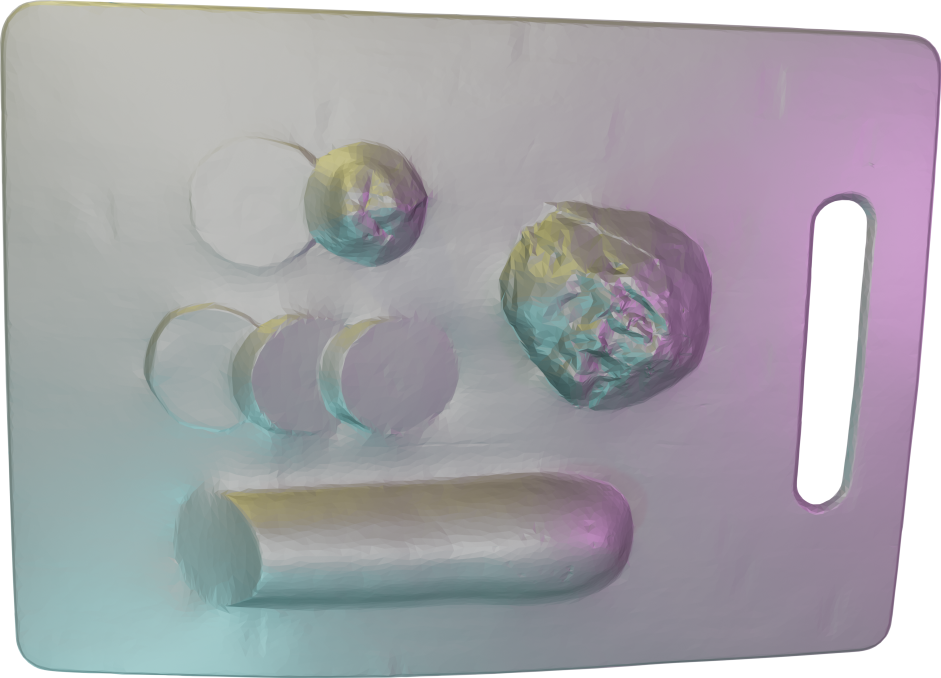}  \\
        \small{12965 (99.7\%)}
    \end{subfigure}
    \\
    \LARGE{\textsc{Leaves}} \\ \vspace{1mm}
    \begin{subfigure}[b]{.22\linewidth}
        \centering
        \includegraphics[width=\linewidth]{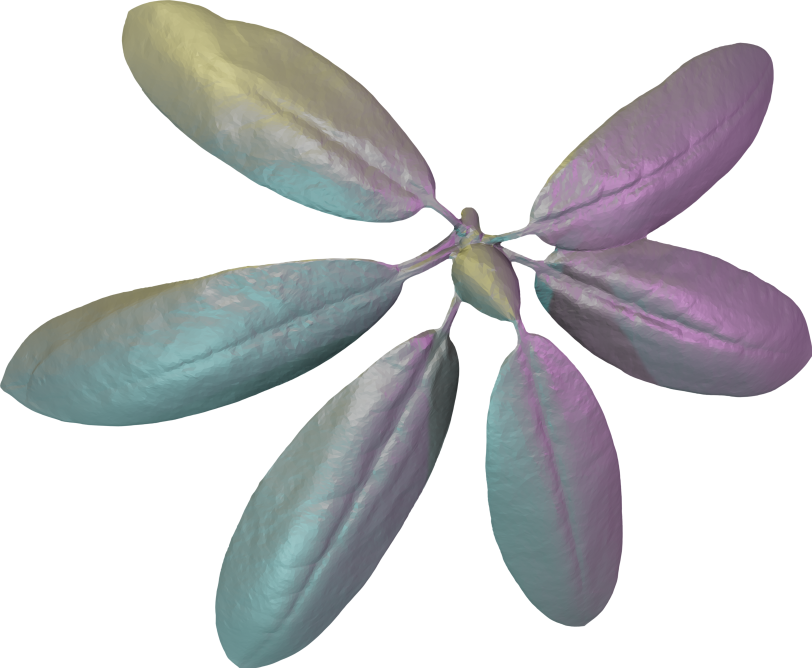}  \\
        \small{29163 (98.3\%)}
    \end{subfigure}
    \begin{subfigure}[b]{.22\linewidth}
        \centering
        \includegraphics[width=\linewidth]{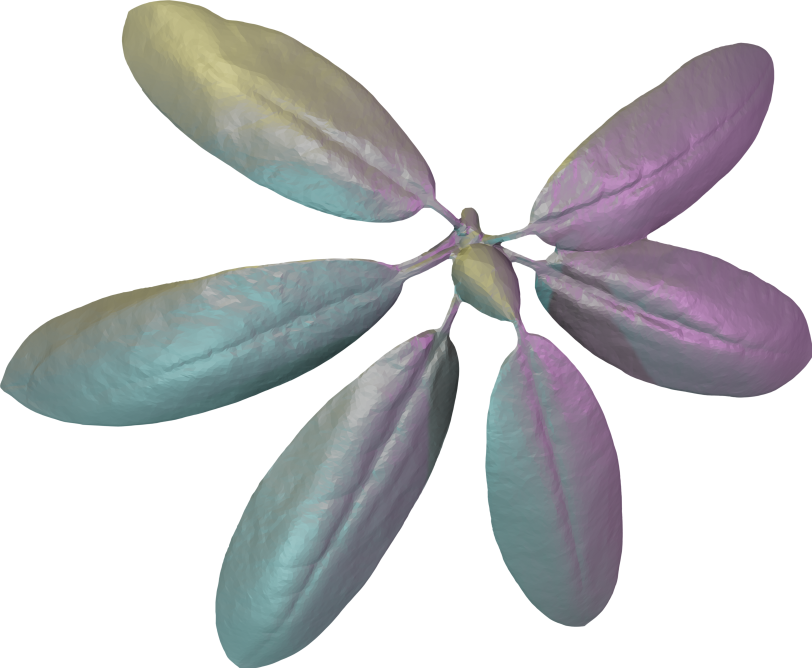}  \\
        \small{21159 (98.8\%)}
    \end{subfigure}
    \begin{subfigure}[b]{.22\linewidth}
        \centering
        \includegraphics[width=\linewidth]{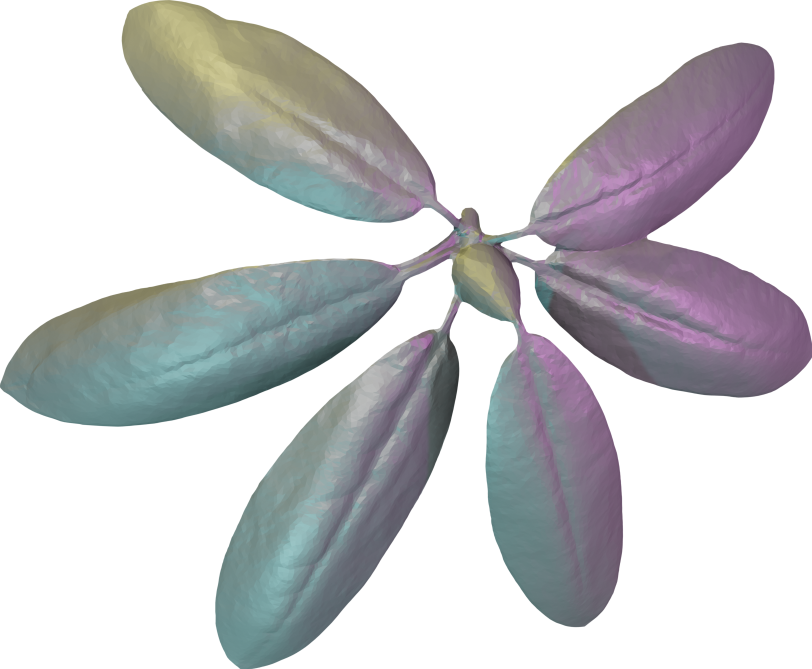}  \\
        \small{15387 (99.1\%)}
    \end{subfigure}
    \begin{subfigure}[b]{.22\linewidth}
        \centering
        \includegraphics[width=\linewidth]{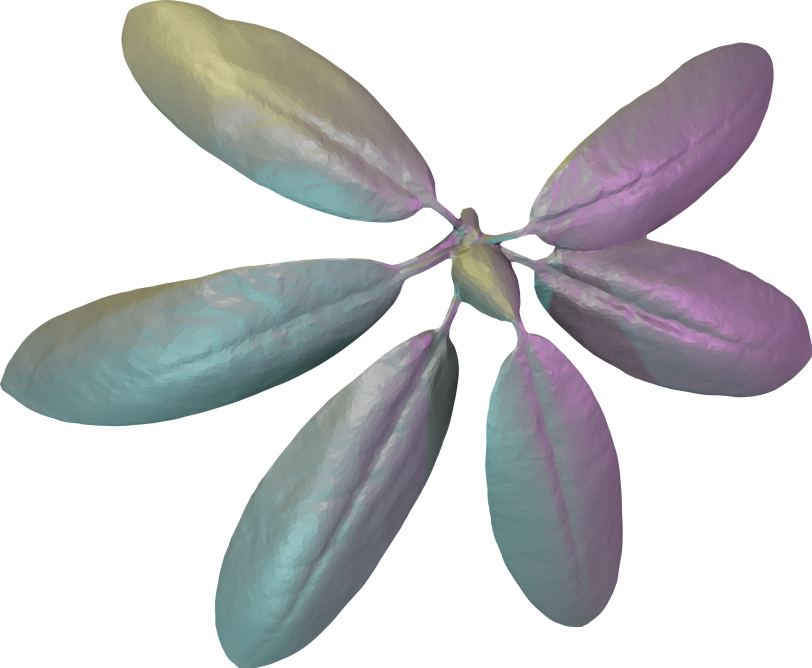}  \\
        \small{11076 (99.4\%)}
    \end{subfigure}
    \\
    \LARGE{\textsc{Shell}} \\ \vspace{1mm}
    \begin{subfigure}[b]{.22\linewidth}
        \centering
        \includegraphics[width=\linewidth]{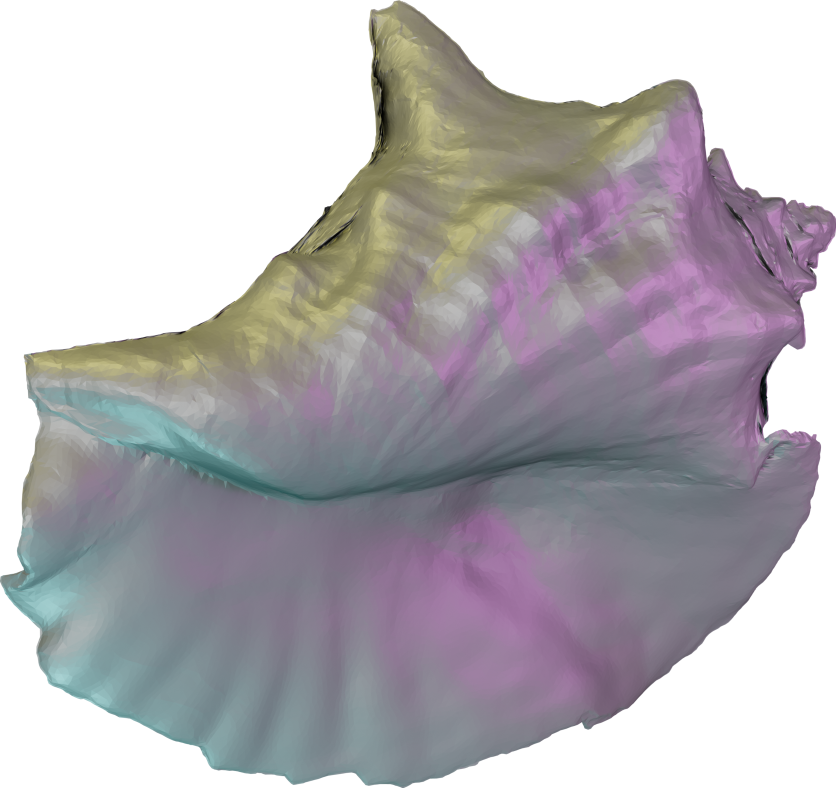}  \\
        \small{21961 (99.1\%)}
    \end{subfigure}
    \begin{subfigure}[b]{.22\linewidth}
        \centering
        \includegraphics[width=\linewidth]{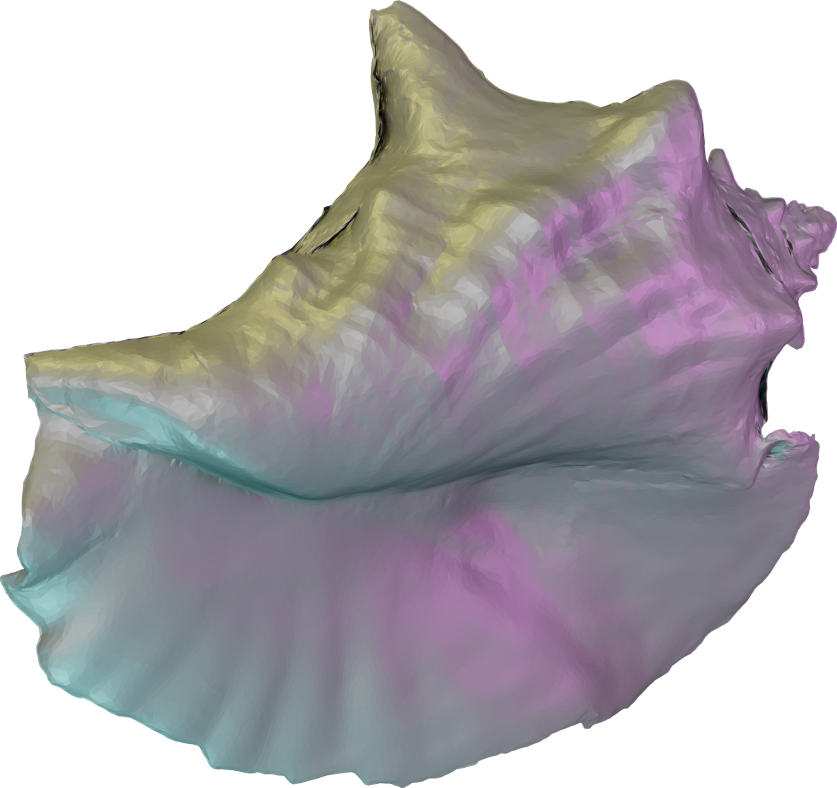}  \\
        \small{15796 (99.3\%)}
    \end{subfigure}
    \begin{subfigure}[b]{.22\linewidth}
        \centering
        \includegraphics[width=\linewidth]{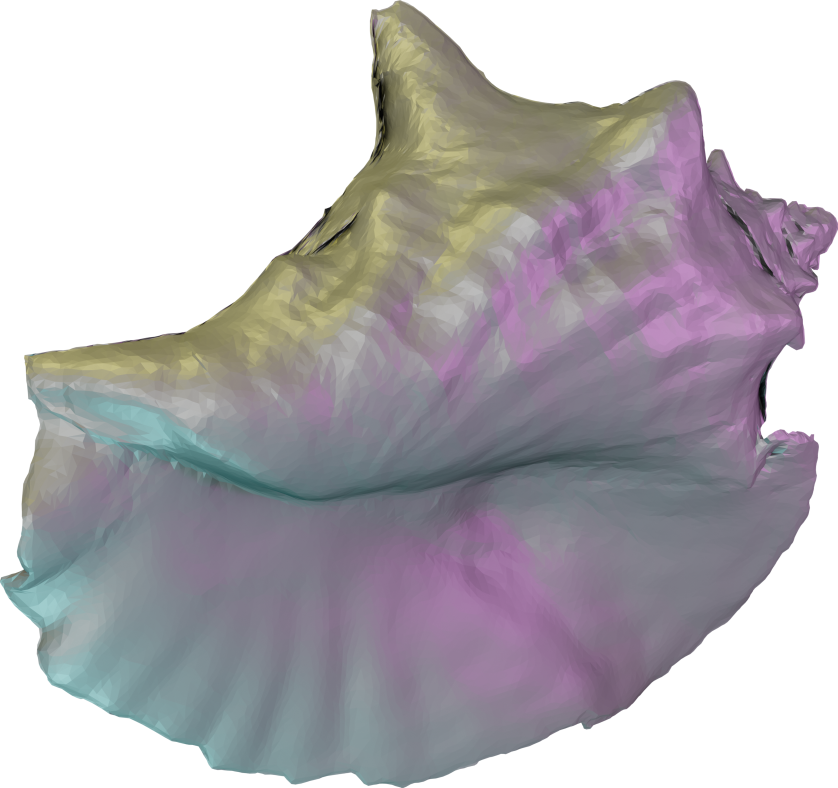}  \\
        \small{11296 (99.5\%)}
    \end{subfigure}
    \begin{subfigure}[b]{.22\linewidth}
        \centering
        \includegraphics[width=\linewidth]{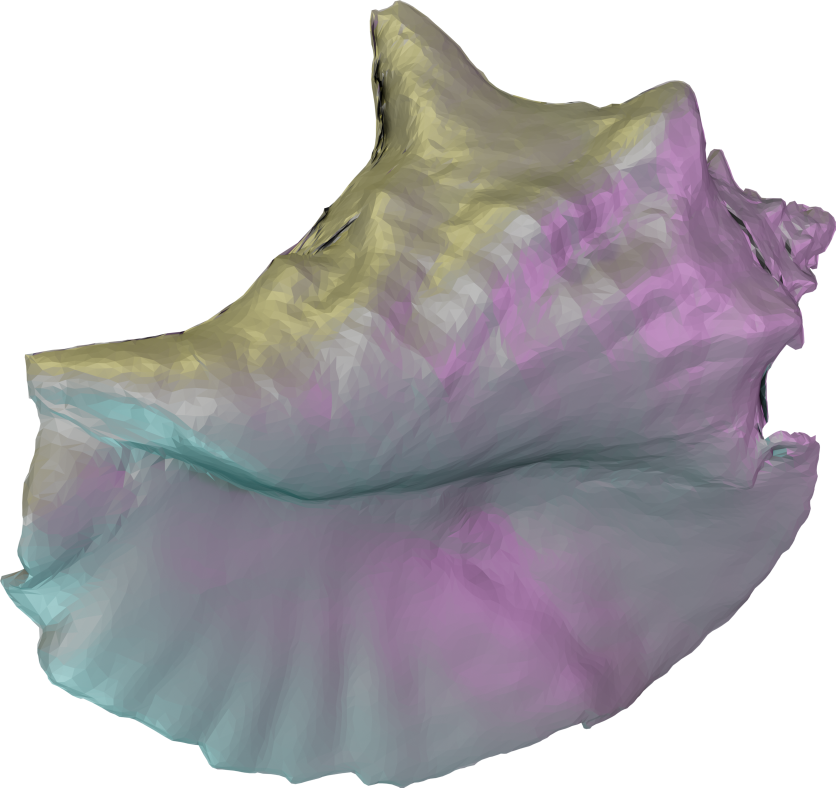}  \\
        \small{8083 (99.7\%)}
    \end{subfigure}
    \\
    \caption{Reconstruction results for the RGBN dataset \cite{Toler-Franklin:2007} for decimation thresholds of 8, 16, 32 and 64.}
    \label{fig:RGBN_first_figure}
\end{figure*}

\begin{figure*}
    \centering
    \LARGE{RGBN Dataset (2 of 2)} \\ \vspace{1mm}
    \Large{High-Res} \hfill \large{Low-Res}\\
    \vspace{-3mm}
    \hrulefill
    \\ \vspace{2mm}   
    \begin{subfigure}[b]{.8\linewidth}
        \centering
        \LARGE{\textsc{Pinecone3}} \\ \vspace{1mm}
        \begin{subfigure}[b]{.23\linewidth}
            \centering
            \includegraphics[angle=90,origin=c,height=40mm]{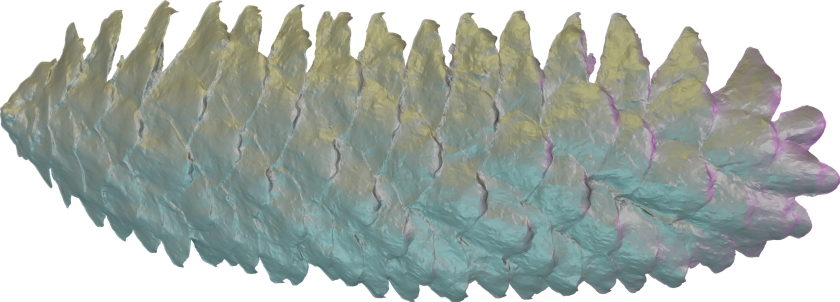}  \\
            \small{39834 (98.1\%)}
        \end{subfigure}
        \begin{subfigure}[b]{.23\linewidth}
            \centering
            \includegraphics[angle=90,origin=c,height=40mm]{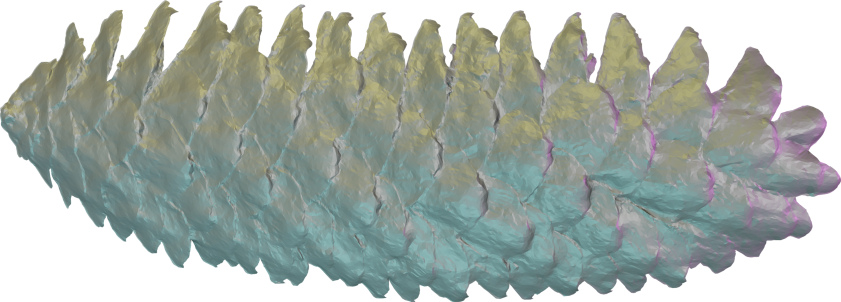}  \\
            \small{28919 (98.6\%)}
        \end{subfigure}
        \begin{subfigure}[b]{.23\linewidth}
            \centering
            \includegraphics[angle=90,origin=c,height=40mm]{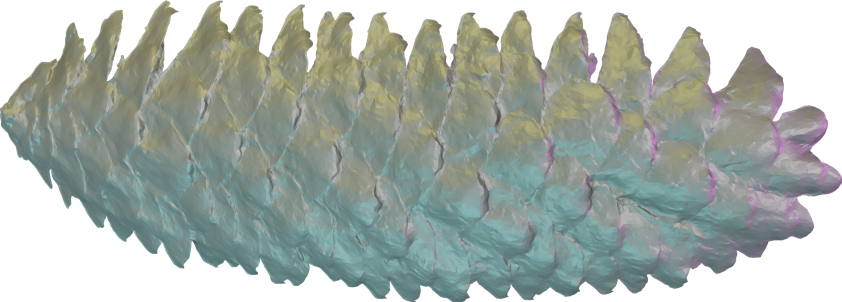}  \\
            \small{20951 (99.0\%)}
        \end{subfigure}
        \begin{subfigure}[b]{.23\linewidth}
            \centering
            \includegraphics[angle=90,origin=c,height=40mm]{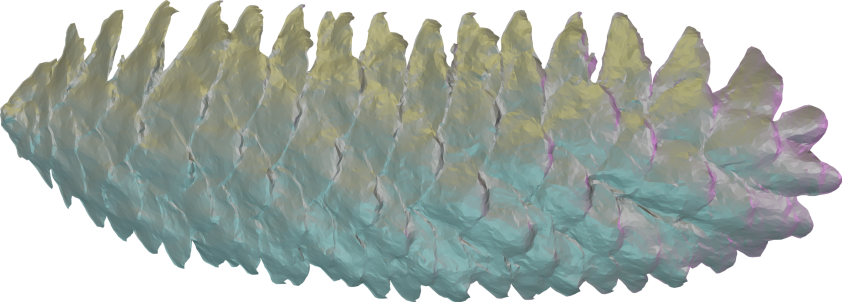}  \\
            \small{15176 (99.3\%)}
        \end{subfigure}
    \end{subfigure}
    \\
    \begin{subfigure}[b]{.8\linewidth}
        \centering
        \LARGE{\textsc{Soldier}} \\ \vspace{1mm}
        \begin{subfigure}[b]{.23\linewidth}
            \centering
            \includegraphics[angle=90,origin=c,height=40mm]{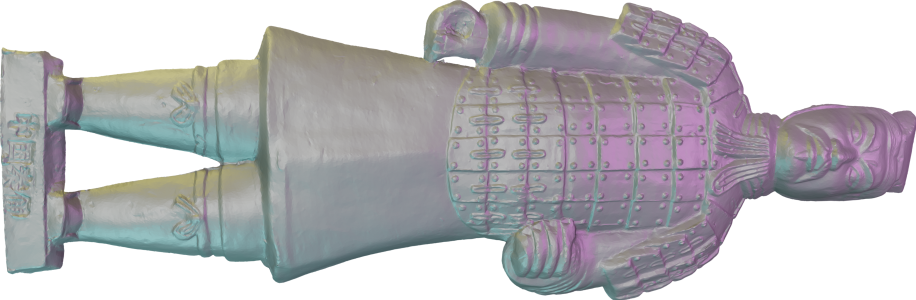}  \\
            \small{49992 (98.5\%)}
        \end{subfigure}
        \begin{subfigure}[b]{.23\linewidth}
            \centering
            \includegraphics[angle=90,origin=c,height=40mm]{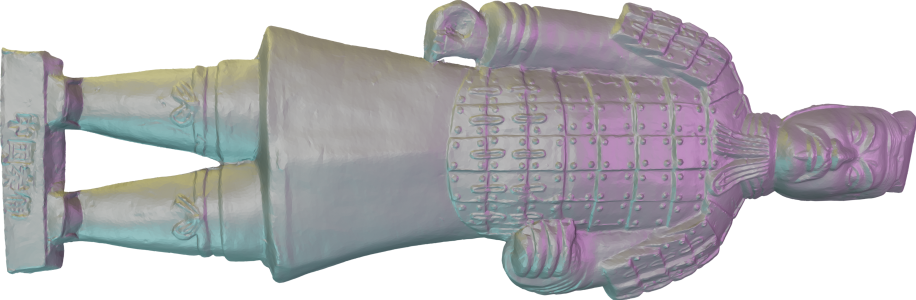}  \\
            \small{36122 (98.9\%)}
        \end{subfigure}
        \begin{subfigure}[b]{.23\linewidth}
            \centering
            \includegraphics[angle=90,origin=c,height=40mm]{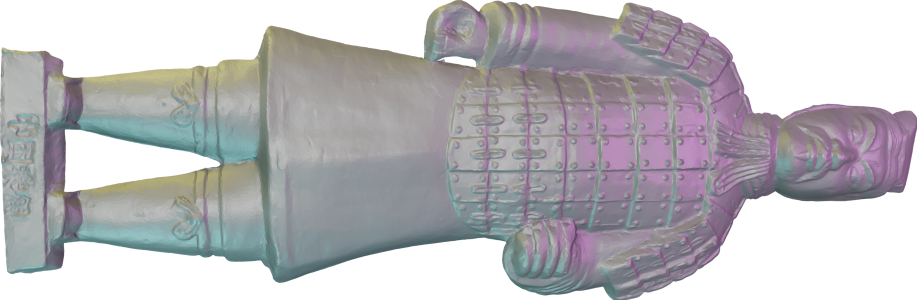}  \\
            \small{26075 (99.2\%)}
        \end{subfigure}
        \begin{subfigure}[b]{.23\linewidth}
            \centering
            \includegraphics[angle=90,origin=c,height=40mm]{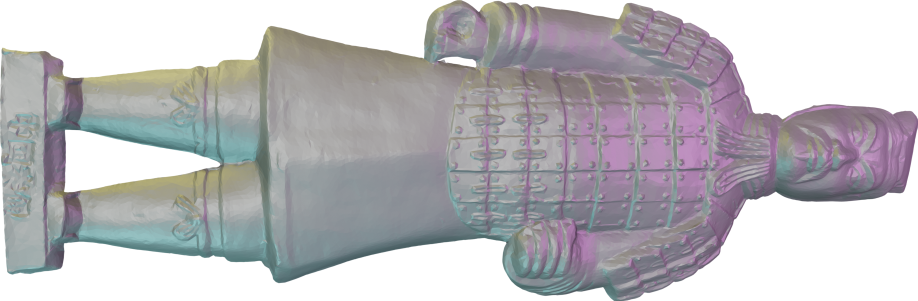}  \\
            \small{18788 (99.4\%)}
        \end{subfigure}
    \end{subfigure}
    \caption{Reconstruction results for the RGBN dataset \cite{Toler-Franklin:2007} for decimation thresholds of 8, 16, 32 and 64. Objects were rotated to the upright position.}
    \label{fig:RGBN_second_figure}
\end{figure*}

\begin{figure*}
    \centering
    \Huge{PS Dataset (1 of 2)}   \\
    \Large{High-Res} \hfill \large{Low-Res}\\
    \vspace{-3mm}
    \hrulefill
    \\  
    \LARGE{\textsc{Cat}} \\ \vspace{1mm}
    \begin{subfigure}[b]{.22\linewidth}
        \centering
        \includegraphics[width=\linewidth]{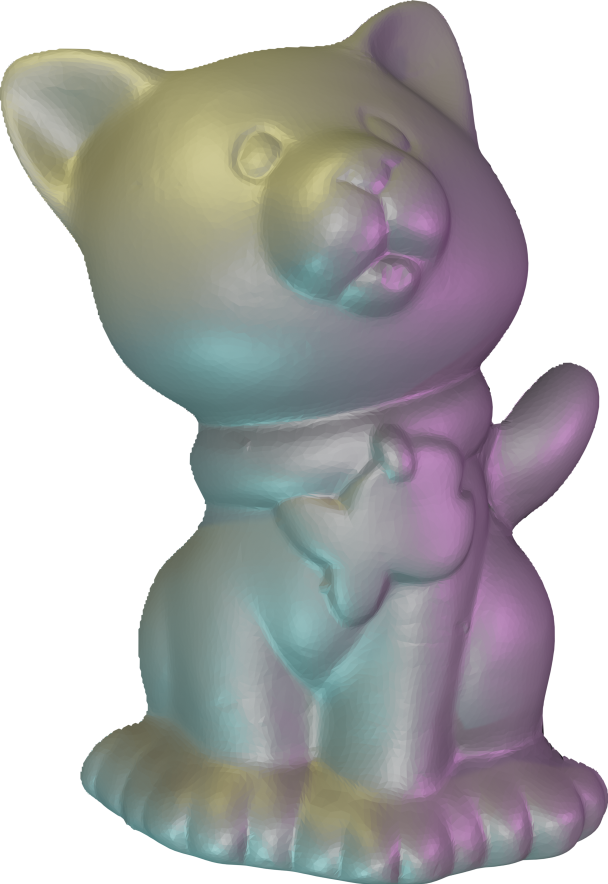}  \\
        \small{14518 (91.5\%)}
    \end{subfigure}
    \begin{subfigure}[b]{.22\linewidth}
        \centering
        \includegraphics[width=\linewidth]{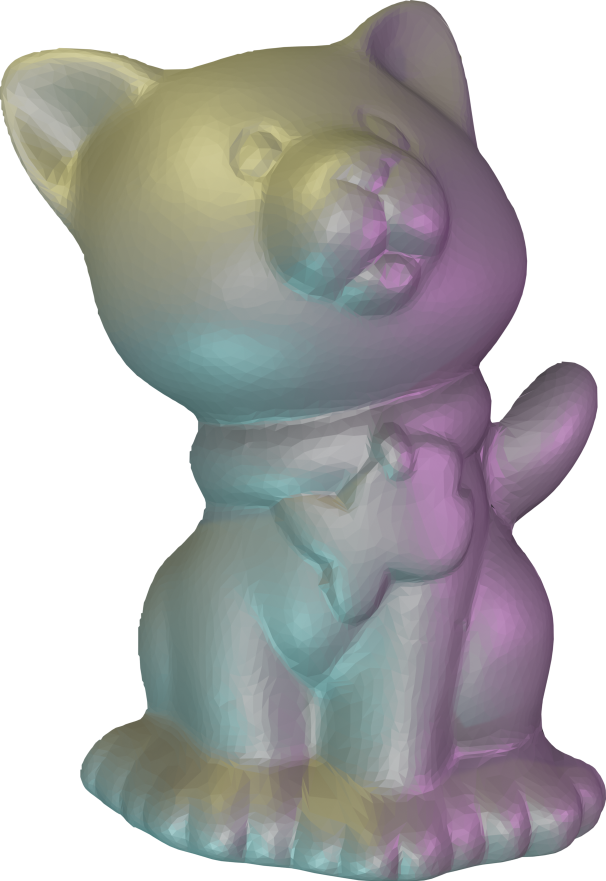}  \\
        \small{5445 (96.8\%)}
    \end{subfigure}
    \begin{subfigure}[b]{.22\linewidth}
        \centering
        \includegraphics[width=\linewidth]{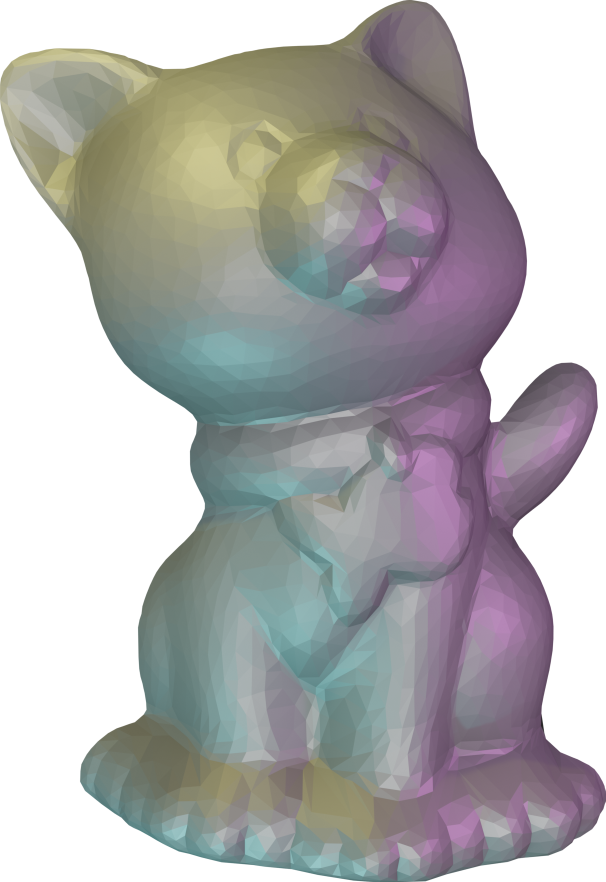}  \\
        \small{1983 (98.8\%)}
    \end{subfigure}
    \begin{subfigure}[b]{.22\linewidth}
        \centering
        \includegraphics[width=\linewidth]{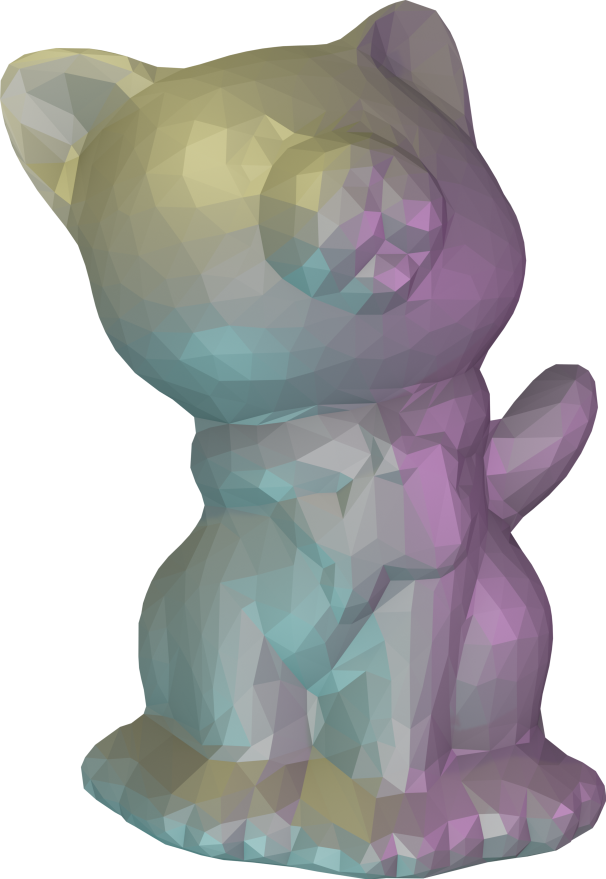}  \\
        \small{754 (99.6\%)}
    \end{subfigure}
    \\
    \LARGE{\textsc{Frog}} \\ \vspace{1mm}
    \begin{subfigure}[b]{.22\linewidth}
        \centering
        \includegraphics[width=\linewidth]{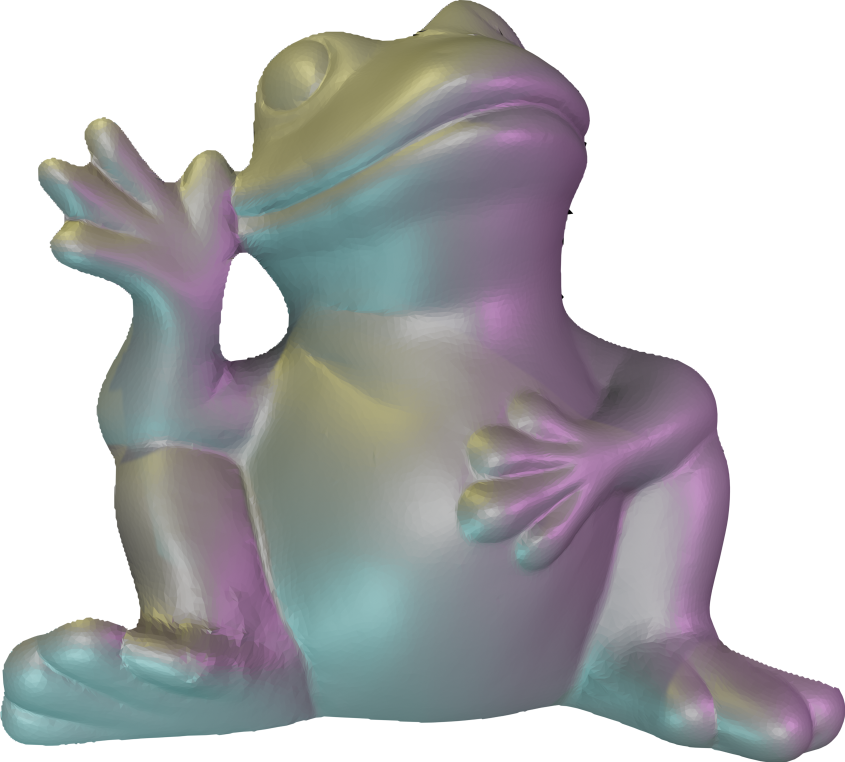}  \\
        \small{16492 (92.2\%)}
    \end{subfigure}
    \begin{subfigure}[b]{.22\linewidth}
        \centering
        \includegraphics[width=\linewidth]{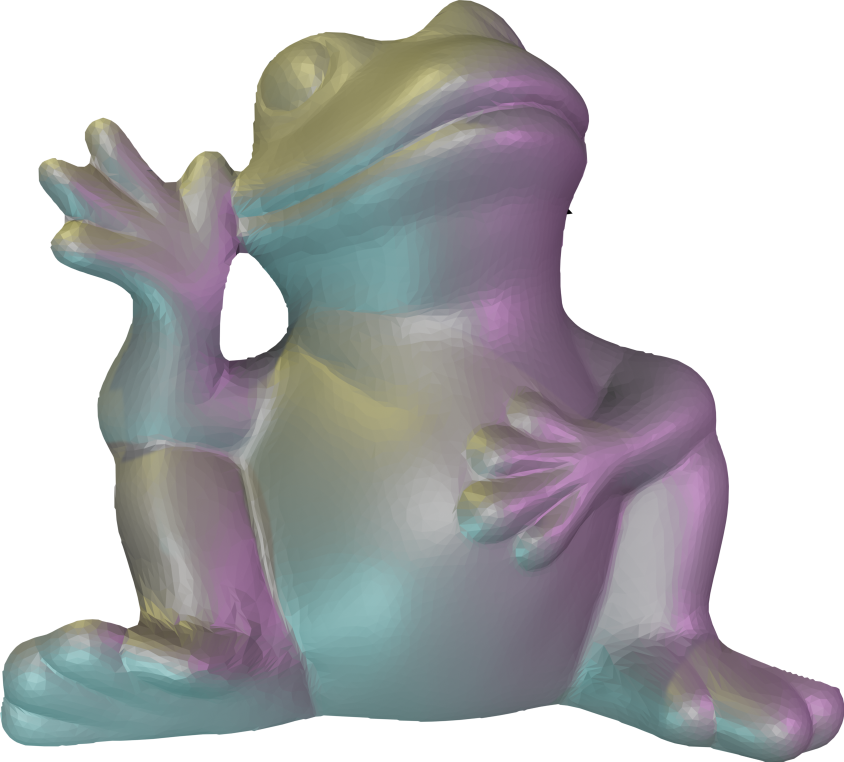}  \\
        \small{6155 (97.1\%)}
    \end{subfigure}
    \begin{subfigure}[b]{.22\linewidth}
        \centering
        \includegraphics[width=\linewidth]{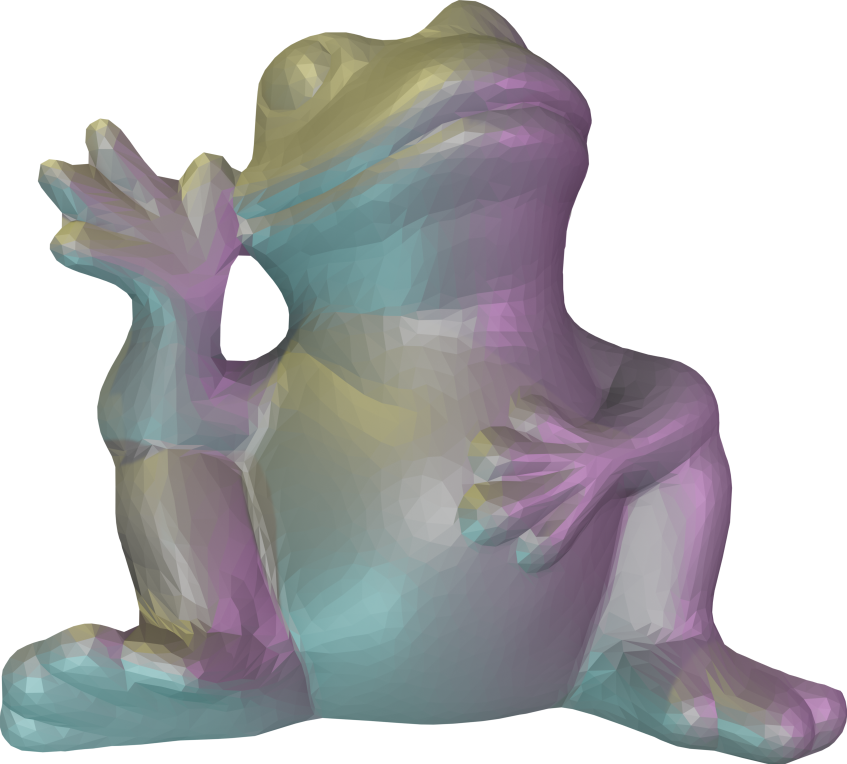}  \\
        \small{2275 (98.9\%)}
    \end{subfigure}
    \begin{subfigure}[b]{.22\linewidth}
        \centering
        \includegraphics[width=\linewidth]{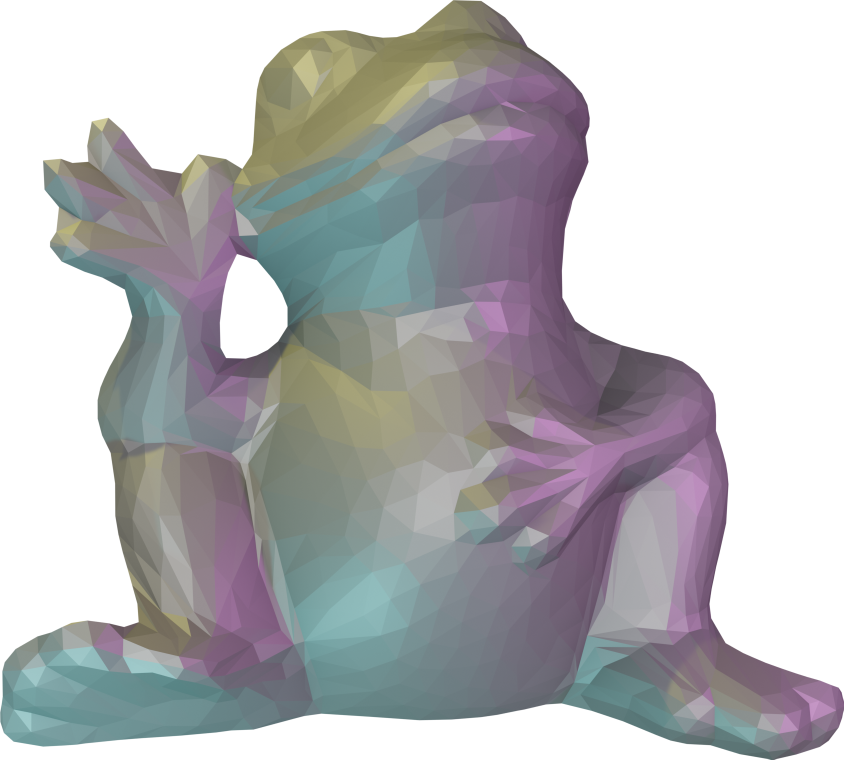}  \\
        \small{844 (99.6\%)}
    \end{subfigure}
    \\
    \LARGE{\textsc{Hippo}} \\ \vspace{1mm}
    \begin{subfigure}[b]{.22\linewidth}
        \centering
        \includegraphics[width=\linewidth]{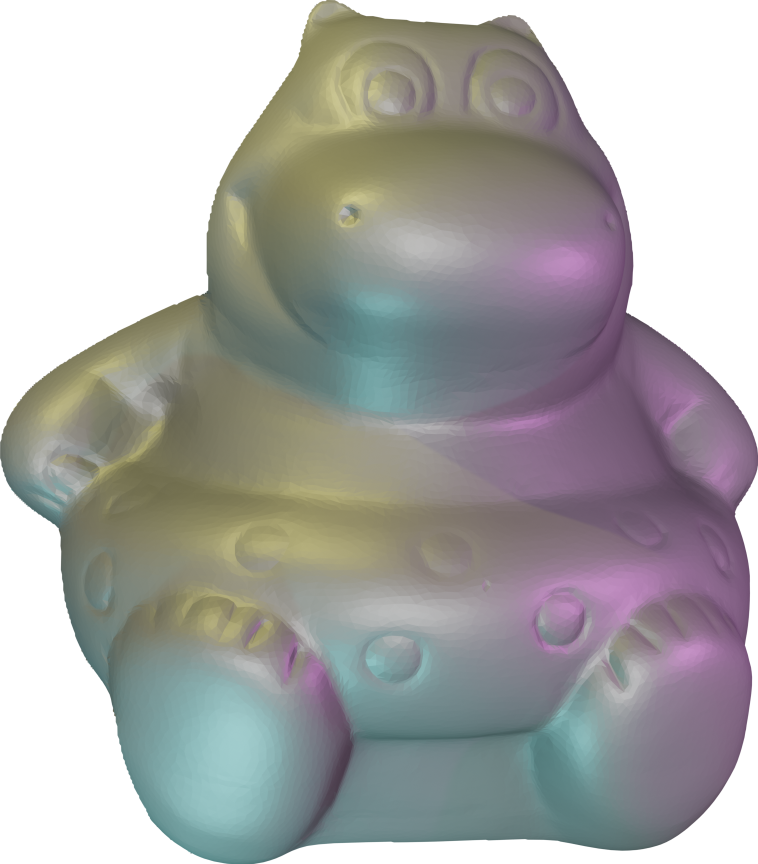}  \\
        \small{15497 (91.7\%)}
    \end{subfigure}
    \begin{subfigure}[b]{.22\linewidth}
        \centering
        \includegraphics[width=\linewidth]{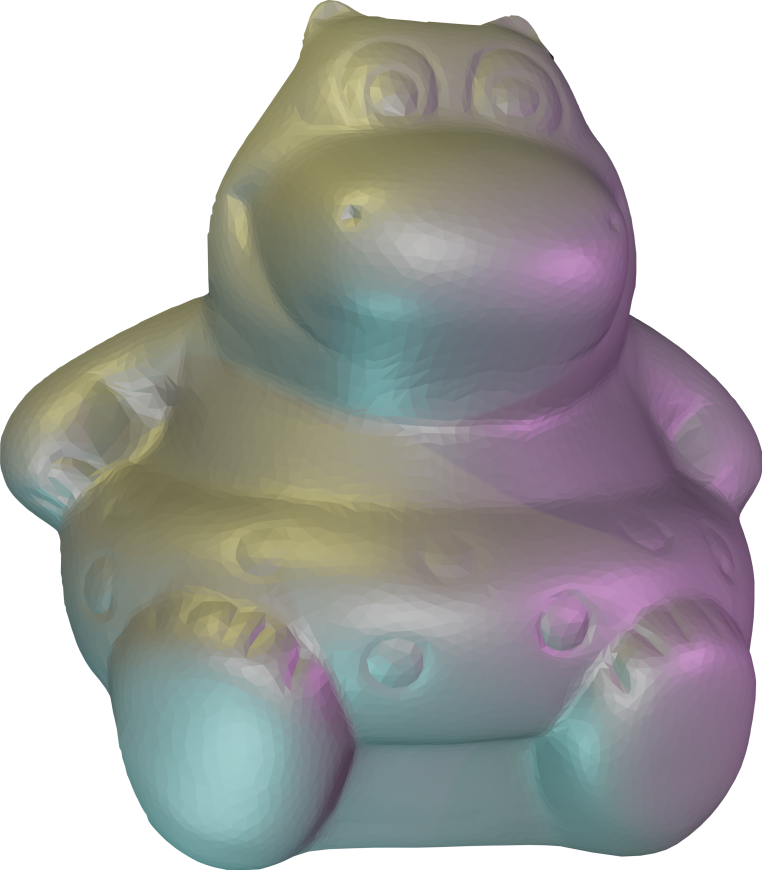}  \\
        \small{5651 (97.0\%)}
    \end{subfigure}
    \begin{subfigure}[b]{.22\linewidth}
        \centering
        \includegraphics[width=\linewidth]{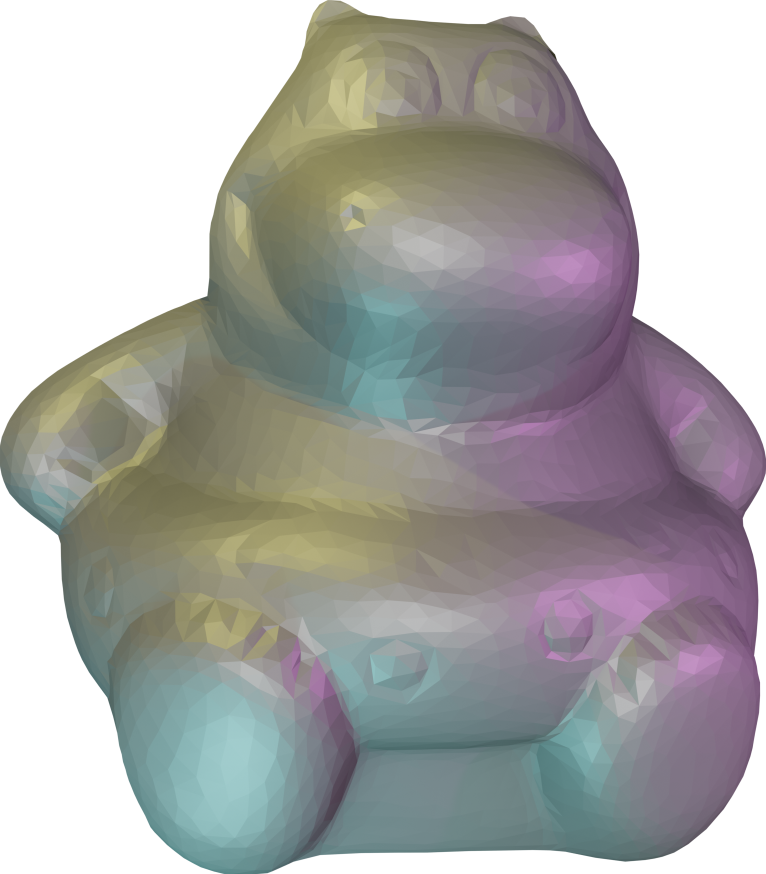}  \\
        \small{2086 (98.9\%)}
    \end{subfigure}
    \begin{subfigure}[b]{.22\linewidth}
        \centering
        \includegraphics[width=\linewidth]{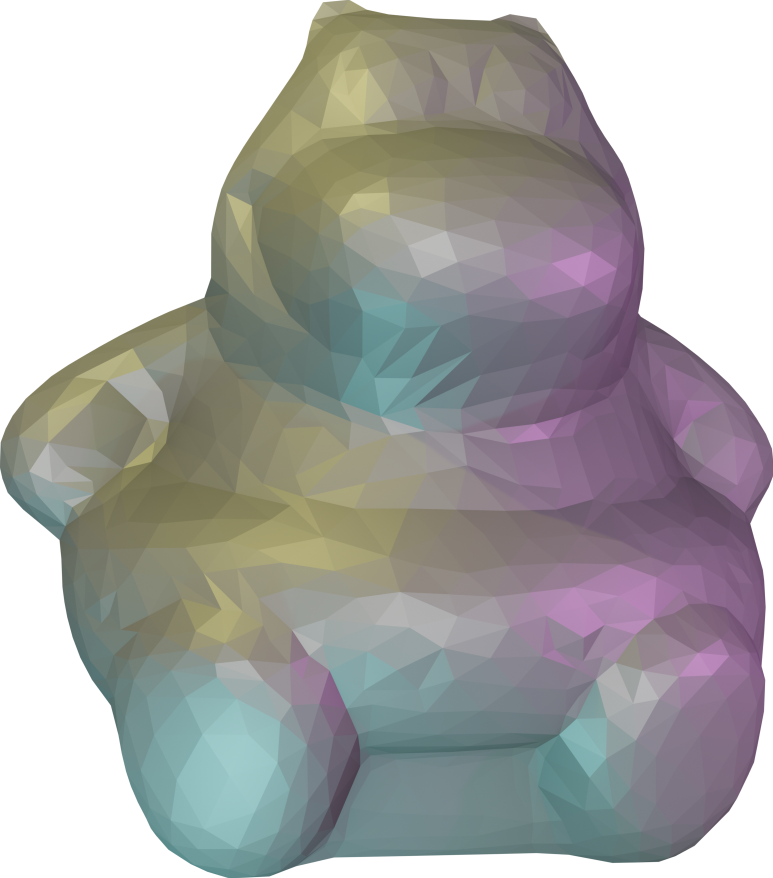}  \\
        \small{757 (99.6\%)}
    \end{subfigure}
    \\
    \caption{Reconstruction results for the PS dataset \cite{Frankot:1988} with decimation thresholds of 0.125, 1, 8 and 64. The decimation threshold increases from left to right, \ie mesh resolution decreases from left to right.}
    \label{fig:PS_first_figure}
\end{figure*}

\begin{figure*}
    \centering
    \Huge{PS Dataset (2 of 2)}   \\
    \Large{High-Res} \hfill \large{Low-Res}\\
    \vspace{-3mm}
    \hrulefill
    \\  
    \LARGE{\textsc{Lizard}} \\ \vspace{1mm}
    \begin{subfigure}[b]{.22\linewidth}
        \centering
        \includegraphics[width=\linewidth]{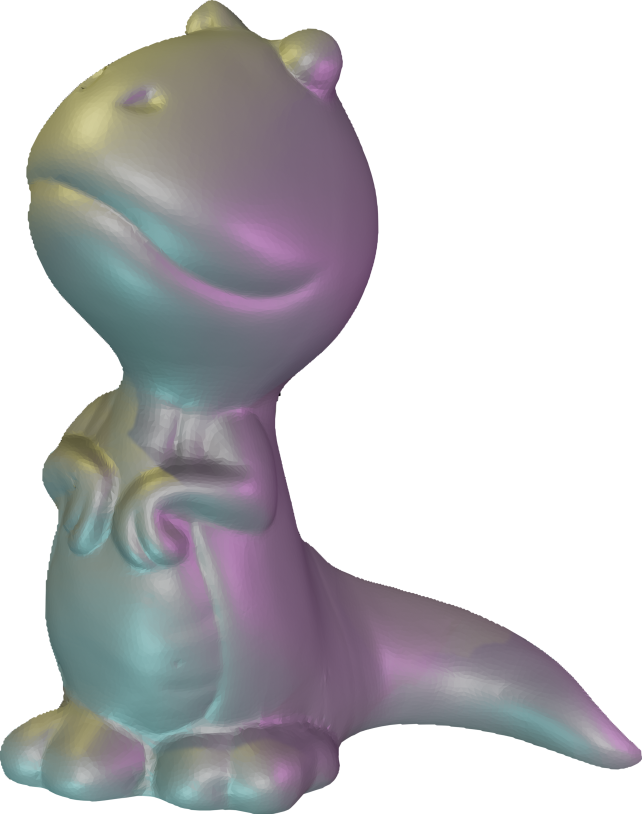}  \\
        \small{14655 (92.0\%)}
    \end{subfigure}
    \begin{subfigure}[b]{.22\linewidth}
        \centering
        \includegraphics[width=\linewidth]{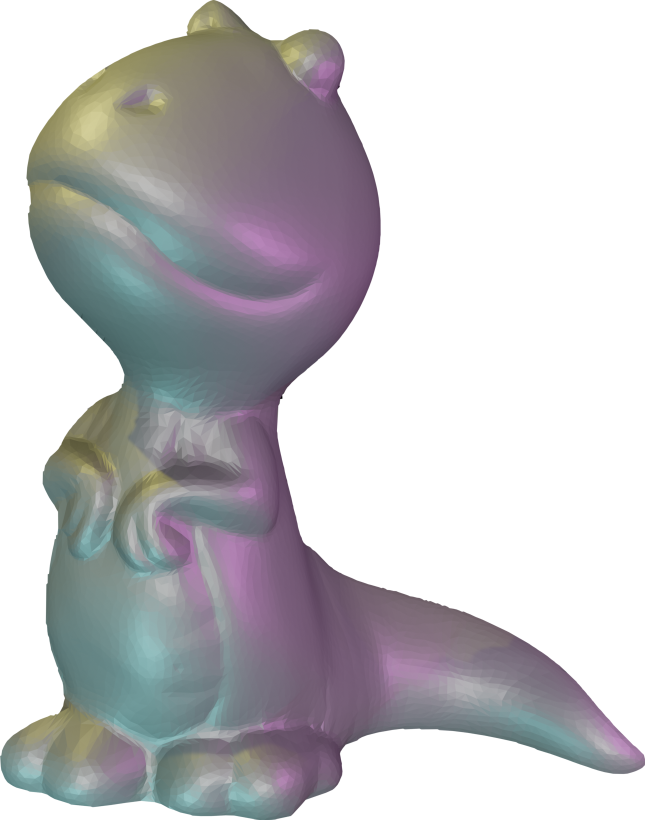}  \\
        \small{5476 (97.0\%)}
    \end{subfigure}
    \begin{subfigure}[b]{.22\linewidth}
        \centering
        \includegraphics[width=\linewidth]{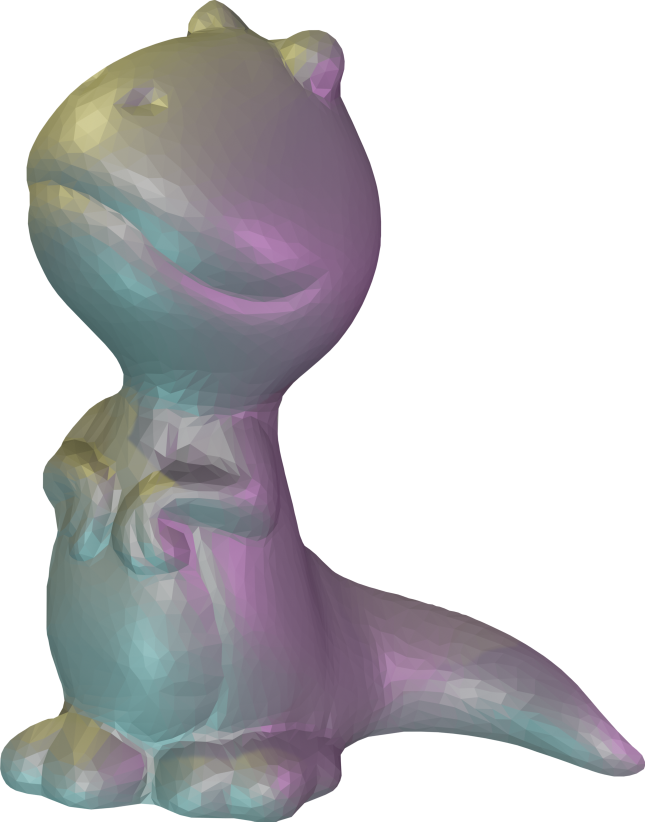}  \\
        \small{1987 (98.9\%)}
    \end{subfigure}
    \begin{subfigure}[b]{.22\linewidth}
        \centering
        \includegraphics[width=\linewidth]{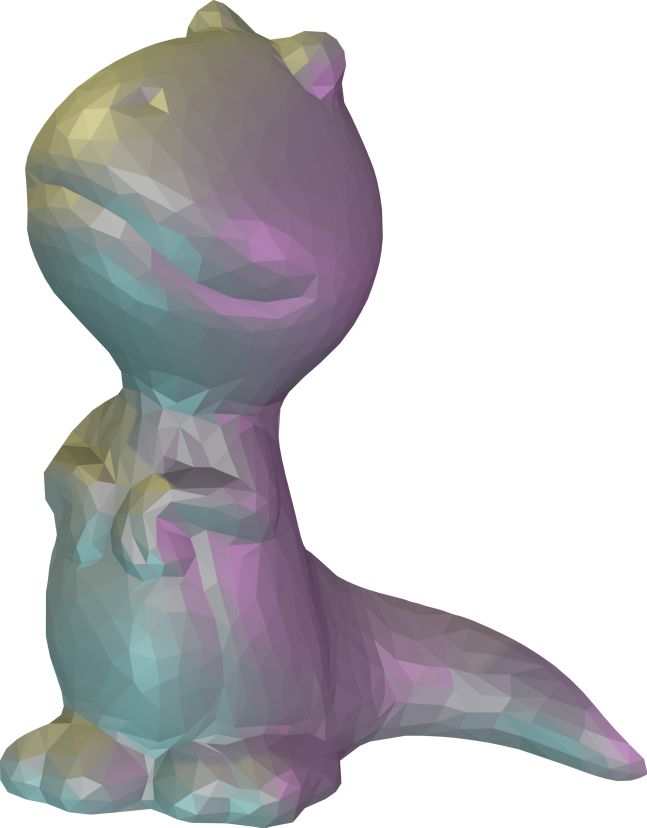}  \\
        \small{742 (99.6\%)}
    \end{subfigure}
    \\
    \LARGE{\textsc{Pig}} \\ \vspace{1mm}
    \begin{subfigure}[b]{.22\linewidth}
        \centering
        \includegraphics[width=\linewidth]{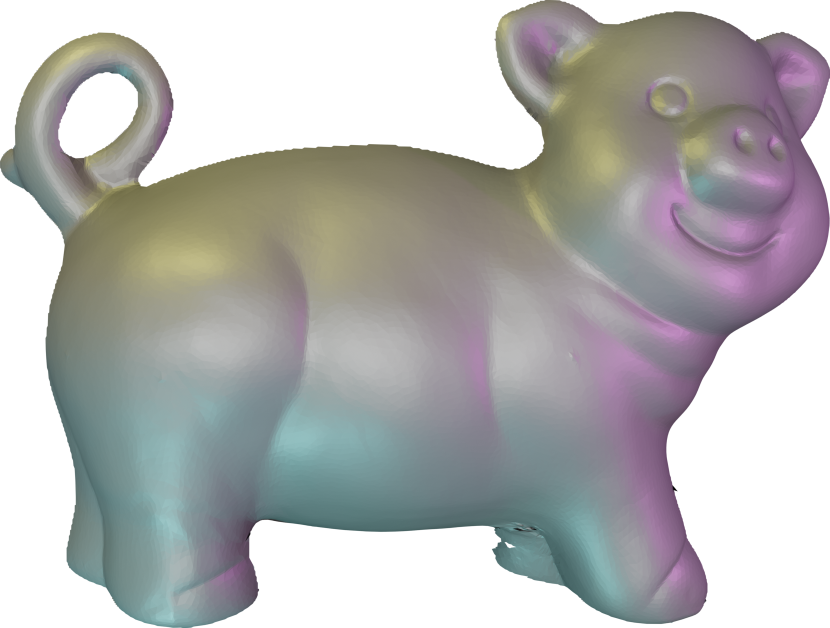}  \\
        \small{15934 (93.0\%)}
    \end{subfigure}
    \begin{subfigure}[b]{.22\linewidth}
        \centering
        \includegraphics[width=\linewidth]{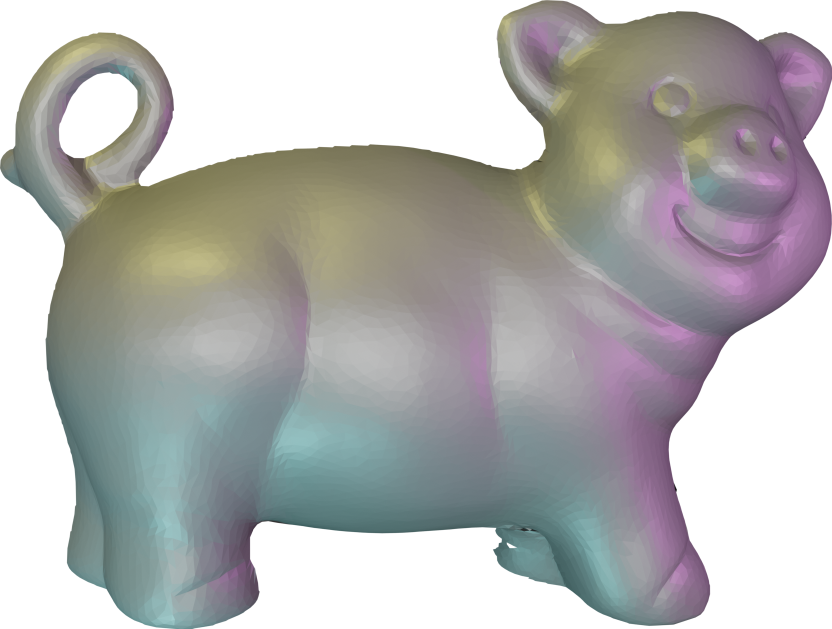}  \\
        \small{5994 (97.4\%)}
    \end{subfigure}
    \begin{subfigure}[b]{.22\linewidth}
        \centering
        \includegraphics[width=\linewidth]{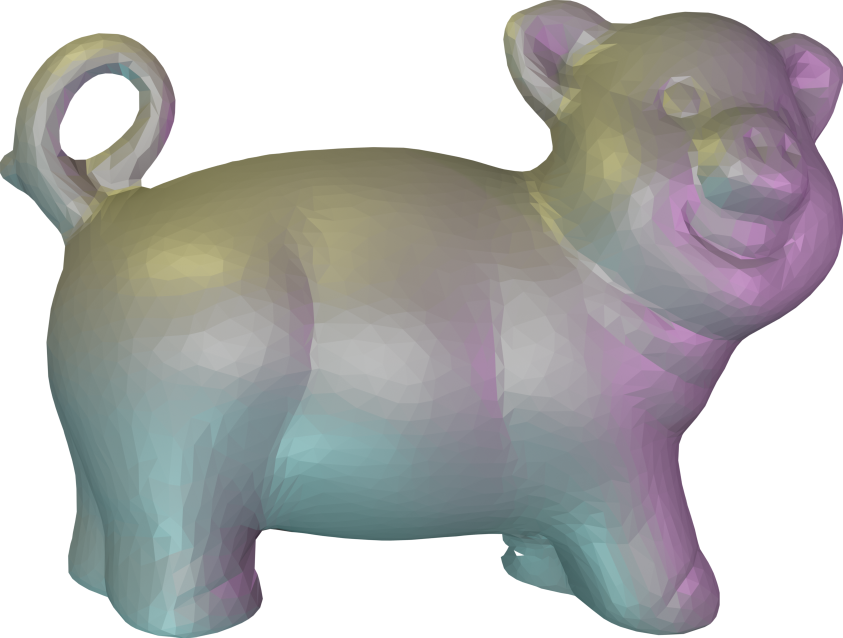}  \\
        \small{2188 (99.0\%)}
    \end{subfigure}
    \begin{subfigure}[b]{.22\linewidth}
        \centering
        \includegraphics[width=\linewidth]{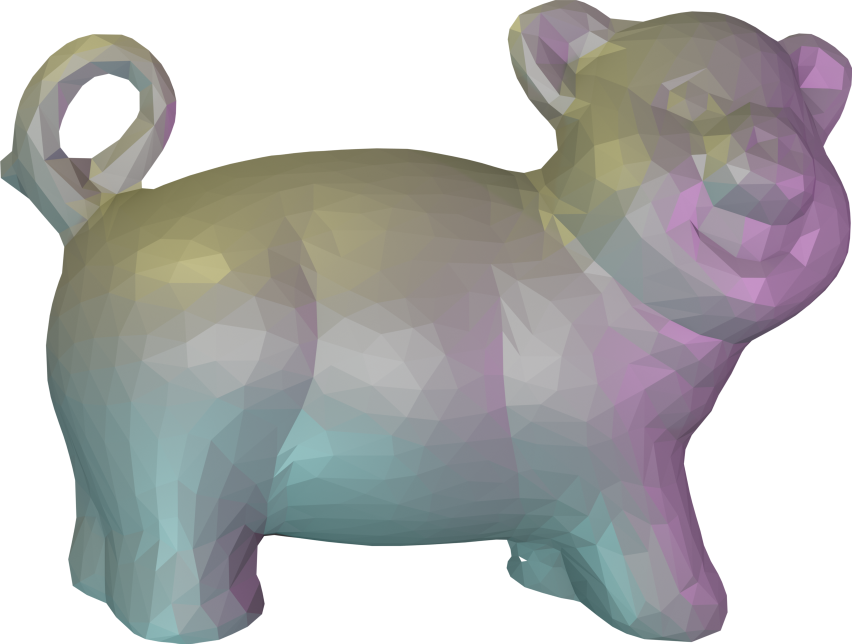}  \\
        \small{827 (99.6\%)}
    \end{subfigure}
    \\
    \LARGE{\textsc{Scholar}} \\ \vspace{1mm}
    \begin{subfigure}[b]{.22\linewidth}
        \centering
        \includegraphics[width=\linewidth]{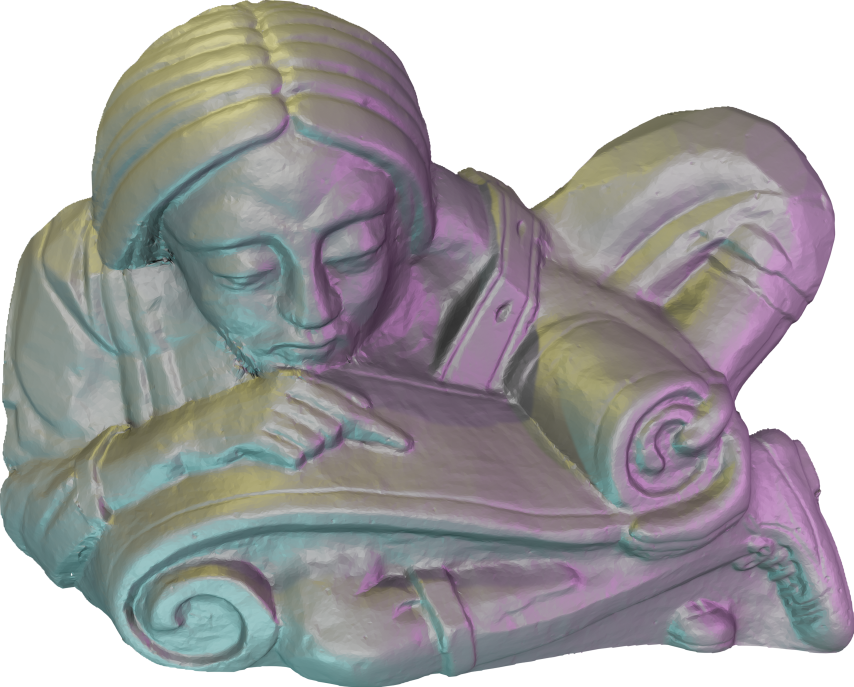}  \\
        \small{64685 (89.3\%)}
    \end{subfigure}
    \begin{subfigure}[b]{.22\linewidth}
        \centering
        \includegraphics[width=\linewidth]{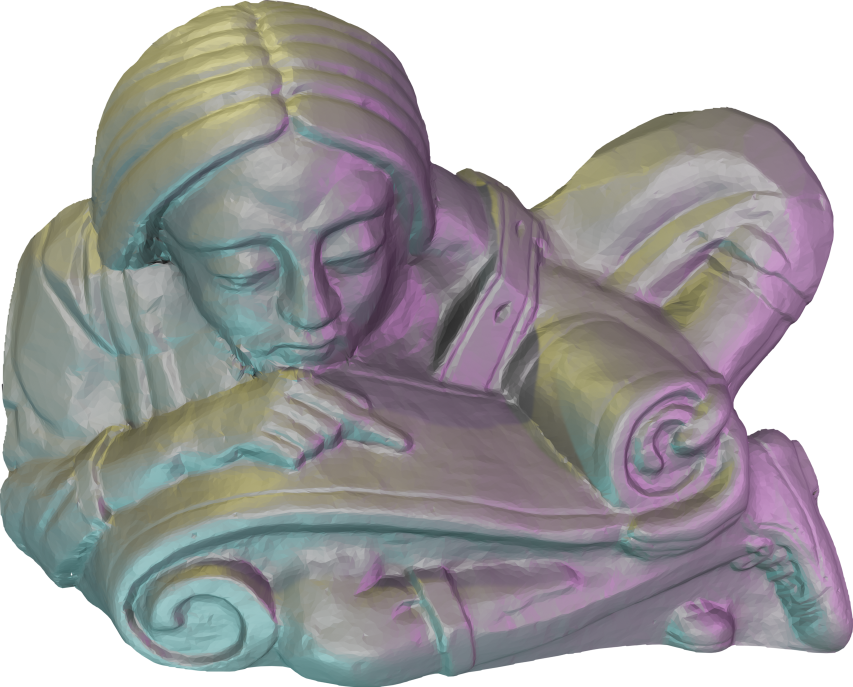}  \\
        \small{24678 (95.9\%)}
    \end{subfigure}
    \begin{subfigure}[b]{.22\linewidth}
        \centering
        \includegraphics[width=\linewidth]{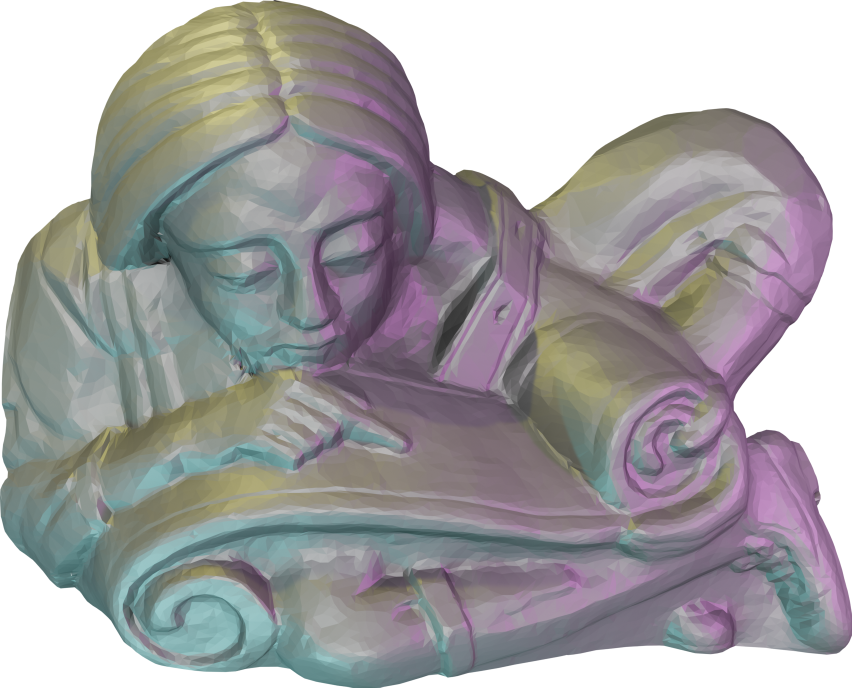}  \\
        \small{9204 (98.5\%)}
    \end{subfigure}
    \begin{subfigure}[b]{.22\linewidth}
        \centering
        \includegraphics[width=\linewidth]{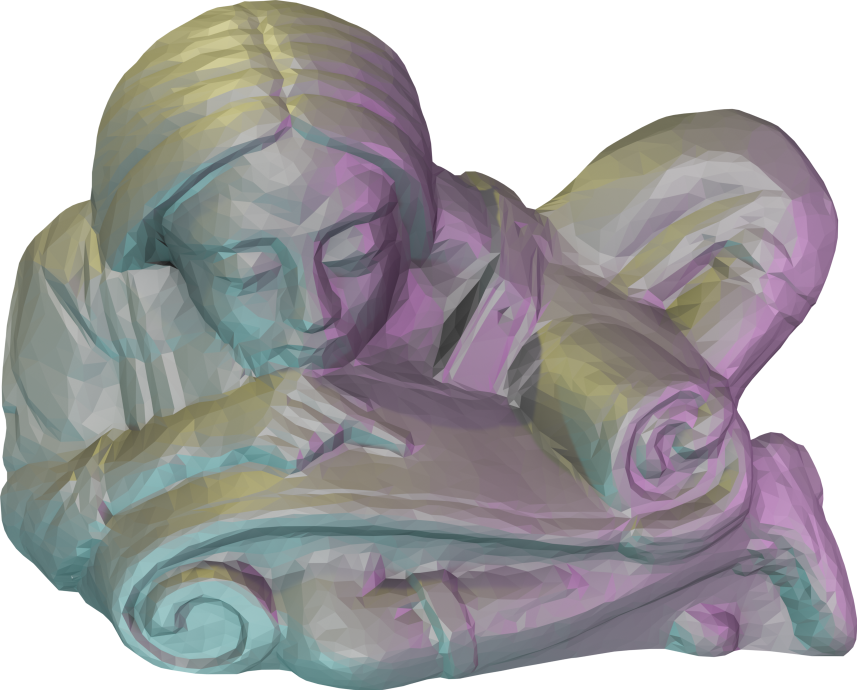}  \\
        \small{3425 (99.4\%)}
    \end{subfigure}
    \\   %
    \LARGE{\textsc{Turtle}} \\ \vspace{1mm}
    \begin{subfigure}[b]{.22\linewidth}
        \centering
        \includegraphics[width=\linewidth]{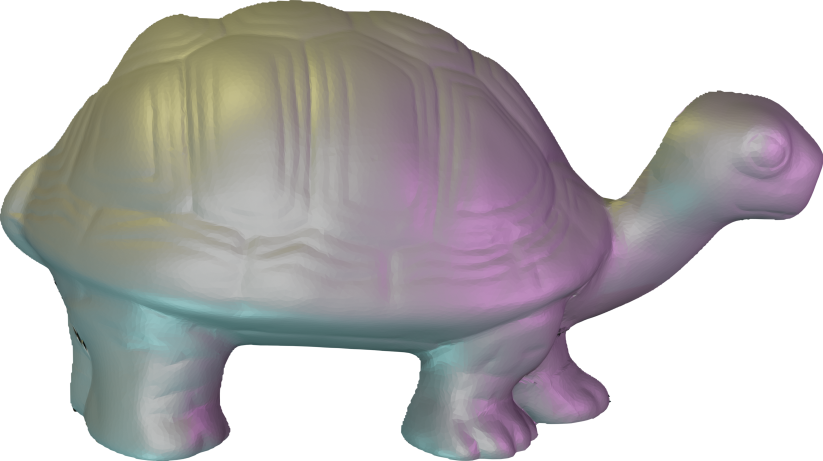}  \\
        \small{16850 (91.7\%)}
    \end{subfigure}
    \begin{subfigure}[b]{.22\linewidth}
        \centering
        \includegraphics[width=\linewidth]{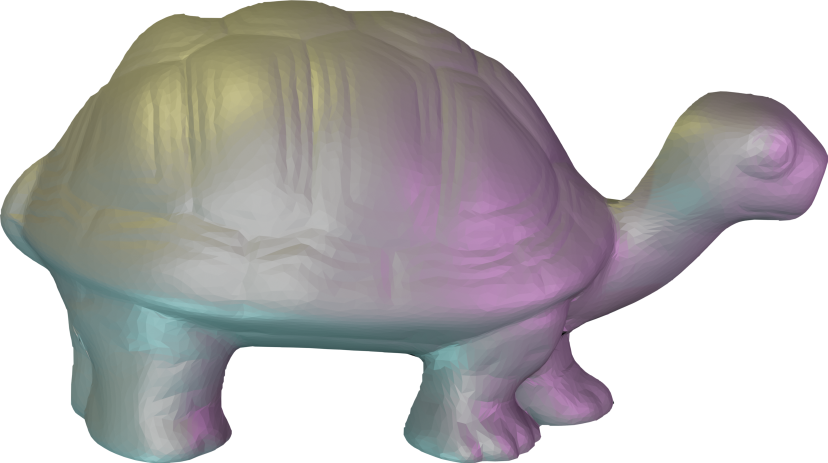}  \\
        \small{6236 (96.9\%)}
    \end{subfigure}
    \begin{subfigure}[b]{.22\linewidth}
        \centering
        \includegraphics[width=\linewidth]{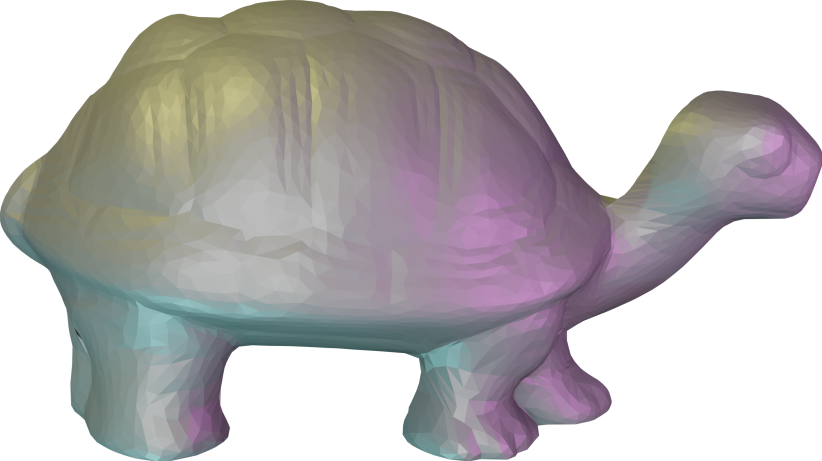}  \\
        \small{2286 (98.9\%)}
    \end{subfigure}
    \begin{subfigure}[b]{.22\linewidth}
        \centering
        \includegraphics[width=\linewidth]{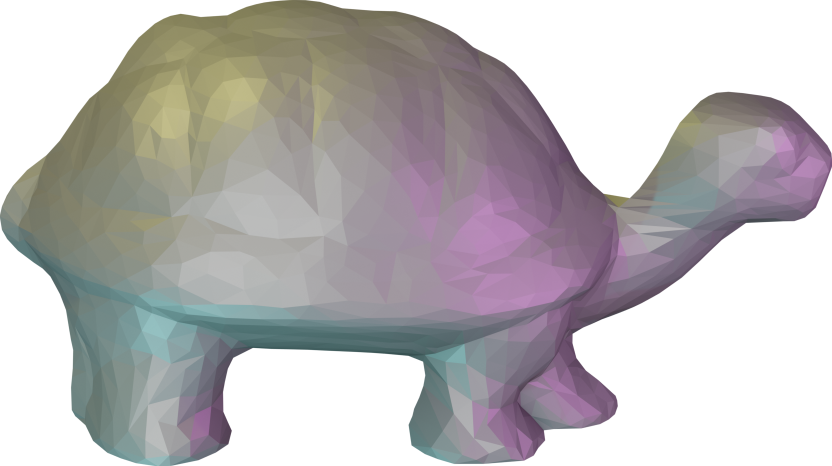}  \\
        \small{865 (99.6\%)}
    \end{subfigure}
    \\
    \caption{Reconstruction results for the PS dataset \cite{Frankot:1988} with decimation thresholds of 0.125, 1, 8 and 64. The decimation threshold increases from left to right, \ie mesh resolution decreases from left to right.}
    \label{fig:PS_second_figure}
\end{figure*}

Finally, we show reconstructions from all the datasets we used. Except for LUCES \cite{Mecca:2021}, these datasets come without ground-truth geometry. The RMSEs for all objects in LUCES were reported in \cref{tab:luces} which complements Tab.\ 2 from the submitted manuscript. For visual inspection, all objects can be found in \cref{fig:LUCES_first_figure} to \cref{fig:PS_second_figure}. We also indicate the vertex count and compression ratio to put the results into perspective.

\clearpage
\clearpage
\FloatBarrier
\clearpage
\section{Overview of all Datasets}
\label{sec:datasets}
In this work, we used the following photometric stereo datasets:
\begin{itemize}
    \item DiLiGenT-MV \cite{Li:2020}: \url{https://sites.google.com/site/photometricstereodata/mv}
    \item LUCES \cite{Mecca:2021}: \url{http://www.robertomecca.com/luces.html}
    \item RGBN \cite{Toler-Franklin:2007}: \url{https://gfx.cs.princeton.edu/gfx/proj/rgbn/}
    \item PS Dataset \cite{Frankot:1988}: \url{https://vision.seas.harvard.edu/qsfs/Data.html}
\end{itemize}
Furthermore, we generated synthetic datasets using the following 3D models:
\begin{itemize}
    \item David Head [1d\_inc]: \url{https://sketchfab.com/models/39a4d01bef37495cac8d8f0009728871/} 
    \item Football Medal 2 [Cain]: \url{https://sketchfab.com/models/54d54534f11d4d23aecb945fd7eb1df4/}
    \item Female Head: \url{https://www.3dscanstore.com/3d-head-models/raw-expression-bundles/female-02-x36-expression-bundle}
    \item Male Head: \url{https://www.3dscanstore.com/3d-head-models/raw-expression-bundles/male-01-36x-expression-scan-bundle}
\end{itemize}
\twocolumn
{
    \small
    \bibliographystyle{ieeenat_fullname}
    \bibliography{cvpr2025}
}